\documentclass{article}

 \PassOptionsToPackage{numbers, compress}{natbib}

\usepackage[final]{nips_2018} 

\usepackage{graphicx}
\usepackage{subfigure}

\usepackage{graphicx} 
\usepackage{wrapfig} 
\usepackage{float}	  
\usepackage{calc} %
\usepackage{caption}
\usepackage{natbib} 

\usepackage{fontawesome} 
\usepackage[nolist,smaller]{acronym}

\usepackage{algorithm}
\usepackage[noend]{algpseudocode}

\usepackage{amsfonts, amsmath, amsthm} 
\usepackage{cancel} 
\usepackage{bbm, bm} 
\usepackage{IEEEtrantools}
\theoremstyle{definition}
\newtheorem{theorem}{Theorem} 

\usepackage{xcolor}

\definecolor{navy}{RGB}{0, 0, 128}
\definecolor{pythonPurple}{RGB}{148,0,211}
\definecolor{darkgreen}{rgb}{0,0.3,0}
\definecolor{pythonOrange}{RGB}{255,165,0}

\usepackage{tikz} 
\usetikzlibrary{arrows,shapes}
\usepackage{bigints} 

\usepackage[utf8]{inputenc} 
\usepackage[T1]{fontenc}    
\usepackage{hyperref}       
\usepackage{url}            
\usepackage{booktabs}       
\usepackage{amsfonts}       
\usepackage{nicefrac}       
\usepackage{microtype}      
\usepackage{xspace}

\usepackage{xcolor} 
\usepackage[utf8]{inputenc}
\usepackage[english]{babel}
\usepackage{amsthm}
\usepackage{multicol}

\DeclareMathOperator*{\argmax}{argmax}
\DeclareMathOperator*{\argmin}{argmin}
\def\*#1{\boldsymbol{#1}} 
\newcommand{\acro}[1]{\textsc{#1}\xspace}

\newcommand{\BOCPD}{\acro{\smaller BOCPD}}
\newcommand{\BD}{\acro{$\beta$-D}}
\newcommand{\brlmm}{\beta_{\text{rlm}}} 
\newcommand{\bpm}{\beta_{\text{p}}}
\newcommand{\brlm}{$\beta_{\text{rlm}}$} 
\newcommand{\bp}{$\beta_{\text{p}}$}
\newcommand{\brlmmtm}{\beta_{\text{rlm},t}}
\newcommand{\bpmtm}{\beta_{\text{p}, t}}

\newcommand{\KLD}{\acro{\smaller KLD}}
\newcommand{\GBI}{\acro{\smaller GBI}}
\newcommand{\CP}{\acro{\smaller CP}}
\newcommand{\CPs}{\acro{\smaller CPs}}
\newcommand{\BLR}{\acro{\smaller BLR}} 
\newcommand{\SGD}{\acro{\smaller SGD}}
\newcommand{\SVRG}{\acro{\smaller SVRG}}

\newcommand{\MCMC}{\acro{\smaller MCMC}}
\newcommand{\ELBO}{\acro{\smaller ELBO}}
\newcommand{\SVI}{\acro{\smaller SVI}} 
\newcommand{\DGM}{\acro{\smaller DGM}} 
\newcommand{\NOX}{\acro{\smaller NOX}}
\newcommand{\AR}{\acro{\smaller AR}}
\newcommand{\ARs}{\acro{\smaller ARs}}
\newcommand{\NIG}{\acro{\smaller NIG}}
\newcommand{\MD}{\acro{\smaller MD}}
\newcommand{\MAP}{\acro{\smaller MAP}}
\newcommand{\FDR}{\acro{\smaller FDR}}
\newcommand{\VAR}{\acro{\smaller VAR}}

\newcommand{\GHz}{\acro{\smaller GHz}}
\newcommand{\iseven}{{i7}}
\newcommand{\GB}{\acro{\smaller GB}}

\newcommand{\RAM}{\acro{\smaller RAM}}
\newcommand{\EPSRC}{\acro{\smaller EPSRC}}
\newcommand{\OxWaSP}{\acro{\smaller OxWaSP}}

\usepackage{mathabx, amssymb}

\newcommand{\E}{\mathbb{E}}
\newcommand{\srv}{\sigma^2}
\newcommand{\berv}{\bm{\mu}}

\newcommand{\an}{\widehat{a}_n}
\newcommand{\bn}{\widehat{b}_n}
\newcommand{\Sn}{\widehat{\bm{\Sigma}}_n}
\newcommand{\Sninv}{\widehat{\bm{\Sigma}}_n^{-1}}
\newcommand{\ben}{\widehat{\bm{\mu}}_n}
\newcommand{\NIGVB}{
	\text{NIG}^{\text{VB}}(
		\*\mu, \sigma^2 |
			\widehat{a}_n, \widehat{b}_n, \widehat{\bm{\mu}}_n, 	
				\widehat{\bm{\Sigma}}_n
		)
}

\newcommand{\az}{{a}_0}
\newcommand{\bz}{{b}_0}
\newcommand{\Sz}{{\bm{\Sigma}}_0}
\newcommand{\Szinv}{{\bm{\Sigma}}_0^{-1}}
\newcommand{\bez}{{\bm{\mu}}_0}
\newcommand{\NIGz}{
	\text{NIG}^{0}(
		\*\mu, \sigma^2 |
			{a}_0, {b}_0, {\bm{\mu}}_0, 	
				{\bm{\Sigma}}_0
		)
}

\newcommand{\Sci}{\widecheck{\bm{\Sigma}}_i}
\newcommand{\Sciinv}{\widecheck{\bm{\Sigma}}_i^{-1}}
\newcommand{\beci}{\widecheck{\bm{\mu}}_i}
\newcommand{\Sti}{\widetilde{\bm{\Sigma}}_i}
\newcommand{\Stiinv}{\widetilde{\bm{\Sigma}}_i^{-1}}
\newcommand{\beti}{\widetilde{\bm{\mu}}_i}

\newcommand{\dpd}{\pi_{\text{DPD}}^{\beta}}
\newcommand{\loss}{\ell^{\beta}}

\newcommand{\p}{\partial}
\newcommand{\tr}{\text{tr}}
\renewcommand{\v}{\text{vech}}

\newcommand{\DL}{
		\frac{\partial}
		{\partial {\text{vech}\left(\bm{L}\right)}}
	}
\newcommand{\Dben}{
		\frac{\partial}
		{\partial {\widehat{{\bm{\mu}}_n}}}
	}
\newcommand{\Dbn}{
		\frac{\partial}
		{\partial {\widehat{b}_n}}
	}
\newcommand{\Dan}{
		\frac{\partial}
		{\partial {\widehat{a}_n}}
	}

\newcommand{\Rnbr}{
		\left[ 
			\widehat{\bm{\Sigma}}_n^{-1}  +\beta \left( \bm{X}_i'\bm{X}_i  \right)
		\right]
	}
\newcommand{\Rninv}{
		\left[ 
			\widehat{\bm{\Sigma}}_n^{-1} +\beta \left( \bm{X}_i'\bm{X}_i  \right)
		\right]^{-1}
	}
\newcommand{\Rndet}{
		\left| 
			\widehat{\bm{\Sigma}}_n^{-1}+\beta \left( \bm{X}_i'\bm{X}_i \right)
		 \right|
	}
\renewcommand{\L}{\bm{L}}
\newcommand{\R}{\bm{R}}
\newcommand{\B}{\bm{B}}

\usepackage[toc,page,header]{appendix}
\usepackage{minitoc}

\title{Doubly Robust Bayesian Inference for Non-Stationary Streaming Data with $\beta$-Divergences 
}

\author{
  Jeremias Knoblauch\\
  The Alan Turing Institute\\
  Department of Statistics\\
  University of Warwick\\
  Coventry, CV4 7AL \\
  \texttt{j.knoblauch@warwick.ac.uk} \\
   \And
  Jack Jewson\\
  Department of Statistics\\
  University of Warwick\\
  Coventry, CV4 7AL \\
  \texttt{j.e.jewson@warwick.ac.uk} \\
   \AND
  Theodoros Damoulas\\
  The Alan Turing Institute\\
  Department of Computer Science \& Department of Statistics\\
  University of Warwick\\
  Coventry, CV4 7AL \\
  \texttt{t.damoulas@warwick.ac.uk} \\
}

\begin{document}

\doparttoc 
\faketableofcontents 

\maketitle

\begin{abstract}
We present the very first robust Bayesian Online Changepoint Detection algorithm through General Bayesian Inference (\GBI) with $\beta$-divergences. The resulting inference procedure is doubly robust for both the parameter and the changepoint (\CP) posterior, with linear time and constant space complexity. We provide a construction for exponential models and demonstrate it on the Bayesian Linear Regression model. 
In so doing, we make two additional contributions: Firstly, we make \GBI scalable using Structural Variational approximations that are exact as $\beta \to 0$. Secondly, we give a principled way of choosing the divergence parameter $\beta$ by minimizing expected predictive loss on-line. 
Reducing False Discovery Rates of \CPs from more than  90\% to 0\% on real world data, this offers the state of the art.
\end{abstract}

\section{Introduction}

Modeling non-stationary time series 
with changepoints (\CPs) is popular \cite{NIPSadd1, NIPSadd3, NIPSadd4} and important in a wide variety of research fields, including genetics \cite{CaronDoucet, NIPSGene1, NIPSGene2}, finance \cite{NIPSFinance}, oceanography \cite{climatologyCPs}, brain imaging and cognition \cite{NIPSMEG, NIPSCognition}, cybersecurity \cite{CybersecurityCPs} and robotics \cite{NIPSMovementSegmentation, NIPSReinforcementLearning}.
%
For streaming data, a particularly important 
subclass are Bayesian On-line Changepoint Detection (\BOCPD) methods that can process data sequentially \citep{BOCD, FearnheadOnlineBCD, ABOCD, MurphyMVTSBCP, HazardLearningBOCD, GPBOCD, CaronDoucet,  CHAMP, TurnerVB, BSCPD2, BOCPDMS} while providing full probabilistic uncertainty quantification.
%
These algorithms declare \CPs if the posterior predictive computed from $\*y_{1:t}$ at time $t$ 
has low density for the  value of the observation $\*y_{t+1}$ at time $t+1$. Naturally, this leads to a high false \CP discovery rate in the presence of outliers and 
as they run on-line, pre-processing is not an option.
In this work, we provide the first robust on-line \CP detection method that is applicable to multivariate data, works with a class of scalable models and quantifies model, \CP and parameter uncertainty in a principled Bayesian fashion.
%
%

\begin{figure}[t!]
\begin{center}
\centerline{\includegraphics[trim= {1.8cm 0.0cm 2.4cm 0.65cm}, clip, 
width=1.00\columnwidth]{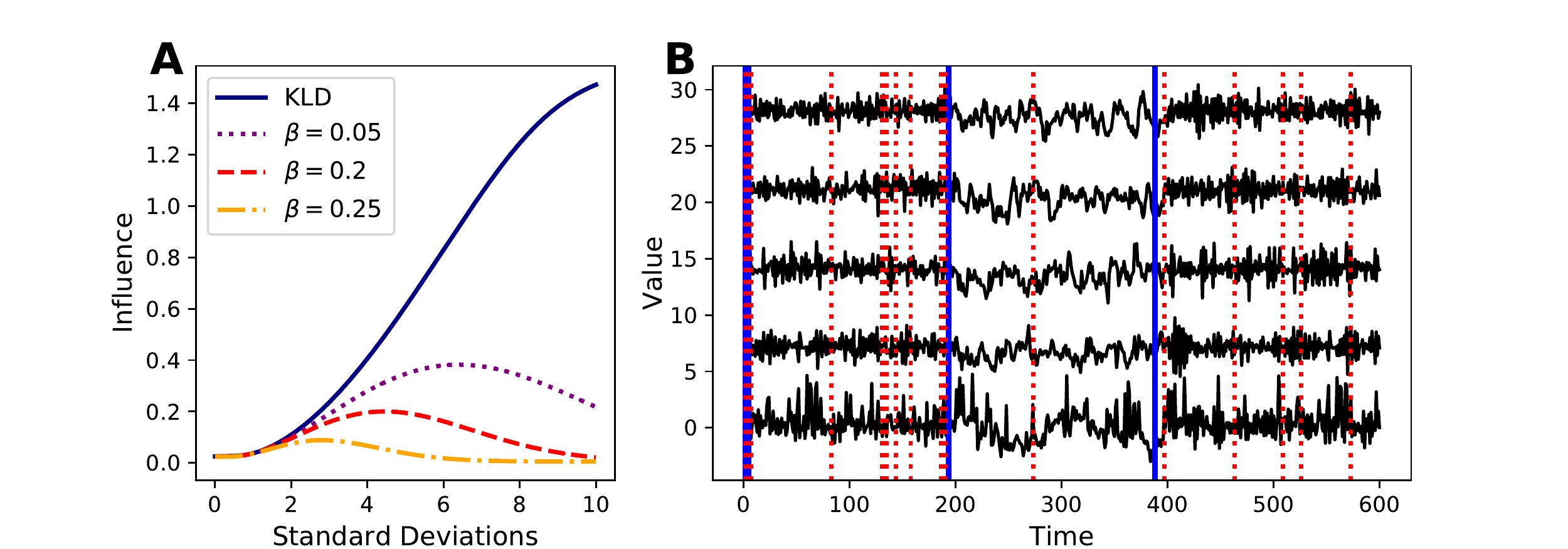}}
\caption{
{\textbf{A:}} 
Influence of $\*y_t$ on inference as function of distance to the posterior expectation in Standard Deviations for $\beta$-divergences with different $\beta$s.
{\textbf{B:}}
Five jointly modeled Simulated Autoregressions (\ARs) with true \CPs at $t=200,400$; bottom-most \AR injected with $t_4$-noise. Maximum A Posteriori \CPs of {\color{blue}robust} ({\color{red}standard}) \BOCPD shown as solid (dashed) vertical lines.}
\label{figure:demo}
\vskip -0.25in
\end{center}
\end{figure}

{Standard Bayesian inference minimizes the Kullback-Leibler divergence (\KLD) between the fitted model and the Data Generating Mechanism (\DGM), but is not robust under outliers or model misspecification due to {its} strictly increasing influence function.
We remedy this by instead minimizing the $\beta$-divergence (\BD) whose influence function has a unique maximum, allowing us to deal with outliers effectively. 
Fig. \ref{figure:demo} \textbf{A} illustrates this: 
Under the \BD, the influence of observations first increases as they move away from the posterior mean, mimicking the \KLD. However, once they move far enough, their influence decreases again. This can be interpreted to mean that they are (increasingly) treated as outliers. As $\beta$ increases, observations are registered as outliers closer to the posterior mean. Conversely, as $\beta \to 0$, one recovers the \KLD which cannot treat any observation as an outlier.
%
In addressing misspecification and outliers this way, our approach builds on the principles of General Bayesian Inference (\GBI) \citep[see][]{Bissiri, Jewson} and robust divergences \cite[e.g.][]{DPDBasuOriginal, DPDBasu}.}
This paper presents three contributions in separate domains that are also illustrated in Figs. \ref{figure:demo} and \ref{figure:contourPlot}:
\begin{itemize}
\item[(1)] \textbf{Robust \BOCPD}: We construct the very first robust \BOCPD inference. The procedure is applicable to a wide class of (multivariate) models and is demonstrated on Bayesian Linear Regression (\BLR). Unlike standard \BOCPD, it discerns outliers and \CPs, see Fig. \ref{figure:demo} \textbf{B}.
\item[(2)] \textbf{Scalable \GBI:} Due to intractable posteriors, \GBI has received little attention in machine learning 
so far. We remedy this with a Structural Variational approximation which
preserves parameter dependence and is exact as $\beta \to 0$, providing a near-perfect fit, see Fig. \ref{figure:contourPlot}. 
\item[(3)] \textbf{Choosing $\beta$:} While Fig. \ref{figure:demo} \textbf{A} shows that $\beta$ regulates the degree of robustness \cite[see also][]{Jewson, DPDBasu}, it is unclear how to set its magnitude. For the first time, we provide a principled way of initializing $\beta$. Further, we show how to refine it on-line by minimizing predictive losses.
\end{itemize}
The remainder of the paper is structured as follows:
In Section \ref{BOCPDAndBD}, we summarize standard \BOCPD and show how to extend it to robust inference using the \BD.
We quantify the degree of robustness and show that inference under the \BD can be designed so that a single outlier never results in false declaration of a \CP, which is impossible under the \KLD.   
Section \ref{GBI} motivates efficient Structural Variational Inference (\SVI) with the \BD posterior. Within \BOCPD, we propose to scale \SVI  
using variance-reduced Stochastic Gradient Descent. 
Next, Section \ref{beta_choice} expands on how $\beta$ can be initialized before the algorithm is run and then optimized on-line during execution time. Lastly, Section \ref{result_section} showcases the substantial gains in performance of robust \BOCPD when compared to its standard version
on real world data in terms of both predictive error and \CP detection.
%
\section{Using Bayesian On-line Changepoint Detection with \texorpdfstring{$\beta$}--Divergences}\label{BOCPDAndBD}
\BOCPD is based on the Product Partition Model \cite{PPM} and introduced independently in \citet{BOCD} and \citet{FearnheadOnlineBCD}. Recently, both formulations have been unified in \citet{BOCPDMS}. The underlying algorithm has extensions ranging from Gaussian Processes \cite{GPBOCD} and on-line hyperparameter optimization \cite{CaronDoucet} to non-exponential families \cite{TurnerVB, CHAMP}.

To formulate \BOCPD probabilistically, define the run-length $r_t$ as the number of observations at time $t$ since the most recent \CP and $m_t$ as the best model in the set $\mathcal{M}$ for the observations since that \CP.  Then, given a real-valued multivariate process $\{  \*y_t \}_{t=1}^{\infty}$ of dimension $d$, a {model universe} $\mathcal{M}$, a {run-length} prior $h$ defined over $\mathbb{N}_0$ and a model prior $q$ over $\mathcal{M}$, the \BOCPD model is 
\begin{IEEEeqnarray}{CCCCCC}
  \IEEEyesnumber\IEEEyessubnumber*
   r_t|r_{t-1} 
   &{\sim} & 
   H(r_t, r_{t-1})
  	\quad\quad & 
  m_t|m_{t-1}, r_t  { }
  &{\sim}& 
  q(m_t|m_{t-1}, r_t)  \label{BOCPD_mrl} \\
 	\*\theta_m|m_t{ } 
 	& \sim & 
 	{  }\pi_{m_t}(\*\theta_{m_t}) 
 	\quad\quad &
 	\*y_t| m_t, \*\theta_{m_t} 
 	& \sim &
 	 f_{m_t}(\*y_t|\*\theta_{m_t})  \label{BOCPD_paramobs}
\end{IEEEeqnarray}
where $q(m_t|m_{t-1}, r_t) = {1}_{m_{t-1}}(m_{t})$  for $r_t>0$ and $q(m_t)$ otherwise, and where $H$ is the conditional run-length prior so that $H(0,r) = h(r+1)$, $H(r+1, r) = 1- h(r+1)$ for any $r \in \mathbb{N}_0$ and $H(r,r') = 0$ otherwise. For example, Bayesian Linear Regression (\BLR) with the $d \times p$ regressor matrix $\*X_t$ and prior covariance $\Sigma_0$ is given by $\*\theta_{m} = (\sigma^2, \*\mu)$, $f_m(\*y_t|\*\theta_{m}) = \mathcal{N}_d(\*y_t; \*X_t\*\mu, I_d )$ and
$\pi_m(\*\theta_{m}) = \mathcal{N}_d(\*\mu; \*\mu_0, \sigma^2 \Sigma_0)\mathcal{IG}(\sigma^2; a_0, b_0)$.
%
If the computations of the parameter posterior
$\pi_{m}(\*\theta_{m}|\*y_{1:t}, r_t)$ 
and the posterior predictive 
$f_m(\*y_t|\*y_{1:(t-1)}, r_t) =  \int_{\Theta_m}f_{m}(\*y_t|\*\theta_{m})\pi_{m}(\*\theta_{m}|\*y_{1:(t-1)}, r_t)d\*\theta_m$
are efficient for all models $m \in \mathcal{M}$, then so is the recursive computation given by
\begin{IEEEeqnarray}{rCl}
 \IEEEyesnumber\IEEEyessubnumber*
	p(\*y_{1}, r_1=0, m_1)
	 & = &  q(m_1)\cdot
     	\int_{\Theta_{m_1}}f_{m_1}(\*y_1|\*\theta_{m_1})\pi_{m_1}(\*\theta_{m_1})d\*\theta_{m_1} 
        = q(m_1)\cdot f_{m_1}(\*y_{1}|\*y_0), \label{BOCPD_recursion1} \\ 
	p(\*y_{1:t}, r_t, m_t)
	& = &  \hspace*{-0.4cm}
	 \sum_{{m_{t-1}, {r_{t-1}}}}\hspace*{-0.4cm}	  \Bigl\{ 
	 	f_{m_t}(\*y_t|\mathcal{F}_{t-1} ) 
	 	q(m_t|\mathcal{F}_{t-1} , m_{t-1}) 
	 	H(r_{t},r_{t-1}) 
	 	p(\*y_{1:(t-1)}, r_{t-1}, m_{t-1})
	 	 \Bigr\}  \label{BOCPD_recursion2} \quad\:\:\:
\end{IEEEeqnarray}
where $\mathcal{F}_{t-1} = \left\{ \*y_{1:(t-1)}, r_{t-1} \right\}$ and $p(\*y_{1:t}, r_t, m_t)$ is the  joint density of $\*y_{1:t}$, $m_t$ and $r_t$. 
The run-length and model posteriors are then available exactly at time $t$, as $p(r_t, m_t|\*y_{1:t}) = {p(\*y_{1:t}, r_t, m_t)}/{\sum_{m_t,r_t}p(\*y_{1:t}, r_t, m_t)}$.
For a full derivation and the resulting inference see \cite{BOCPDMS}. 


\subsection{General Bayesian Inference (\GBI) with $\beta$-Divergences (\BD)}\label{SubSec:GBIwithBD}

Standard Bayesian inference minimizes the \KLD between the  Data Generating Mechanism (\DGM) and its probabilistic model (see Section 2.1 of \cite{Bissiri} for a clear illustration). 
In the M-closed world where one assumes that {the} \DGM and model coincide, the \KLD is the most efficient way of
updating posterior beliefs. However, this is no longer the case in the M-open world \cite{bernardo2001bayesian} where they match only approximately \cite{Jewson}, e.g.\ in the presence of outliers. 
\GBI \cite{Bissiri, Jewson} generalizes standard Bayesian updating based on the \KLD to a family of divergences. 
In particular, it uses the relationship between losses $\ell$ and divergences $D$ to deduce for $D$ a corresponding loss $\ell^D$. It can then be shown that for model $m$, the posterior update optimal for $D$ yields the distribution 
\begin{IEEEeqnarray}{rCl}
	\pi^{D}_m(\*\theta_m|\*y_{(t-r_t):t}) 
	& \propto & 
	\pi_m(\theta)\exp\left\{ 
		-\textstyle{\sum_{	i=t-r_t}^t} \ell^D(\*\theta_m|\*y_i)
	\right\}. \label{GBI_posterior}
\end{IEEEeqnarray}
For parameter inference with the \KLD and \BD, these losses are the log score and the Tsallis score:
\begin{IEEEeqnarray}{rCl}
	\ell^{\KLD} (\*\theta_m|\*y_t) &=& -\log\left(f_{m}(\*y_t|\*\theta_m
	\right) 
		\label{KLD_loss}\\
	\ell^{\beta}(\*\theta_m|\*y_t) &=& -\left(
			\frac{1}{\bpm}f_{m}(\*y_t|\*\theta_m
			 )^{\bpm} 
			- \frac{1}{1+\bpm}\int_{\mathcal{Y}}f_{m}(\*z|\*\theta_m 
			)^{1+	\bpm}d\*z
			\right).
		\label{BD_loss}
\end{IEEEeqnarray}
Eq. \eqref{BD_loss} shows why the \BD excels at robust inference: Similar to tempering, $\ell^{\beta}$ exponentially downweights the density, attaching less influence to observations in the tails {of the model}. 
This phenomenon is depicted with influence functions $I(\*y_t)$ in Figure \ref{figure:demo} \textbf{A}.
$I(\*y_t)$ is a divergence between the posterior with and without an observation $\*y_t$ \cite{kurtek2015bayesian}. 
%

\GBI with the \BD yields robust inference without the need to specify a heavy-tailed or otherwise robustified model. Hence, one estimates the same model parameters as in standard Bayesian inference while down-weighting the influence of observations that are overly inconsistent with the model. 
Accordingly, \GBI provides robust inference for a much wider class of models and situations than the ones illustrated here.
%
Though other divergences such as $\alpha$-Divergences \cite[e.g.][]{alphaBB} also accommodate robust inference, we restrict ourselves to the \BD. We do this because unlike other divergences, it does not require estimation of the \DGM's density. Density estimation increases estimation error, is computationally cumbersome and works poorly for small run-lengths (i.e.\ sample sizes). 
%
Note that versions of \GBI have been proposed before \cite{VBFormulationGBI,RenyiDiv,OperatorDiv,XiDiv},  but have framed the procedure as alternative to Variational Bayes instead.

{Apart from the computational gains of {S}ection \ref{SVI}}, we tackle robust inference via the \BD rather than via Student's $t$ errors for three reasons:
%
Firstly, robust run-length posteriors need robustness in \textit{ratios} rather than \textit{tails} (see {S}ection \ref{QuantRobustness} and the simulation results for Student's $t$ errors in the {A}ppendix).
Secondly, Student's $t$ errors model outliers as part of the \DGM, which compromises the inference target:
Consider a \BLR with error $e_t = \varepsilon_t + w_t\nu_t$, where $w_t \sim \text{Ber}(p)$ for $p=0.01$, $\varepsilon_t\sim \mathcal{N}(0,\sigma^2)$ with outliers $\nu_t \sim t_{1}(0, \gamma)$.
Appropriate choices of $\bpm$ give most influence to the $(1-p) \cdot 100\% = 99\%$ of typical observations one can explain well with the \BLR model.
In contrast, modeling $e_t$ as Student's $t$ under the \KLD lets $\nu_t$ dominate parameter inference and lets 1\% of observations inflate the predictive variance substantially.  
%
Thirdly, using Student's $t$ errors is a technique only applicable to symmetric, continuous models. In contrast, \GBI with the \BD is valid for any setting, e.g. for asymmetric errors as well as point and count processes.


%

\subsection{Robust \BOCPD}

The literature on robust on-line \CP detection so far is sparse and covers limited settings without Bayesian uncertainty quantification
\cite[e.g.][]{robust2, robust3, FearnheadOutliersConstant}. 
For example, the method in \citet{FearnheadOutliersConstant} only produces point estimates 
and is limited to fitting a piecewise constant function to univariate data. In contrast, \BOCPD can be applied to multivariate data and a set of models $\mathcal{M}$ while quantifying uncertainty about these models, their parameters and potential \CPs, but is not robust. 
%
%
Noting that for standard \BOCPD the posterior expectation is given by
\begin{IEEEeqnarray}{rCl}
 \mathbb{E}\left(\*y_t|\*y_{1:(t-1)}\right) & = &
 	\sum_{r_t,m_t}\mathbb{E}\left(\*y_t|\*y_{1:(t-1)}, r_{t-1}, m_{t-1}\right)p(r_{t-1}, m_{t-1}|\*y_{1:(t-1)}), \label{Posterior_expectation}
\end{IEEEeqnarray}
%
%
the key observation is that prediction is driven by two probability distributions: The run-length and model posterior $p(r_t, m_t|\*y_{1:t})$ and parameter posterior distributions 
$\pi_m(\*\theta_m|\*y_{1:t})$. Thus, we make \BOCPD robust by using \BD posteriors $p^{\brlmm}
(r_t, m_t|\*y_{1:t})$, $\pi_m^{\bpm}(\*\theta_m|\*y_{1:t})$
for $\*\beta = (\brlmm, \bpm) > 0$\footnote{In fact, \bp = $\beta_p^m$, i.e.\ the robustness is model-specific, but this is suppressed for readability}. 

$\brlmm$ prevents abrupt changes in $p^{\brlmm}(r_t, m_t|\*y_{1:t})$ caused by a small number of observations, see section \ref{QuantRobustness}. This form of robustness is easy to implement and retains the closed forms of \BOCPD: In Eqs. \eqref{BOCPD_recursion1} and \eqref{BOCPD_recursion2}, one simply replaces $f_{m_t}(\*y_t|\*y_0 )$ and $f_{m_t}(\*y_t|\mathcal{F}_{t-1} )$ by their \BD-counterparts $\exp\{ \ell^{\brlmm}(\*\theta_{m_t}|\*y_t) \}$, where
\begin{IEEEeqnarray}{rCl}
 \ell^{\brlmm}(\*\theta_{m_t}|\*y_t) & = &  
	-\left(
			\frac{1}{\brlmm}f_{m}(\*y_t|\mathcal{F}_{t-1}
			 )^{\brlmm} 
			- \frac{1}{1+\brlmm}\int_{\mathcal{Y}}f_{m}(\*z| \mathcal{F}_{t-1} 
			)^{1+	\brlmm}d\*z
			\right). \label{eq:rlm_robustness}
\end{IEEEeqnarray}
While {the posterior $p^{\brlmm}(r_t, m_t|\*y_{1:t})$ is only available up to a constant, it is discrete and thus easy to normalize.} 
%
Complementing this, $\bpm$ regulates the robustness of $\pi_m^{\bpm}(\*\theta|\*y_{1:t})$ by preventing it from being dominated by tail events.
Section \ref{SVI} overcomes the intractability of $\pi_m^{\bpm}(\*\theta|\*y_{1:t})$ 
using Structural Variational Inference (\SVI) that recovers the approximated distribution exactly as $\bpm \to 0$.
%
%
\begin{figure}[t!]
\vskip -0.08in
\begin{center}
{\includegraphics[width=1\columnwidth, 
trim= {0.4cm 0.4cm 0.5cm 0.15cm}, clip]{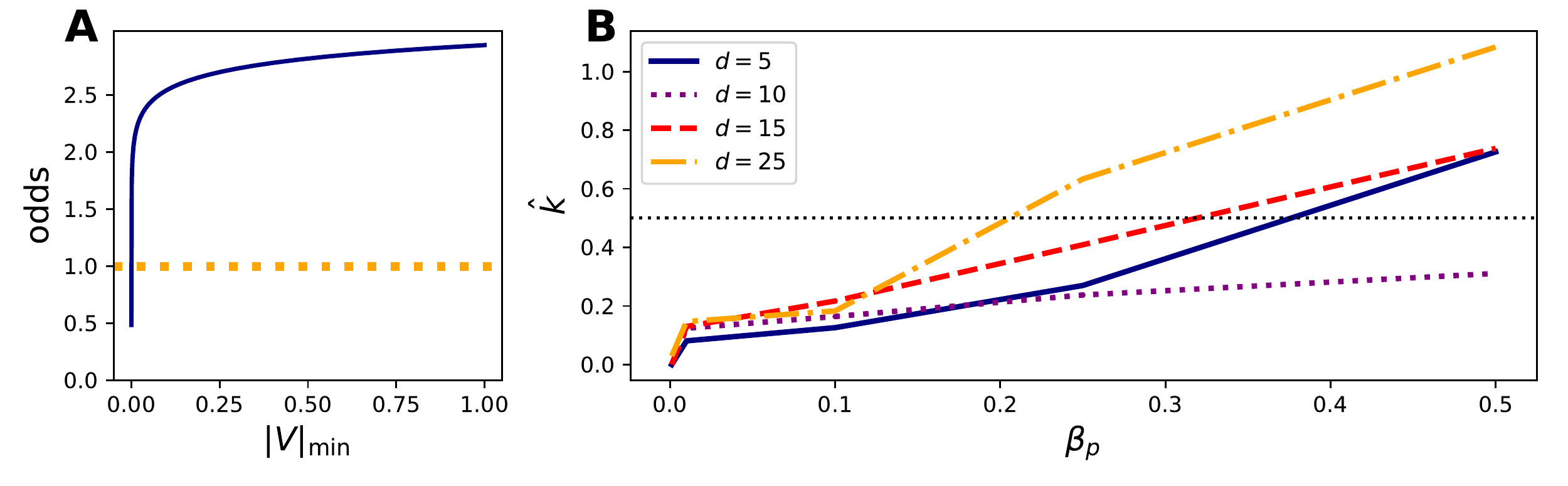}}
\caption{\textbf{A}: {\color{navy}Lower bound} on the odds of Thm. \ref{Thm:DeclareChange}  for priors used for Figure \ref{figure:demo} \textbf{B} and $h(r) = 1/100$.
\textbf{B}: $\hat{k}$ for different choices of $\bpm$ and output (input) dimensions $d$ ($2d$) in an autoregressive \BLR 
}  
\label{figure:THM_VB}
\end{center}
\vskip -0.2in
\end{figure}
\subsection{Quantifying robustness}\label{QuantRobustness}
{The algorithm of \citet{FearnheadOutliersConstant} is robust because hyperparameters enforce that a single outlier is insufficient for declaring a \CP. 
Analogously, we 
investigate conditions under which a {single} (outlying) observation $\*y_{t+1}$ is able to force a \CP. An intuitive way of achieving this is by studying the odds of $r_{t+1} \in \{0,r+1\}$ conditional on $r_t=r$:
\begin{IEEEeqnarray}{lCr}
	\frac{p(r_{t+1}=r+1|\*y_{1:t+1},r_t=r,m_t)}
		{p(r_{t+1}=0|\*y_{1:t+1},r_t=r,m_t)}
&=&\frac{\cancel{p(\*y_{1:t}, r_t=r, m_t)}\cdot(1-H(r_{t+1}, r_t))f_{m_t}^D(\*y_{t+1}|\mathcal{F}_{t} )}
		{\cancel{p(\*y_{1:t}, r_t=r, m_t)}\cdot H(r_{t+1}, r_t)f_{m_t}^D(\*y_{t+1}|\*y_0)}. \quad \label{Odds}
\end{IEEEeqnarray}
%
{Here, $f_{m_t}^D$ denotes the negative exponential of the score under divergence $D$. In particular, $f_{m_t}^{\KLD}(\*y_{t+1}|\mathcal{F}_{t} ) = f_{m_t}(\*y_{t+1}|\mathcal{F}_{t} )$ and $f_{m_t}^{\brlmm}(\*y_{t+1}|\mathcal{F}_{t} ) = \exp\left\{-\ell^{\brlmm}(\*\theta_m|\*y_t)\right\}$ as in Eq. \eqref{eq:rlm_robustness}. 
Taking a closer look at Eq. \eqref{Odds}, if $\*y_{t+1}$ is an outlier with low density under $f_{m_t}^D(\*y_{t+1}|\mathcal{F}_{t} )$, the odds will move in favor of a \CP provided that the prior is sufficiently uninformative to make $f_{m_t}^D(\*y_{t+1}|\*y_0) > f_{m_t}^D(\*y_{t+1}|\mathcal{F}_{t} )$.
In fact, even very small differences have a substantial impact on the odds.
This is why using the Student's $t$ error for the \BLR model with standard Bayes will not provide robust run-length posteriors: 
While an outlying observation $\*y_{t+1}$ will have greater density $f_{m_t}^{\KLD}(\*y_{t+1}|\mathcal{F}_{t})$ under a Student's $t$ error model than under a normal error model, $f_{m_t}^{\KLD}(\*y_{t+1}|\*y_0)$ (the density under the prior) will also be larger under the Student's $t$ error model. As a result, changing the tails of the model only has a very limited effect on the ratio in Eq. (\ref{Odds}).
In fact, the perhaps unintuitive consequence is that Student's $t$ error models will yield \CP inference that very closely resembles that of the corresponding normal model. A range of numerical examples in the Appendix illustrate this surprising fact.
In contrast, \CP inference robustified via the \BD does not suffer from this phenomenon. In fact, 
Theorem \ref{Thm:DeclareChange} provides very mild  conditions for the \BD robustified \BLR model ensuring that the odds never favor a \CP after \textit{any} single outlying observation $\*y_{t+1}$. 
\begin{theorem}\label{Thm:DeclareChange}
If $m_t$ in Eq. \eqref{Odds} is the Bayesian Linear Regression (\BLR) model with $\*\mu \in \mathbb{R}^p$ 
and priors $a_0$, $b_0$, $\mu_0$, $\Sigma_0$; and if the posterior predictive's variance determinant is larger than $\left|V\right|_{\min}>0$, then one can choose any $(\brlmm, H(r_t,r_{t+1})) \in S\left(p,\brlmm,a_0,b_0,\mu_0,\Sigma_0, \left|V\right|_{\min}\right)$ to guarantee that
\begin{IEEEeqnarray}{rCl}
 \frac{(1-H(r_{t+1}, r_t))f_{m_t}^{\brlmm}(\*y_{t+1}|\mathcal{F}_{t} )}
		{H(r_{t+1}, r_t)f_{m_t}^{\brlmm}(\*y_{t+1}|\*y_0)} & \geq & 1,
\end{IEEEeqnarray}
where the set $S\left(p,\brlmm,a_0,b_0,\mu_0,\Sigma_0, \left|V\right|_{\min}\right)$ is defined by an inequality given in the Appendix.
\end{theorem}
{
{Thm. \ref{Thm:DeclareChange} says that one can bound the odds for a \CP independently of $\*y_{t+1}$. 
The requirement for a lower bound $\left|V\right|_{\min}$ results from the integral term in Eq. \eqref{BD_loss}, 
 which dominates \BD-inference if $\left|V\right|$ is extremely small. In practice, this is not restrictive:} 
{E.g.\  for $p=5$, $h(r)=\frac{1}{\lambda}$,  $a_0=3,b_0=5,\Sigma_0=\textrm{diag}(100,5)$ used in Fig. \ref{figure:demo} \textbf{B}, Thm. \ref{Thm:DeclareChange} holds for $(\brlmm,\lambda)=(0.15,100)$ used for inference if  $\left|V\right|_{\min}\geq 8.12 \times 10^{-6}$.
Fig. \ref{figure:THM_VB} \textbf{A} plots the lower bound (see Appendix) as function of  $\left|V\right|_{\min}$. 
}

\begin{figure}[ht!]
\begin{center}
{\includegraphics[width=0.325\columnwidth, 
trim= {0.3cm 0.3cm 0.3cm 0.5cm}, clip]{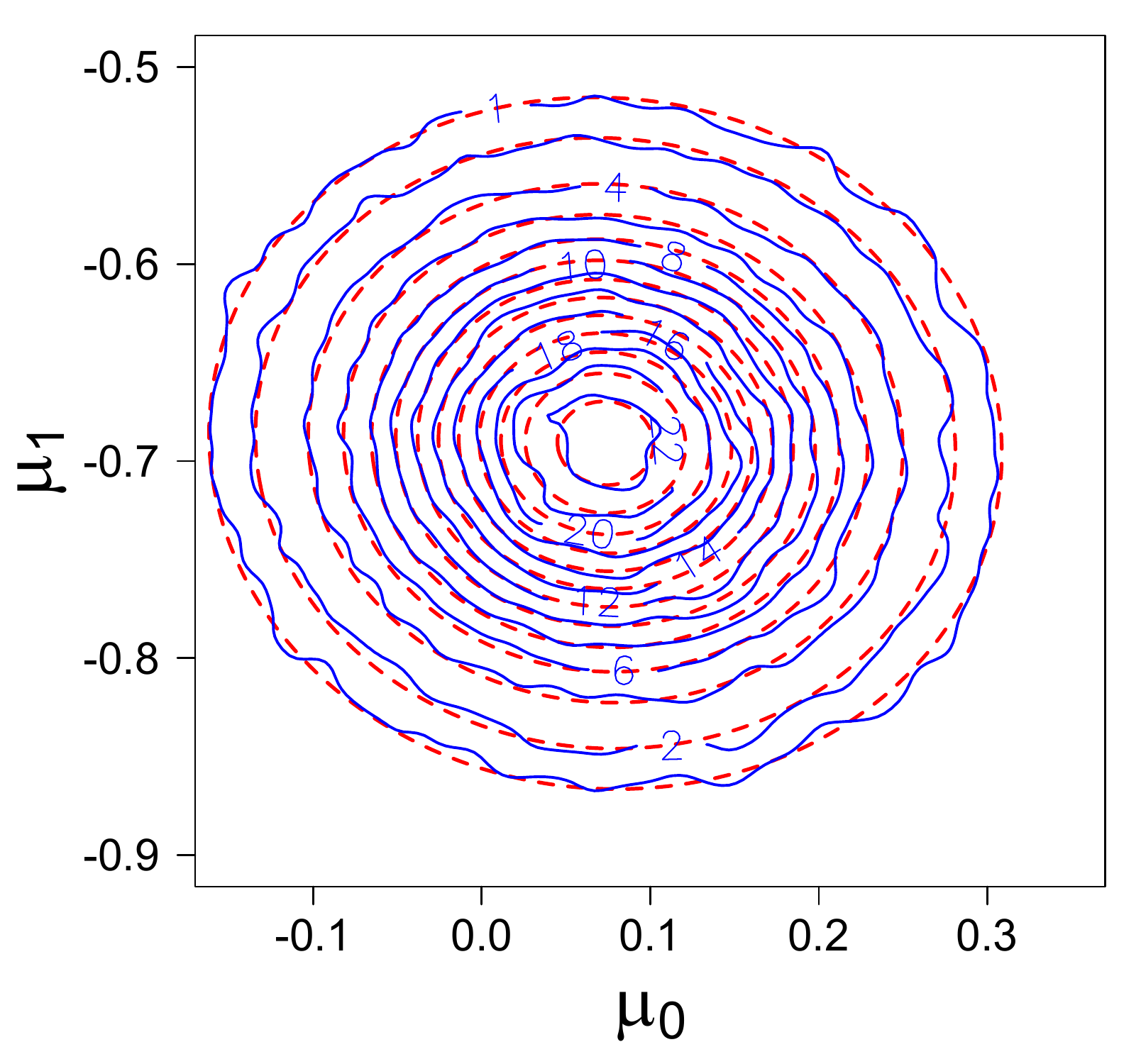}}
{\includegraphics[width=0.325\columnwidth,
trim= {0.25cm 0.3cm 0.3cm 0.5cm}, clip]{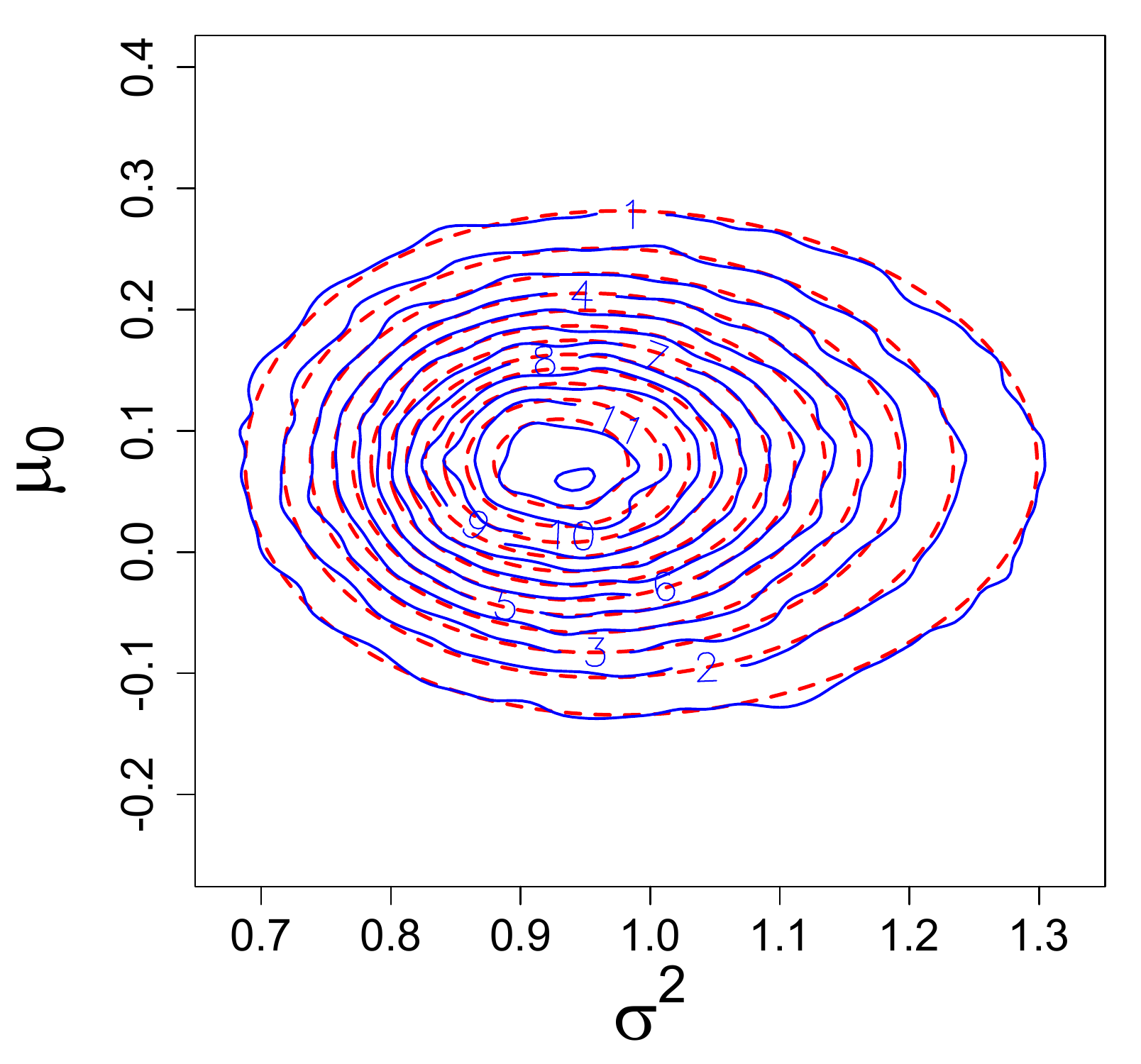}}
{\includegraphics[width=0.325\columnwidth,
trim= {0.25cm 0.3cm 0.5cm 0.5cm}, clip]{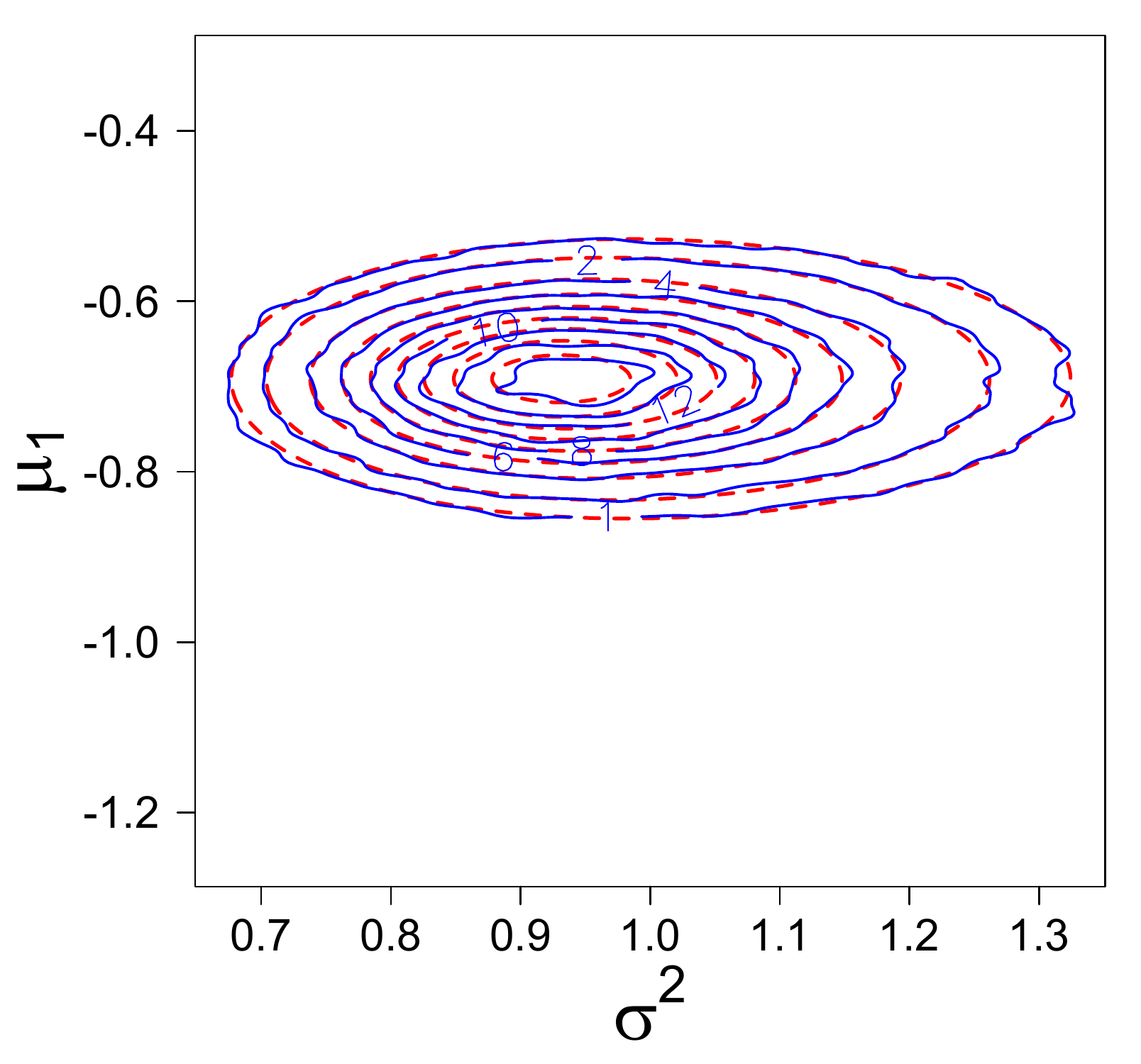}}
\caption{Exemplary contour plots of bivariate marginals for the approximation $\widehat{\pi}^{\bpm}_m(\*\theta_m)$  of Eq. \eqref{KL_equation} ({\color{red} dashed}) and the target $\pi^{\bpm}_m(\*\theta_m|\*y_{(t-r_t):t})$ ({\color{blue} solid}) estimated and smoothed from $95,000$ Hamiltonian Monte Carlo samples for the \BD posterior of \BLR with $d=1$, two regressors and $\bpm = 0.25$.
}  
\label{figure:contourPlot}
\end{center}
\vskip -0.2in
\end{figure}

\section{On-line General Bayesian Inference (\GBI)}\label{GBI}

\subsection{Structural Variational Approximations for Conjugate Exponential Families}\label{SVI}
{While there has been a recent surge in theoretical work on \GBI  \cite{Bissiri,DPDBasu,Jewson,VBFormulationGBI}, applications have been sparse, in large {part} due to intractability. While sampling methods have been used successfully for \GBI \cite{Jewson,DPDBasu}, it is not easy to scale {these} for the robust \BOCPD setting.
%
%
%
Thus, most work on \BOCPD has focused on conjugate distributions \cite{BOCD,ABOCD,FearnheadOnlineBCD} and approximations \cite{TurnerVB,CHAMP}. We extend the latter branch of research by deploying 
Structural Variational Inference (\SVI). Unlike mean-field approximations, this preserves parameter dependence in the posterior, see Figure \ref{figure:contourPlot}. While it is in principle possible to solve the inference task by sampling, this is computationally burdensome and makes the algorithm on-line in name only:
%
%
Any sampling approach needs to (I) sample from $\pi_m^{\beta_{p}}(\mathbf{\theta}_m|\*y_{t-r_t:t})$ in Eq. \eqref{GBI_posterior}, (II) numerically integrate to obtain $f_m(\*y_t|\*y_{1:(t-1)}, r_t)$ and lastly (III) sample and numerically integrate the integral in Eq. \eqref{eq:rlm_robustness} which {no longer has} a closed form. Moreover, this has to be performed for each $(r_t, m)$ at times $t=1,2,\dots$. On top of this increased computational cost, it  creates three sources of approximation error propagated forward through time via Eqs. \eqref{BOCPD_recursion1} and \eqref{BOCPD_recursion2}.
%
%
%
Since $\pi_m^{\KLD}$ is available in closed form and as \BD $\to$ \KLD as $\beta \to 0$ \cite{DPDBasuOriginal}, there is an especially compelling way of doing \SVI for conjugate models using the \BD based on the fact that 
\begin{IEEEeqnarray}{rCl}
	\pi^{\bpm}_m(\*\theta_m|\*y_{(t-r_t):t})  
    & \approx & 
    \pi^{\KLD}_m(\*\theta_m|\*y_{(t-r_t):t})
    \label{SVI_inference_idea}
\end{IEEEeqnarray}
is exact as $\beta \to 0$. Thus we approximate the \BD posterior for model $m$ and run-length $r_t$ as 
\begin{IEEEeqnarray}{rCl}
	\widehat{\pi}^{\bpm}_m(\*\theta_m) & = &
    \argmin_{\pi^{\KLD}_m(\*\theta_m)}
		\left\{ 
			\text{KL}\left(  \pi^{\KLD}_m(\*\theta_m)
            		\left\| \pi^{\bpm}_m(\*\theta_m|\*y_{(t-r_t):t}) 
                    \right)	\right.
		\right\}. \label{KL_equation}
\end{IEEEeqnarray}
While this ensures that the densities $\widehat{\pi}^{\bpm}_m$ and $\pi^{\KLD}_m$ belong to the same family, the variational parameters can be very different from those  implied by the \KLD-posterior. This approximation mitigates multiple issues that would arise with sampling approaches: By forcing $\pi_m^{\beta_{p}}(\mathbf{\theta}_m|\*y_{1:t})$ into the conjugate closed form, steps (II) and (III) are solved analytically. Thus, inference is orders of magnitude faster, while the resulting approximation error remains negligible (see Figs 2B, 3).

Moreover, for many models, the Evidence Lower Bound (\ELBO) associated with the optimization in Eq. \eqref{KL_equation} is available in closed form. As a result, off-the-shelf optimizers are sufficient and no black-box or sampling-based techniques are required to efficiently tackle the problem.
Theorem \ref{Thm:QuasiConjugacyJK} provides the conditions for a conjugate exponential family to admit such a closed form \ELBO. The proof alongside the derivation of the \ELBO for \BLR can be found in the Appendix
%


}

\begin{theorem}\label{Thm:QuasiConjugacyJK}
The \ELBO objective corresponding to the \BD posterior approximation in Eq. \eqref{KL_equation} 
of an exponential family likelihood model 
$f_m(\*y;\theta_m)=\exp\left(\eta(\theta_m)^TT(\*y)\right)g(\eta(\theta_m))A(x)$ 
with conjugate prior 
$\pi_0(\theta_m|\nu_0,\mathcal{X}_0)=g(\eta(\theta_m))^{\nu_0}\exp\left(\nu_0\eta(\theta_m)^T\mathcal{X}_0\right)h(\mathcal{X}_0,\nu_0)$ 
and variational posterior $\widehat{\pi}^{\bpm}_m(\theta_m|\nu_m,\mathcal{X}_m)=g(\eta(\theta_m))^{\nu_m}\exp\left(\nu_m\eta(\theta_m)^T\mathcal{X}_m\right)h(\mathcal{X}_m,\nu_m)$ within the same conjugate family is analytically available iff the following three quantities have closed form:
\begin{IEEEeqnarray}{rCl}
	\mathbb{E}_{\widehat{\pi}^{\bpm}_m}\left[\eta(\theta_m)\right], \;
    \mathbb{E}_{\widehat{\pi}^{\bpm}_m}\left[\log g(\eta(\theta_m))\right], \;
    \bigintssss \hspace*{-0.15cm}{A(z)^{1+\bpm}}\left[{h\left(\frac{(1+\bpm) T(z)+\nu_m\mathcal{X}_m}{1+\bpm+\nu_m},1+\beta+\nu_m\right)}\right]^{-1}\hspace*{-0.3cm}dz. \nonumber 
\end{IEEEeqnarray}
\end{theorem}
%
%
The conditions of  Theorem \ref{Thm:QuasiConjugacyJK} are met by many exponential models, e.g.\ the Normal-Inverse-Gamma, the Exponential-Gamma, and the Gamma-Gamma.
%
%
For a simulated autoregressive \BLR, we assess the quality of $\widehat{\pi}^{\bpm}$ following \citet{yao2018yes}, who estimate a difference $\hat{k}$ between ${\pi}^{\bpm}_m$ and $\widehat{\pi}^{\bpm}_m$ relative to a posterior expectation. We use this on the posterior predictive, which is 
an expectation relative to $\pi^{\bpm}_m$ and drives the \CP detection. 
\citet{yao2018yes} rate $\widehat{\pi}^{\bpm}_m$ as \textit{close} to ${\pi}^{\bpm}_m$ 
if $\hat{k}<0.5$. 
Figs \ref{figure:contourPlot} and \ref{figure:THM_VB} \textbf{B} show that our approximation lies well below this threshold for choices of $\bpm$ decreasing reasonably fast with the dimension. Note that these are exactly the values of $\bpm$ one will want to select for inference:
{As $d$ increases, the magnitude of $f_{m_t}(\*y_t|\mathcal{F}_{t-1} )$ decreases rapidly. Hence, $\beta_{\text{p}}$ needs to decrease as $d$ increases to prevent the \BD inference from being dominated by the integral in Eq. \eqref{BD_loss} and disregarding $\*y_t$ \cite{Jewson}.} This is also reflected in our experiments in section \ref{result_section}, for which we initialize $\bpm = 0.05$ and $\bpm = 0.005$ for $d=1$ and $d=29$, respectively. However, as Figs. \ref{figure:contourPlot} and \ref{figure:THM_VB} \textbf{B} illustrate, the approximation is still excellent for values of $\bpm$ that are much larger than that.

\subsection{Stochastic Variance Reduced Gradient (\SVRG) for \BOCPD}

While highest predictive accuracy within \BOCPD is achieved using full optimization of the variational parameters at each of $T$ time periods, this has space and time complexity of $\mathcal{O}(T)$ and $\mathcal{O}(T^2)$. 
%
In comparison, Stochastic Gradient Descent (\SGD) has space and time complexity of $\mathcal{O}(1)$ and $\mathcal{O}(T)$, but yields a loss in accuracy, substantially so for small run-lengths. 
In the \BOCPD setting, there is an obvious trade-off between accuracy and scalability:
Since the posterior predictive distributions $f_{m_t}(\*y_t|\*y_{1:(t-1)}, r_{t})$ for all run-lengths $r_t$ drive \CP detection, \SGD estimates are insufficiently accurate for small run-lengths $r_t$. On the other hand, once $r_t$ is sufficiently large, 
the variational parameter estimates only need minor adjustments
and computing an optimum is costly.

\begin{algorithm}[h!]
	\caption*{\textbf{Stochastic Variance Reduced Gradient (\SVRG) inference for \BOCPD}}
   \label{Algorithm_BOCPDMS}
\begin{algorithmic}
   \State {\bfseries Input at time $0$:} Window \& batch sizes $W$, $B^{\ast}$, $b^{\ast}$; frequency $m$, prior $\*\theta_0$, \#steps  $K$, step size $\eta$
   \State \hspace*{2.4cm}  s.t. $W > B^{\ast} > b^{\ast}$; and $\sim$ denotes sampling without replacement 
   \For{ next observation $\*y_t$ at time $t$}   
   \For{retained run-lengths $r \in R(t)$ }
   	\If{$\tau_r = 0$} 
   		\If{$r < W$}
   			\State $\*\theta_r \leftarrow \*\theta^{\ast}_r \leftarrow \text{FullOpt}\left(\text{\ELBO}(\*y_{t-r:t})\right)$; 
            $\tau_r \leftarrow  m$
   		\ElsIf{ $r \geq W$}
   			\State $\*\theta^{\ast}_r \leftarrow \*\theta_r$;
            $\tau_r \leftarrow  \text{Geom}\left(B^{\ast}/(B^{\ast}+b^{\ast})\right)$
   		\EndIf 
        \State $B \leftarrow \min(B^{\ast}, r)$
   		\State $g_r^{\text{anchor}}  \leftarrow \frac{1}{B}\sum_{i \in \mathcal{I}} \nabla \text{\ELBO}(\*\theta^{\ast}_r, \*y_{t-i})$, 
   			where $\mathcal{I} \sim \text{Unif}\{0,\dots,\min(r,W)\}$,
   			 $|\mathcal{I}| = B$
    \EndIf
    \For{ $j = 1,2,\dots, K$}
    	  		\State $b \leftarrow \min(b^{\ast}, r)$ and $\widetilde{\mathcal{I}} \sim  \text{Unif}\{0,\dots,\min(r,W)\}$ and $|\widetilde{\mathcal{I}}| = b$
    	  		\State $g_r^{\text{old}} \leftarrow  \frac{1}{b}\sum_{i \in \widetilde{\mathcal{I}}} \nabla \text{\ELBO}(\*\theta^{\ast}_r, \*y_{t-i})$, \:
    	  		$g_r^{\text{new}} \leftarrow  \frac{1}{b}\sum_{i \in \widetilde{\mathcal{I}}} \nabla \text{\ELBO}(\*\theta_r, \*y_{t-i})$
    	  		\State $\*\theta_r \leftarrow \*\theta_r + \eta\cdot\left(g_r^{\text{new}} - 
    	  						g_r^{\text{old}} + g_r^{\text{anchor}}\right) $; 
                $\tau_r \leftarrow  \tau_r -1$
    \EndFor
   \EndFor
     \State $r \leftarrow r+1$ for all $r \in R(t)$; $R(t) \leftarrow R(t) \cup \{0\}$
   \EndFor
\end{algorithmic}
\end{algorithm}
Recently, a new generation of algorithms interpolating \SGD and global optimization have addressed this trade-off. They achieve substantially better convergence rates by anchoring the stochastic gradient to a point near an optimum \cite{SVRG, SAGA, ProposedBatches, PracticalSVRG, SCSG}. 
We propose a memory-efficient two-stage variation of these methods tailored to \BOCPD. First, the variational parameters are moved close to their global optimum using a variant of \citep{SVRG,ProposedBatches}. 
Unlike standard versions, we anchor the gradient estimates to a (local) optimum by calling a convex optimizer FullOpt every $m$ steps for the first $W$ iterations. While our implementation uses Python scipy's L-BFSG-B optimization routine, any convex optimizer could be used for this step. Compared to standard \SGD or \SVRG, full optimization substantially decreases variance and increases accuracy for small $r_t$.
%
Second, once $r_t>W$ we do not perform full optimization anymore. Instead, we anchor optimization to the current value as in standard \SVRG, by updating the anchor at stochastic time intervals determined by a geometric random variable with success probability $B^{\ast}/(B^{\ast} + b^{\ast})$. Whether the anchor is based on global optimization or not, the next step consists in sampling $B = \min(r_t, B^{\ast})$ observations without replacement from a window with the $\min(r_t, W)$ most recent observations to initiate the \SVRG procedure. Following this, for the next $K$ observations, we incrementally refine the estimates while keeping their variance low using a stochastic-batch variant of \cite{SCSG, SCSG2} by sampling a batch of size $b = \min(r_t, b^{\ast})$ without replacement from the $\min(r_t, W)$ most recent observations. The resulting on-line inference has constant space and linear time complexity like \SGD, but produces good estimates for small $r_t$ and converges faster \cite{SVRG, SCSG, SCSG2}. 
We provide a detailed complexity analysis of the procedure in the Appendix, where we also demonstrate numerically that it is orders of magnitude faster than \MCMC-based inference.
%





\section{Choice of $\beta$}\label{beta_choice}

\textbf{Initializing} $\bpm$: The \BD has been used in a variety of settings \cite{DPDBasu, DPDBasuOriginal,VBFormulationGBI,DPDForTensorFactorization}, but there is no principled framework for selecting $\beta$. We remedy this by minimizing the expected predictive loss with respect to $\beta$ on-line.
{As the losses need not be convex in $\bpm$, initial values can matter for the optimization. A priori, we pick $\bpm$ maximizing the \BD influence for a given Mahalanobis Distance (\MD) $\*x^{\ast}$ under $\pi(\theta_m)$. As Figure \ref{figure:demo} \textbf{A} shows, $\bpm>0$ induces a point of maximum influence ${\MD}( \bpm, \pi_m(\theta_m))$: Points further in the tails are treated as outliers, while points closer to the mode receive similar influence as under the \KLD.
A Monte Carlo estimate of ${\MD}( \bpm, \pi_m(\theta_m))$ is found via $\widehat{\MD}( \bpm, \pi_m(\theta_m))=\argmax_{{x}\in\mathbb{R}_{+}} \hat{I}( \bpm, \pi_m(\theta_m))({x})$ \cite{kurtek2015bayesian}.
%
We initialize $\bpm$ by solving the inverse problem: For $x^{\ast}$, we seek $\bpm$ such that $\widehat{\MD}( \bpm, \pi_m(\theta_m)) = x^{\ast}$. (The Appendix contains a pictorial illustration of this procedure.)
The $k$-th standard deviation under the prior is a good choice of $x^{\ast}$ for low dimensions \cite[see also][]{FearnheadOutliersConstant}, 
but not appropriate as delimiter for high density regions even in moderate dimensions $d$. Thus, we propose $x^{\ast} = \sqrt{d}$ for larger values of $d$, inspired by the fact that under normality, 
$\text{\MD}\to\sqrt{d}$ as $d \to \infty$ \cite{hall2005geometric}. 
One then finds $\bpm$ by approximating the gradient of $\widehat{\MD}( \bpm, \pi_m(\theta_m))$ with respect to $\bpm$. 
{As $\brlmm$ does not affect $\pi^{\bpm}_m$, its initialization matters less and generally, initializing  $\beta_{\text{rlm}} \in [0,1]$  produces reasonable results.}

\textbf{Optimizing $\*\beta$ on-line}: For $\*\beta = (\brlmm, \bpm)$ and prediction $\widehat{\*y}_t(\*\beta)$ of $\*y_t$ obtained as posterior expectation via Eq. \eqref{Posterior_expectation}, define $\*\varepsilon_t(\*\beta) = \*y_t - \widehat{\*y}_t(\*\beta)$. 
For predictive loss $L:\mathbb{R}\to\mathbb{R}_{+}$, we target $\*\beta^{\ast} = \argmin_{\*\beta}\left\{\mathbb{E}\left(L(\*\varepsilon_t(\*\beta)) \right)\right\}$. 
Replacing expected by empirical loss and deploying \SGD, 
we seek to find the partial derivatives of $\nabla_{\beta} L\left(\varepsilon_t(\*\beta)\right) $. Noting that
	$\nabla_{\*\beta} L\left(\varepsilon_t(\*\beta)\right)) = L'\left(\varepsilon_t(\*\beta)\right))  \cdot \nabla_{\*\beta } \: \widehat{\*y}_t(\*\beta)$,
the issue reduces to finding the partial derivatives $\nabla_{\brlmm } \widehat{\*y}_t(\*\beta)$ and $\nabla_{\bpm } \widehat{\*y}_t(\*\beta)$. Remarkably, $\nabla_{\brlmm } \widehat{\*y}_t(\*\beta)$ can be updated sequentially and efficiently by differentiating the recursion in Eq. \eqref{BOCPD_recursion2}. The derivation is provided in the Appendix. The gradient $\nabla_{\bpm } \widehat{\*y}_t(\*\beta)$ on the other hand is not available analytically and thus is approximated numerically. Now, $\beta$ can be updated on-line via
\begin{IEEEeqnarray}{CCCCCC}
	\*\beta_t & =& \*\beta_{t-1} - \eta \cdot \begin{bmatrix}
	\nabla_{\brlmmtm} L\left(\varepsilon_t(\*\beta_{1:(t-1)})\right) \\
	 \nabla_{\bpmtm} L\left(\varepsilon_t(\*\beta_{1:(t-1)})\right))\\
	\end{bmatrix}
\end{IEEEeqnarray}
In spirit, this procedure resembles existing approaches for model hyperparameter optimization \cite{CaronDoucet}. 
For robustness, $L$ should be chosen appropriately. In our experiments $L$ is a bounded absolute loss. 





\begin{figure}[b!]
\begin{center}
\centerline{\includegraphics[trim= {2.95cm 1.6cm 4.5cm 2.1cm}, clip, 
width=1.00\columnwidth]{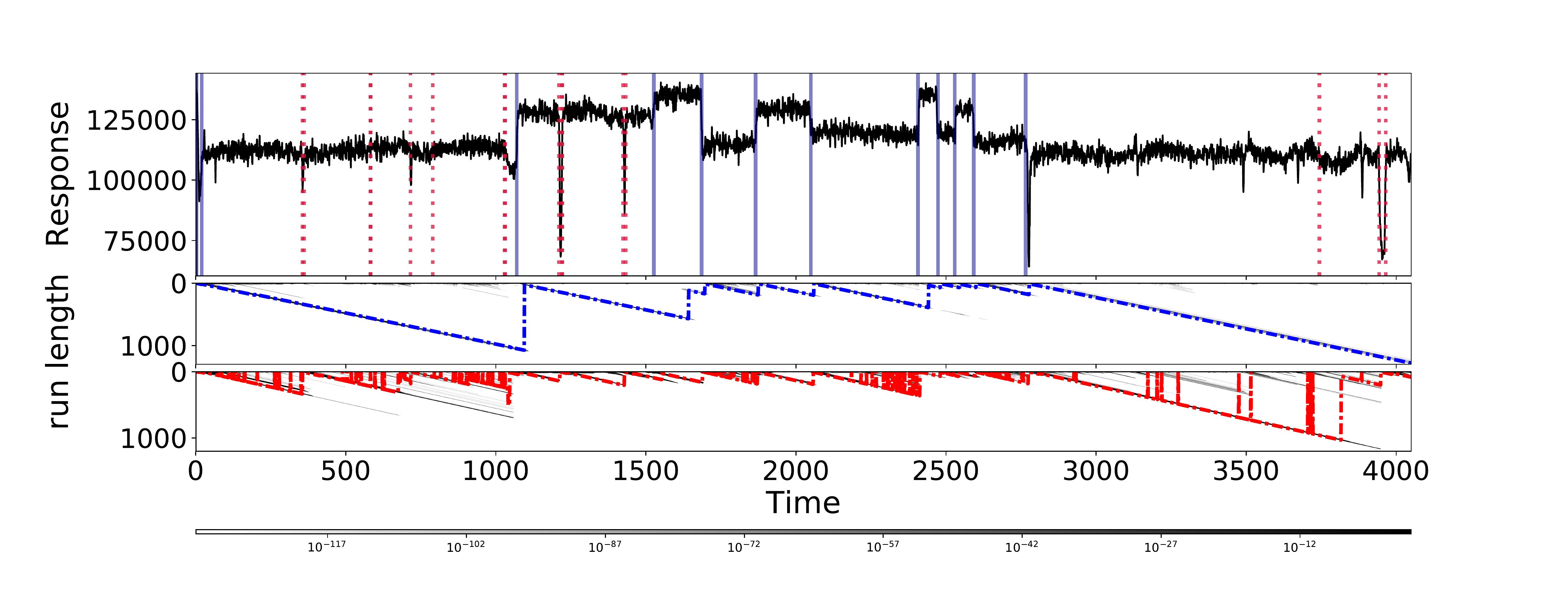}}
\caption{{ Maximum A Posteriori (\MAP) segmentation} and run-length distributions of the well-log data. 
{\color{blue} Robust } segmentation depicted using solid lines, \CPs additionally declared under {\color{red} standard} \BOCPD with dashed lines.
The corresponding run-length distributions for robust (middle) and standard (bottom) \BOCPD
are shown in grayscale. The most likely run-lengths are dashed.}
\label{figure:well_log}
\end{center}
\vskip -0.2in
\end{figure}

\section{Results}\label{result_section}

Next, we illustrate the most important improvements this paper makes to \BOCPD. First, we show how robust \BOCPD deals with outliers on the well-log data set.
Further, we show that standard \BOCPD breaks down in the M-open world whilst \BD yields useful inference by analyzing noisy measurements of Nitrogen Oxide (\NOX) levels in London.
%
In both experiments, we use the methods in section \ref{beta_choice}, 
 on-line hyperparameter optimization \cite{CaronDoucet} and pruning for $p(r_t, m_t|\*y_{1:t})$ \citep[][]{BOCD}. 
%
%
%
Detailed information is provided in the Appendix. Software and simulation code is available as part of a reproducibility award at \url{https://github.com/alan-turing-institute/rbocpdms/}.

\subsection{Well-log}\label{well_log}

The well-log data set was first studied in \citet{FirstWellLog} and has become a benchmark data set for univariate \CP detection.
However, except in \citet{FearnheadOutliersConstant} its outliers have been removed before \CP detection algorithms are run \citep[e.g.][]{BOCD, LassoCP, BSCPD2}. With $\mathcal{M}$ containing one \BLR model of form $y_t = \mu + \varepsilon_t$, Figure \ref{figure:well_log} shows that robust \BOCPD deals with outliers on-line. 
The maximum of the run-length distribution for standard \BOCPD is zero $145$ times, so declaring \CPs based on the run-length distribution's maximum \cite[see e.g.][]{GPBOCD} yields a False Discovery Rate (\FDR) > 90\%.
This problem persists even with non-parametric, Gaussian Process, models \cite[p. 186, ][]{TurnerThesis}.
Even using Maximum A Posteriori (\MAP) segmentation \cite{FearnheadOnlineBCD}, standard \BOCPD mislabels $8$ outliers as \CPs, making for a \FDR  > 40\%.
In contrast, the segmentation of the \BD version does not mislabel any outliers. 
Morevoer and in accordance with Thm. \ref{Thm:DeclareChange}, its run-length distribution's maximum never drops to zero in response to outliers. 
%
Further, a natural byproduct of the robust segmentation is a reduction in squared (absolute) prediction error by 10\% (6\%) compared to the standard version.
%
The robust version has more computational overhead than standard \BOCPD, but still needs less than 0.5 seconds per observation using a 3.1 \GHz Intel \iseven{ }and 16\GB \RAM. 

Not only does robust \BOCPD's segmentation in Figure \ref{figure:well_log} match that in \citet{FearnheadOutliersConstant}, but it also offers three additional on-line outputs: Firstly, it produces probabilistic (rather than point) forecasts and parameter inference. Secondly, it self-regulates its robustness via $\*\beta$. Thirdly, it can compare multiple models and produce model posteriors (see section \ref{AirPollution}).  
Further, unlike \citet{FearnheadOutliersConstant}, it is not restricted to fitting univariate data with piecewise constant functions. 

\subsection{Air Pollution}\label{AirPollution}


The example in Fig. \ref{figure:demo} \textbf{B} gives an illustration of the importance of robustness in medium-dimensional (\BOCPD) problems: It suffices for a \textit{single} dimension of the problem to be misspecified or outlier-prone for inference to fail. Moreover, the presence of misspecification or outliers in this plot can hardly be spotted -- and this effect will worsen with increasing dimensionality.  
%
To illustrate this point on a multivariate real world data set, we also analyze Nitrogen Oxide (\NOX) levels across $29$ stations in London using spatially structured Bayesian Vector Autoregressions \cite[see][]{BOCPDMS}. 
Previous robust on-line methods \cite[e.g.][]{ robust2, robust3, FearnheadOutliersConstant} cannot be applied to this problem because they assume univariate data or do not allow for dependent observations.
%
%
\begin{figure}[b!]
\begin{center}
\centerline{\includegraphics[trim= {2.4cm 2.4cm 2.85cm 1.95cm}, clip, 
width=1.00\columnwidth]{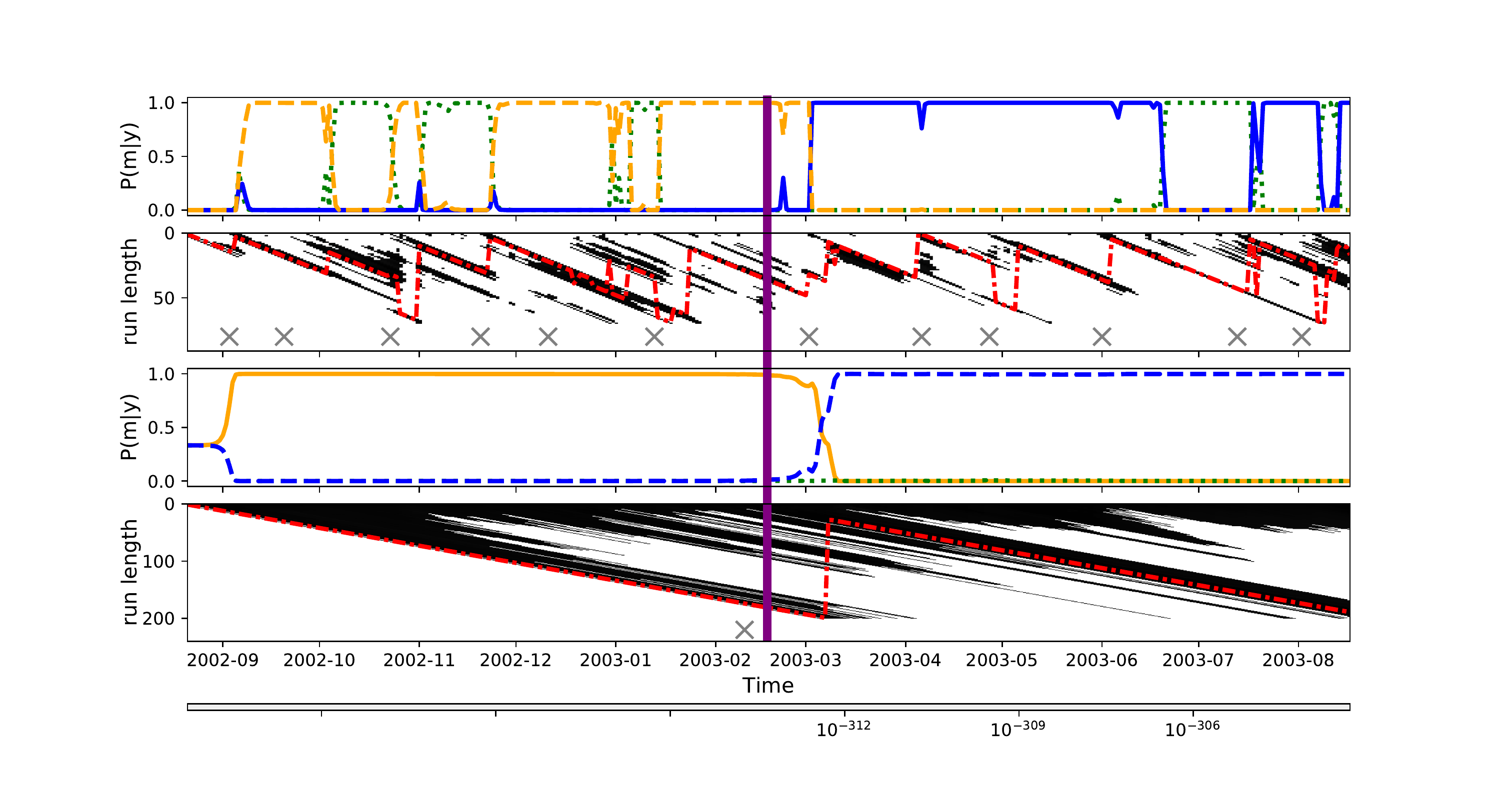}}
\caption{ On-line model posteriors for three different \VAR models ({\color{blue} solid}, {\color{pythonOrange} dashed}, {\color{darkgreen} dotted}) and run-length distributions in grayscale with {\color{red} most likely run-lengths} dashed for standard (top two panels) and robust (bottom two panels) \BOCPD. Also marked are the
{ \color{pythonPurple}congestion charge introduction}, 17/02/2003 (solid vertical line) and the {\MAP segmentations} ({\color{gray}crosses})}
\label{figure:AP}
\end{center}
\vskip -0.2in
\end{figure}
As Figure \ref{figure:AP} shows, robust \BOCPD finds one \CP corresponding to the introduction of the congestion charge, while standard \BOCPD produces an \FDR >90\%. Both methods find a change in dynamics (i.e.\ models) after the congestion charge introduction, but variance in the model posterior is substantially lower for the robust algorithm. 
Further, it increases the average one-step-ahead predictive likelihood  
by 10\% compared to standard \BOCPD. 

%
%


\section{Conclusion}

This paper has presented the very first robust Bayesian on-line changepoint (\CP) detection algorithm and the first ever scalable General Bayesian Inference (\GBI) method.
While \CP detection is a particularly salient example of unaddressed heterogeneity and outliers leading to poor inference, the capabilities of \GBI and the Structural Variational approximations presented extend far beyond this setting. 
With an ever increasing interest in the field of machine learning to efficiently and reliably quantify uncertainty, robust probabilistic inference will only become more relevant.
In this paper, we 
give a particularly striking demonstration of the inferential power  that can be unlocked through divergence-based General Bayesian inference.
%


\section*{Acknowledgements}

We would like to cordially thank both Jim Smith and Chris Holmes for fruitful discussions and help with some of the theoretical results. 
JK and JJ are funded by \EPSRC grant EP/L016710/1 as part of the Oxford-Warwick  Statistics Programme (\OxWaSP). TD is funded by the Lloyds Register Foundation programme on Data Centric Engineering through the London Air Quality project. This work was supported by The Alan Turing Institute for Data Science and AI under \EPSRC grant EP/N510129/1. In collaboration with the Greater London Authority.

\bibliography{library}
\bibliographystyle{plainnat}

\newpage

\appendix
\part*{Appendix} 
\parttoc 


\tableofcontents

\section{Student-t Experiments}

\begin{figure}[h!]
\begin{center}
\centerline{\includegraphics[trim= {0.5cm 2.25cm 0.5cm 1.6cm}, clip,  
width=1.00\columnwidth]{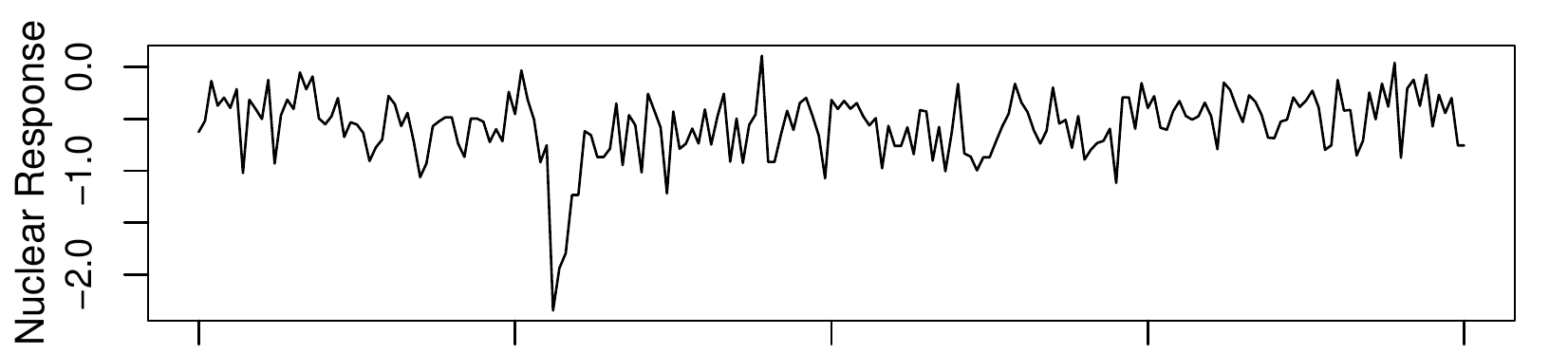}}
\centerline{\includegraphics[trim= {0.5cm 2.25cm 0.5cm 1.8cm}, clip,  
width=1.00\columnwidth]{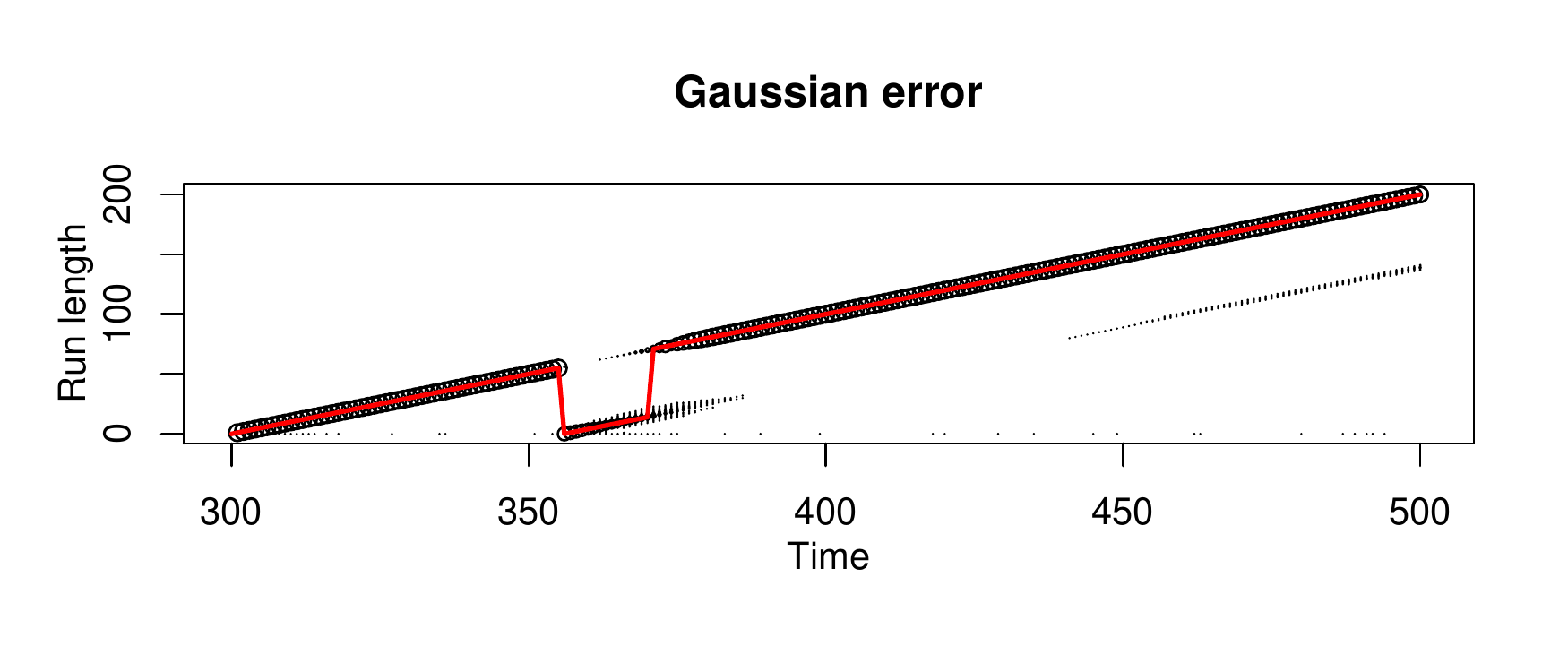}}
\centerline{\includegraphics[trim= {0.5cm 2.25cm 0.5cm 1.8cm}, clip,  
width=1.00\columnwidth]{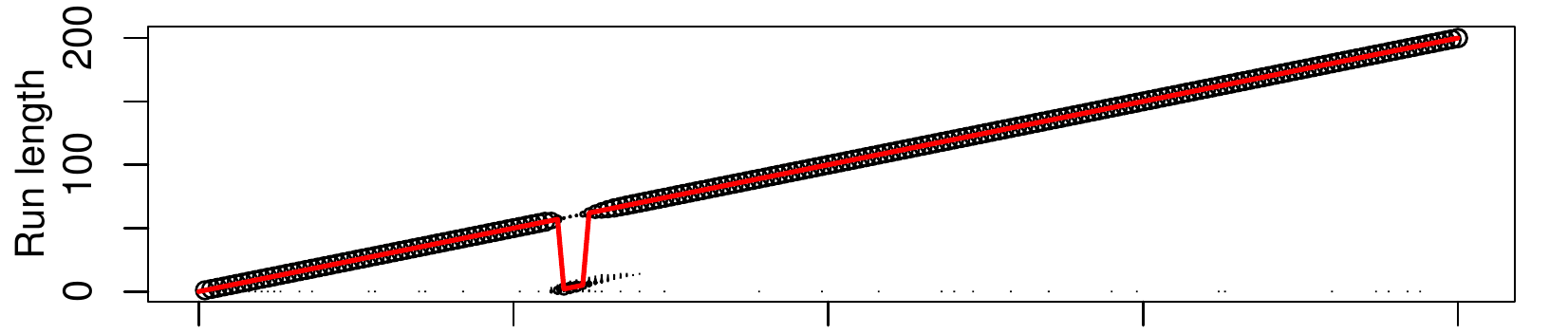}}
\centerline{\includegraphics[trim= {0.5cm 1.0cm 0.5cm 1.8cm}, clip,  
width=1.00\columnwidth]{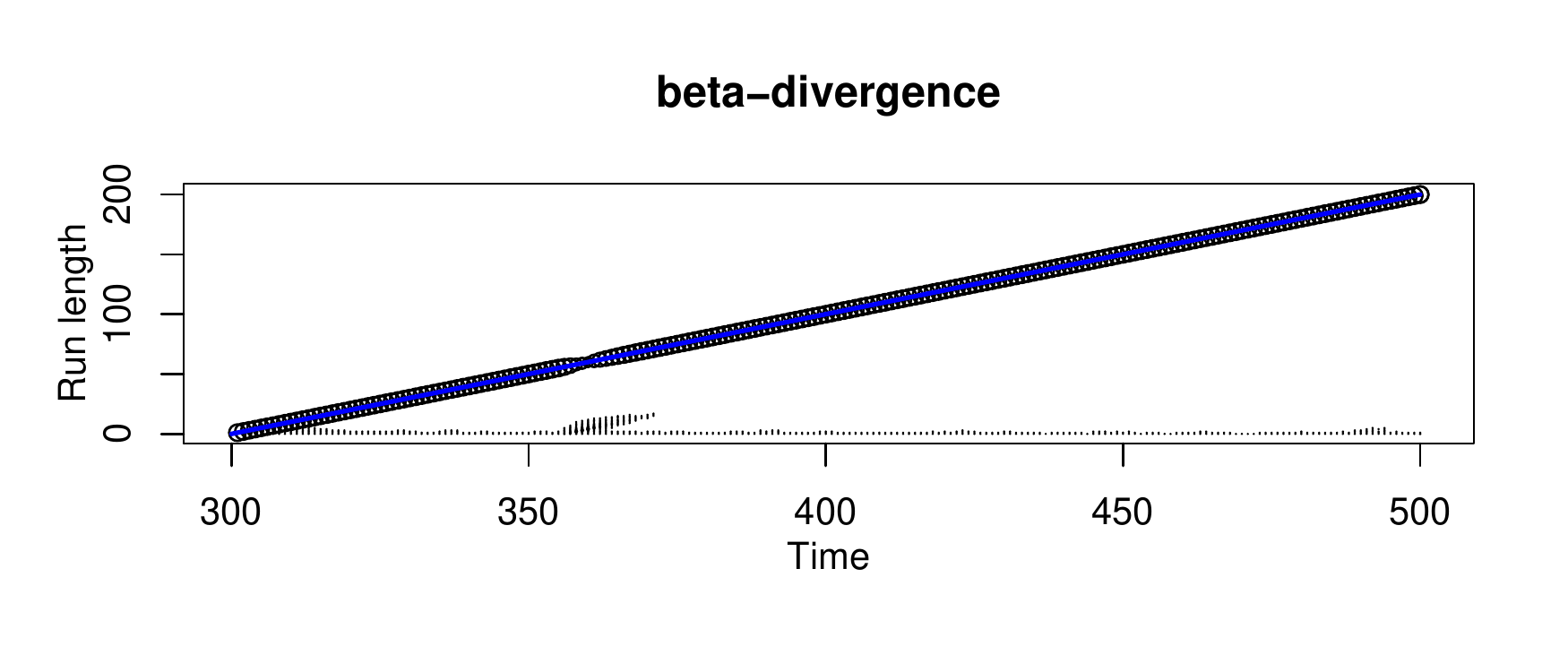}}
\caption{
{\textbf{Top:}} 
Plot of the \textit{well-log} data between $t=300$ and $t=500$ with one obvious outlying period.
{\textbf{Middle Top:}}
\KLD run length posterior under the Gaussian error model with the MAP of the run length posterior at each time point overlayed in \textcolor{red}{red}.
{\textbf{Middle Bottom:}}
\KLD run length posterior under the Student's $t$ with 5 degrees of freedom error model with the MAP of the run length posterior at each time point overlayed in \textcolor{red}{red}.
{\textbf{Bottom:}}
\BD run length posterior under the Gaussian error model with $\beta_p=0.25$ and \brlm$=0.5$ with the MAP of the run length posterior at each time point overlayed in \textcolor{blue}{blue}.}
\label{figure:Studentt1}
\vskip -0.25in
\end{center}
\end{figure}


{In Section 2.3 (Quantifying Robustness) of the paper we argue that substituting the Gaussian error model in the \BLR setting for a Student's $t$ error model -- a traditional solution for robust parameter inference -- will be insufficient to ensure that standard Bayesian run-length posteriors are robust. Here, the type of robustness we refer to is defined in Theorem 1. To demonstrate this, we implement a version of \BOCPD using both the Gaussian error model and the Student's $t$ error model on two subsets of the well-log data. The Student's $t$ distribution is no longer an exponential family and thus cannot be implemented in analytical form or via our structural variational approximation. Hence, we used \textit{stan} \citep{carpenter2016stan} for MCMC sampling from the parameter posterior under the Student's $t$ error model. 
For comparability, hyperparameters were fixed for both the Gaussian and Student's $t$ error models at $\mu_0=0$, $\Sigma_0=\sqrt{5}$, $a_0=0.5$, $b_0=2$, $h(r_{t+1})=0.01$ $\forall r_{t+1}$, where $N=1000$ values were sampled from the parameter posterior, $M=25$ run lengths were stored and the degrees of freedom of the Student's $t$ error model were set to be $\nu=5$. 
Figures \ref{figure:Studentt1} and \ref{figure:Studentt2} plot the \KLD run-length posteriors of the Gaussian and Student's-$t$ error models as well as the \BD run-length posteriors of the Gaussian error models for the two subsets of the \textit{well-log} data. In both examples, the \KLD run-length posteriors favor declaring a \CP under both the Gaussian and Student's $t$ error model at the first sign of an outlier. In the second example, the outlier is severe enough to permanently disrupt the run-length inference for both \KLD-based methods, while the \BD-based method remains robust. Theorem 1 outlines situations were this desireable behaviour of \BD-based inference can be guaranteed to happen when it would not happen under the \KLD with \textit{any} error model.}

\begin{figure}[h!]
\begin{center}
\centerline{\includegraphics[trim= {0.5cm 2.25cm 0.5cm 1.6cm}, clip,  
width=1.00\columnwidth]{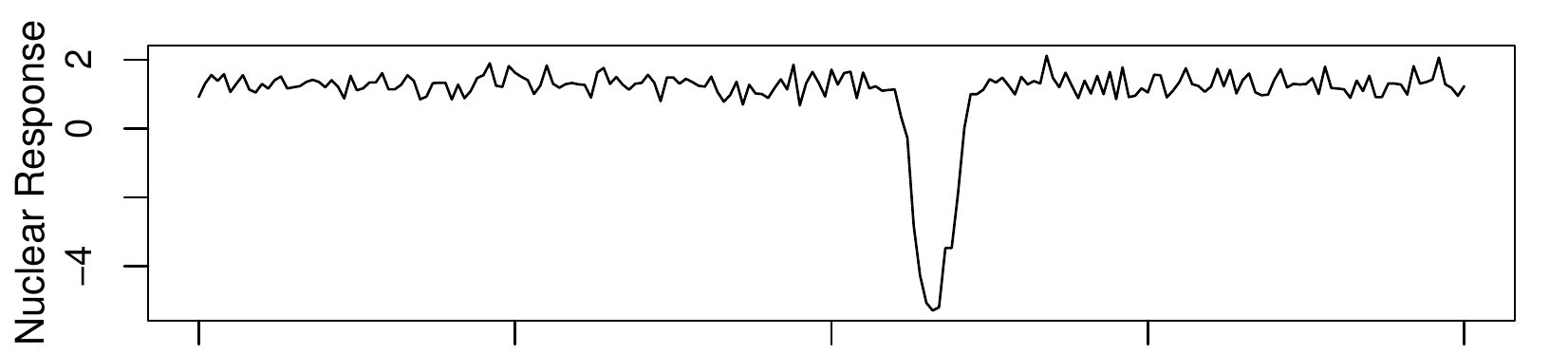}}
\centerline{\includegraphics[trim= {0.5cm 2.25cm 0.5cm 1.8cm}, clip,  
width=1.00\columnwidth]{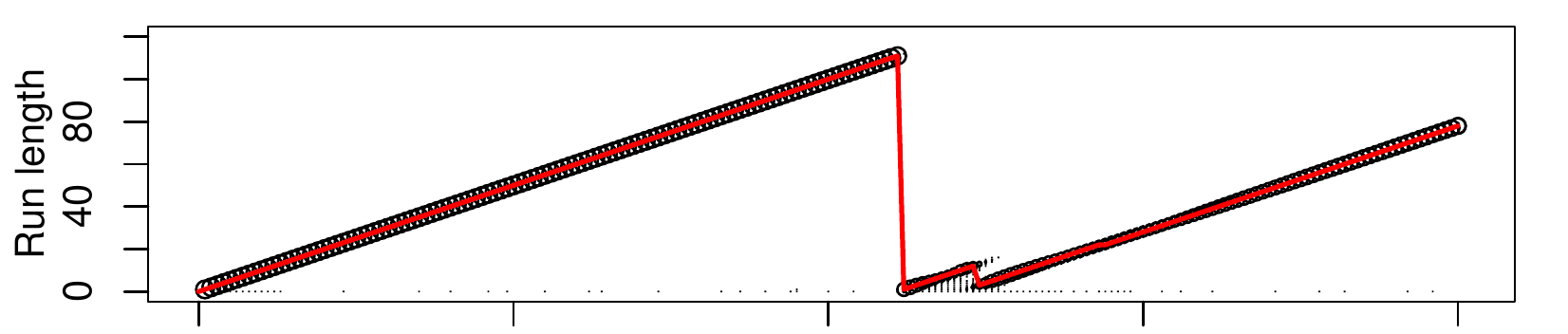}}
\centerline{\includegraphics[trim= {0.5cm 2.25cm 0.5cm 1.8cm}, clip,  
width=1.00\columnwidth]{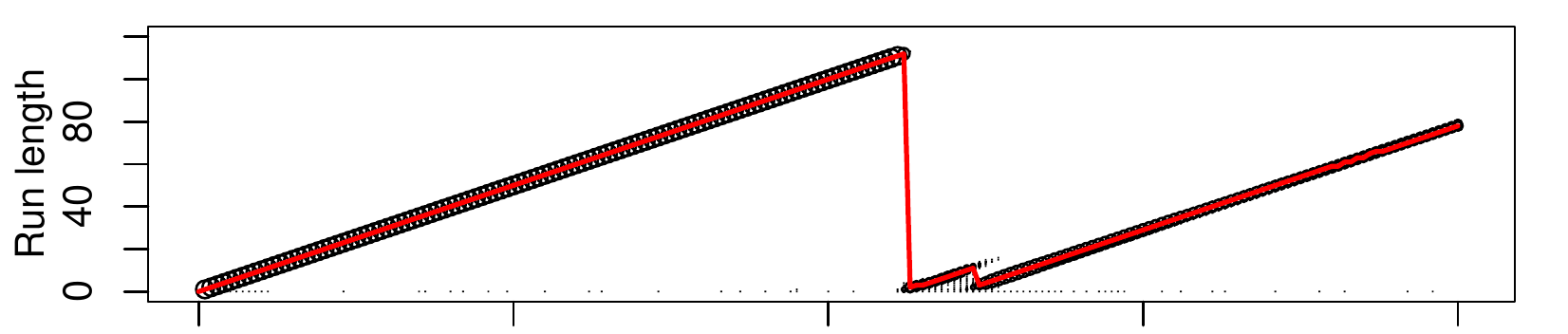}}
\centerline{\includegraphics[trim= {0.5cm 1.0cm 0.5cm 1.8cm}, clip,  
width=1.00\columnwidth]{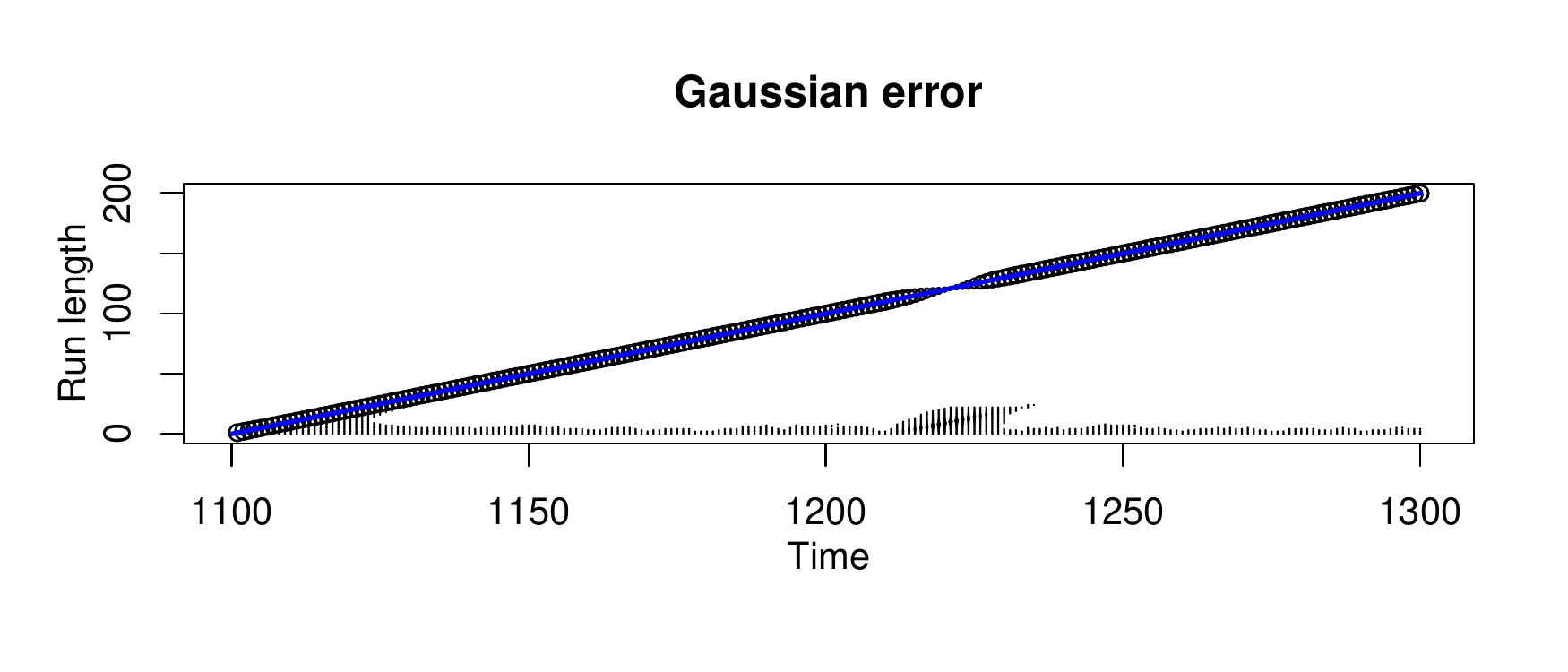}}
\caption{
{\textbf{Top:}} 
Plot of the \textit{well-log} data between $t=1100$ and $t=1300$ with one obvious outlying period.
{\textbf{Middle:}}
\KLD run length posterior under the Gaussian error model with the MAP of the run length posterior at each time point overlayed in \textcolor{red}{red}.
{\textbf{Bottom:}}
\KLD run length posterior under the Student's $t$ with 5 degrees of freedom error model with the MAP of the run length posterior at each time point overlayed in \textcolor{red}{red}.
{\textbf{Bottom:}}
\BD run length posterior under the Gaussian error model with $\beta_p=0.5$ and \brlm$=1$ with the MAP of the run length posterior at each time point overlayed in \textcolor{blue}{blue}.}
\label{figure:Studentt2}
\vskip -0.5in
\end{center}
\end{figure}

\newpage

\section{Proof of Theorem 1}

\begin{proof}
This proof looks at the run length posterior parameterised by \brlm, however to ease notation we refer to \brlm=$\beta$ throughout. Condition on the event $r_t=r$ then after one time step either $r_{t+1}=r+1$ or $r_{t+1}=0$. The odds of these two possibilities are as in Thm. 1.
Now substituting the definitions of 
$f_{m_t}^{\beta}(\*y_{t+1}|\mathcal{F}_{t} )$ and
$f_{m_t}^{\beta}(\*y_{t+1}|\*y_0)$
leaves

 \begin{IEEEeqnarray}{rCl}
&&\frac{f^{\beta}_{m_t}(\*y_{t+1}|\mathcal{F}_{t} )}{f^{\beta}_{m_t}(\*y_{t+1}|\*y_0)}=\frac{\exp\left(\frac{1}{\beta}p(\*y_{t+1}|\*y_{1:t})^{\beta}-\frac{1}{1+\beta}\int p(\*z|\*y_{1:t})^{1+\beta}d\*z\right)}{\exp\left(\frac{1}{\beta}p(\*y_{t+1}|\*y_{0})^{\beta}-\frac{1}{1+\beta}\int p(\*z|\*y_{0})^{1+\beta}d\*z\right)}\\
&&\quad=\exp\left(\frac{1}{\beta}\left(p(\*y_{t+1}|\*y_{1:t})^{\beta}-p(\*y_{t+1}|\*y_{0})^{\beta}\right)-\frac{1}{1+\beta}\int p(\*z|\*y_{1:t})^{1+\beta}- p(\*z|\*y_{0})^{1+\beta}d\*z\right).
\end{IEEEeqnarray}

This proof first seeks a lower bound for this ratio. A lower bound on $\frac{1}{\beta}p(\*y_{t+1}|\*y_{1:t})^{\beta}$ is 0, while the maximal value of $\frac{1}{\beta}p(\*y_{t+1}|x_{0})^{\beta}$ will occur at the prior mode. For the multivariate $t$-distribution prior predictive with NIG hyperparameters $a_0$, $b_0$, $\mu_0$, $\*\Sigma_0$ of dimensions $p$ the prior mode has density

\begin{IEEEeqnarray}{rCl}
p(\boldsymbol{\mu}_0|\nu_0,\boldsymbol{\mu}_0,\*V_0,p)&=&\frac{\Gamma((\nu_0+p)/2)}{\Gamma(\nu_0/2)\nu_0^{p/2}\pi^{p/2}\left|\*V_0\right|^{1/2}}\left[1+\frac{1}{\nu_0}(\boldsymbol{\mu}_0-\boldsymbol{\mu}_0)\*\Sigma_0^{-1}(\boldsymbol{\mu}_0-\boldsymbol{\mu}_0)\right]^{-(\nu_0+p)/2}\\
&=&\frac{\Gamma((\nu_0+p)/2)}{\Gamma(\nu_0/2)\nu_0^{p/2}\pi^{p/2}\left|\*V_0\right|^{1/2}}\\
&=&\frac{\Gamma(a_0+p/2)}{\Gamma(a_0)\left(2b_0\pi\right)^{p/2}\left|\*I+\*X\*\Sigma_0\*X^T\right|^{1/2}}.
\end{IEEEeqnarray}

As a result the only term in the lower bound of  $f_{m_t}^{\beta}(\*y_{t+1}|\mathcal{F}_{t} )/f_{m_t}^{\beta}(\*y_{t+1}|\*y_0)$ that does not solely depend on the prior parameters is $\frac{1}{1+\beta}\int p(\*z|\*y_{1:t})^{1+\beta}d\*z$. This term appears in the negative and thus to lower bound ${f^{\beta}_{m_t}(\*y_{t+1}|\mathcal{F}_{t} )}/{f^{\beta}_{m_t}(\*y_{t+1}|\*y_0)}$, an upper bound for $\frac{1}{1+\beta}\int p(\*z|\*y_{1:t})^{1+\beta}d\*z$ must be found. The multivariate t-distribution can be integrated as

\begin{IEEEeqnarray}{rCl}
\frac{1}{1+\beta}&\int& \textrm{MVSt}_{\nu}(\*z|\boldsymbol{\mu},\*V)^{1+\beta}d\*z=\frac{\Gamma((\nu+p)/2)^{\beta+1}\Gamma((\beta\nu+\beta p+\nu)/2)}{\Gamma(\nu/2)^{\beta+1}\Gamma((\beta\nu+\beta p+\nu+p)/2)}\frac{1}{(1+\beta)(\nu\pi)^{(\beta p)/2}\left|\*V\right|^{\beta/2}}\\
&=&\frac{\Gamma((\nu+p)/2)^{\beta}\Gamma((\nu+p)/2)\Gamma((\beta\nu+\beta p+\nu)/2)}{\Gamma(\nu/2)^{\beta}\Gamma(\nu/2)\Gamma((\beta\nu+\beta p+\nu+p)/2)}\frac{1}{(1+\beta)(\pi\nu)^{(\beta p)/2}\left|\*V\right|^{\beta/2}}\\
&\leq&\frac{\Gamma((\nu+p)/2)^{\beta}}{\Gamma(\nu/2)^{\beta}}\frac{1}{(1+\beta)(\pi\nu)^{(\beta p)/2}\left|\*V\right|^{\beta/2}}\label{Equ:IntMVtGammaBound}.
\end{IEEEeqnarray}

The inequality is derived from the fact that $\frac{\Gamma\left(x+\frac{p}{2}\right)}{\Gamma(x)}$ is increasing in $x$ and as $\beta\geq 0$ and $\nu\geq 0$ then $(\beta\nu+\beta p+\nu)/2)\geq \nu/2$ which implies $\frac{\Gamma((\nu+p)/2)\Gamma((\beta\nu+\beta p+\nu)/2)}{\Gamma(\nu/2)\Gamma((\beta\nu+\beta p+\nu+p)/2)}\leq 1$. 

Now employing the well-known result using Stirling's formula to bound the gamma function 

\begin{IEEEeqnarray}{rCl}
(2\pi)^{1/2}x^{x-1/2}\exp(-x)\leq \Gamma(x)\leq (2\pi)^{1/2}x^{x-1/2}\exp(1/(12x)-x)
\end{IEEEeqnarray}

we can therefore rewrite the ratio of gamma functions leaving

\begin{IEEEeqnarray}{rCl}
&&\frac{1}{1+\beta}\int \textrm{MVSt}-t_{\nu}(\*z|\boldsymbol{\mu},\*V)^{1+\beta}d\*z
\leq\frac{\Gamma((\nu + p)/2)^{\beta}}{\Gamma(\nu/2)^{\beta}}\frac{1}{(1+\beta)(\pi\nu)^{(\beta p)/2}\left|\*V\right|^{\beta/2}}\\
&\quad&\leq\frac{\left(\sqrt{2\pi}((\nu+p)/2)^{(\nu+p-1)/2)}\exp(-(\nu+p)/2+1/6(\nu+p)\right)^{\beta}}{\left(\sqrt{2\pi}(\nu/2)^{(\nu-1)/2}\exp(-\nu/2)\right)^{\beta}(1+\beta)(\pi\nu)^{(\beta p)/2}\left|\*V\right|^{\beta/2}}\\
&\quad&=((1+\frac{p}{\nu})^{\beta(\nu+p-1)/2)}\exp(\beta(1/(6(\nu+p)) -p/2))\frac{1}{(1+\beta)(\pi)^{(\beta p)/2}\left|\*V\right|^{\beta/2}}.\label{Equ:StirlingExpansion}
\end{IEEEeqnarray} 
         
Clearly $\exp\left(\beta(1/(6(\nu+p)) -p/2)\right)$ is decreasing in $\nu$ for all $p$ and to demonstrate when $((1+\frac{p}{\nu})^{\beta(\nu+p-1)/2)}$ is decreasing in $\nu$ we examine its derivative 
         
\begin{IEEEeqnarray}{rCl}
w&=&\left(1+\frac{p}{\nu}\right)^{\beta(\nu+p-1)/2}\\   &=&\exp\left(\left(\beta(\nu+p-1)/2\right)\log\left(\left(1+\frac{p}{\nu}\right)\right)\right)\\
\frac{dw}{d\nu}&=&\frac{\beta}{2}\left(\log \left(1+\frac{p}{\nu}\right)-(\nu+p-1)\frac{\frac{p}{\nu^2}}{1+\frac{p}{\nu}}\right)\left(1+\frac{p}{\nu}\right)^{\beta(\nu+p-1)/2)}.
\end{IEEEeqnarray}

The sign of $\frac{dw}{d\nu}$ is dictated by $\left(\log \left(1+\frac{p}{\nu}\right)-(\nu+p-1)\frac{\frac{p}{\nu^2}}{1+\frac{p}{\nu}}\right)$, which can be demonstrated to be positive always if $p=1$ and negative always if $p>1$. 

\textbf{Case 1}: when $p>1$, $\frac{1}{1+\beta}\int p(\*z|\*y_{1:t})^{1+\beta}d\*z$ is decreasing in $\nu$ and thus we can upper bound it by substituting the smallest value of $\nu$. Here we bound $\nu$ above 1 in order to enforce that the mean of the predictive $t$-distribution exists. Under the \KLD posterior it is clear that $a_0$ rises as more data is seen and while we do not have closed forms associated with the variational approximation to the \BD posterior we expect this to be the case here. As more data is seen the finite sampling uncertainty, represented by $\nu$ in the \NIG case, should be decreasing. Therefore provided $a_0$ is set such that $2a_0>1$, then this lower bound should never be violated.

\textbf{Case 2}: when $p=1$, Stirling's formula has failed to provide a decreasing upper bound for $\frac{1}{1+\beta}\int p(\*z|\*y_{1:t})^{1+\beta}dz$.  However in the univariate case

\begin{IEEEeqnarray}{rCl}
\frac{1}{1+\beta}\int \textrm{St}_{\nu}(\*z|\boldsymbol{\mu},\*V)^{1+\beta}d\*z
&\leq&\frac{\Gamma((\nu+1)/2)^{\beta}}{\Gamma(\nu/2)^{\beta}}\frac{1}{(1+\beta)(\nu\left|\*V\right|)^{(\beta)/2}\pi^{(\beta)/2}}\\
&\leq&\frac{1}{(1+\beta)\left|\*V\right|^{(\beta)/2}\pi^{(\beta)/2}}
\end{IEEEeqnarray}
Where $p=1$ is substituted into the bound from equation (\ref{Equ:IntMVtGammaBound}) and the inequality comes from that fact that $\frac{\Gamma((x+1)/2)}{\Gamma(x/2)}\leq \sqrt{x}$. This bound conveniently does not depend on the degrees of freedom $\nu$ at all. 

We can therefore lower bound ${f^{\beta}_{m_t}(\*y_{t+1}|\mathcal{F}_{t} )}/{f^{\beta}_{m_t}(\*y_{t+1}|\*y_0)}$ as
\begin{IEEEeqnarray}{rCl}\label{Equ:Inequality1}
\frac{f^{\beta}_{m_t}(\*y_{t+1}|\mathcal{F}_{t} )}{f^{\beta}_{m_t}(\*y_{t+1}|\*y_0)}&\geq&\begin{cases}
&\exp\left\lbrace-\frac{1}{\beta}\left(\frac{\Gamma(a_0+1/2)}{\Gamma(a_0)\left(2b_0\pi\right)^{1/2}\left|\*I+\*X\*\Sigma_0\*X^T\right|^{1/2}}\right)^{\beta}-\frac{1}{(1+\beta)\left|\*V\right|^{(\beta)/2}\pi^{(\beta)/2}}+\right.\\
&\quad\left.\frac{\Gamma(a_0+1/2)^{\beta+1}\Gamma(\beta a_0+\beta/2+a_0)}{\Gamma(a_0)^{\beta+1}\Gamma(\beta a_0+\beta/2+a_0+1/2)}\frac{1}{(1+\beta)(2\pi b_0)^{(\beta)/2}\left|\*I+\*X\*\Sigma_0\*X^T\right|^{\beta/2}}\right\rbrace  \textrm{ if } p=1\\
&\exp\left\lbrace-\frac{1}{\beta}\left(\frac{\Gamma(a_0+p/2)}{\Gamma(a_0)\left(2b_0\pi\right)^{p/2}\left|\*I+\*X\*\Sigma_0\*X^T\right|^{1/2}}\right)^{\beta}+\right.\\
&\quad\left.\frac{\Gamma(a_0+p/2)^{\beta+1}\Gamma(\beta a_0+\beta p/2+a_0)}{\Gamma(a_0)^{\beta+1}\Gamma(\beta a_0+\beta p/2+a_0+p/2)}\frac{1}{(1+\beta)(2\pi b_0)^{(\beta p)/2}\left|\*I+\*X\*\Sigma_0\*X^T\right|^{\beta/2}}-\right.\\
&\quad\left.((1+p)^{\beta p/2)}\exp(\beta(1/(6(1+p)) -p/2))\frac{1}{(1+\beta)(\pi)^{(\beta p)/2}\left|\*V\right|^{\beta/2}}\right\rbrace\textrm{ if } p>1
\end{cases}
\end{IEEEeqnarray}

Now fixing $p,a_0,b_0,\mu_0,\Sigma_0$ and $\left|V\right|_{\min}$ which values of $\beta$ and $H(r_t,r_{t+1})$ would leave 
\begin{IEEEeqnarray}{rCl}\label{Equ:MAPCondition}
\frac{1-H(r_t,r_{t+1}))}{H(r_t,r_{t+1})}\frac{f^{\beta}_{m_t}(\*y_{t+1}|\mathcal{F}_{t} )}{f^{\beta}_{m_t}(\*y_{t+1}|\*y_0)}\geq 1?
\end{IEEEeqnarray}

We demonstrate this for $p>1$ but it is straightforward to see that it extends to when $p=1$. Rearranging the inequality in equation (\ref{Equ:Inequality1}) gives us that (\ref{Equ:MAPCondition}) holds providing

\begin{IEEEeqnarray}{rCl}\label{Equ:Inequality2}
&&\frac{1}{\left|\*V\right|^{\beta/2}}\leq \nonumber\\
&&\left(\frac{\Gamma(a_0+p/2)^{\beta}}{\Gamma(a_0)^{\beta}\left(2b_0\pi\right)^{\beta p/2}\left|\*I+\*X\*\Sigma_0\*X^T\right|^{\beta/2}}\left(\frac{\Gamma(a_0+p/2)\Gamma(\beta a_0+\beta p/2+a_0)}{\Gamma(a_0)\Gamma(\beta a_0+\beta p/2+a_0+p/2)}\frac{1}{(1+\beta)}-\frac{1}{\beta}\right)\right.\\
&&\left.+\log\left(\frac{1-H(r_t,r_{t+1}))}{H(r_t,r_{t+1})}\right)\right)\frac{(1+\beta)(\pi)^{(\beta p)/2}}{((1+\frac{p}{2a_0})^{\alpha(2a_0+p-1)/2)}\exp(\beta(1/(6(2a_0+p)) -p/2))}\nonumber
\end{IEEEeqnarray}

We define the set defined by inequality (\ref{Equ:Inequality2}) as $S\left(p,\beta,a_0,b_0,\mu_0,\*\Sigma_0, \left|\*V\right|_{\min}\right) = \left\{ (\beta, H(r_t, r_{t+1})): (\beta, H(r_t, r_{t+1})) \text{ satisfy \eqref{Equ:Inequality2} for } p,\beta,a_0,b_0,\mu_0,\*\Sigma_0, \left|\*V\right|_{\min}\right\}$. As a result we can see that for fixed of $a_0,b_0,\mu_0,\Sigma_0$ and $\left|\*V\right|\geq \left|\*V\right|_{\min}$ it is always possible to choose values of $\beta$ and  $H(r_t,r_{t+1})$ such that this holds. To see this consider fixing $\beta$, the the upper bound is simply increasing in $\log\left(\frac{1-H(r_t,r_{t+1})}{H(r_t,r_{t+1})}\right)$ which takes values in $\mathbb{R}$ and thus can be set large enough so that the inequality holds.
\end{proof}

We note that in practice this results is likely to be stronger than is necessary. The observation that is most likely to generate a change-point will have 0 mass under the predictive associated with the current segment but also appears at the prior mode. While this was necessary to demonstrate this result for all situations this is incredibly unlikely to occur. The requirement for $\left|\*V_{\min}\right|$ is a result of the beta-divergence loss function depending on $\int p(z|\*y_{1:t})^{1+\beta}dz$. In the proof of this result we demonstrate that ${f^{\beta}_{m_t}(\*y_{t+1}|\mathcal{F}_{t} )}/{f^{\beta}_{m_t}(\*y_{t+1}|\*y_0)}$ is increasing in $\left|\*V\right|$ and as a result if it is allowed to get too small the inequality in equation (\ref{Equ:Inequality2}) would not hold. This is an undesirable consequence of the beta-divergence score not being completely local, that is to say not solely depending on the predictive probability of the observation, thus the score under the prior can be quite a lot bigger than the score under the continuing run length independent of the observations seen and solely based on the predictive covariances.

\section{Variational Bayes Approximation for $\beta$-divergence based General Bayesian Inference with the Bayesian Linear Model}

For ease of notation, we use $\beta = \bpm$.
We wish to approximate the posterior belief distribution $\dpd(\berv, \srv|\*y)$ which for observations $\*y = (\*y_1, \*y_2, \dots, \*y_n)^T$ with $\*y_i \in \mathbb{R}^d$, prior $\NIGz$, model likelihood $f$ and density power divergence (DPD) loss 
\begin{IEEEeqnarray}{rCl}
	\loss(\berv, \srv| \*y_i) & =  & \frac{1}{\beta} f(\*y_i|\berv, \srv)^{\beta} - \frac{1}{1+\beta} \int_{\mathcal{Y}} f(\*y_i|\berv, \srv)^{1+\beta}d\*y
\end{IEEEeqnarray}
 is given by
\begin{IEEEeqnarray}{rCl}
	 \dpd(\berv, \srv|\*y) & = & \NIGz \cdot 
	 	\text{exp} \left\{
			-\sum_{i=1}^n  \loss(\berv, \srv| \*y_i) 	 	
	 	\right\}.
\end{IEEEeqnarray}
In particular, we want to approximate it with a posterior $\NIGVB$ via  Variational Bayes. This can be done by minimizing the variational parameters in a Kullback-Leibler sense:
\begin{IEEEeqnarray}{rCl}
(a^{\ast}, b^{\ast}, \berv^{\ast}, \*\Sigma^{\ast} ) & = & 
	\arg\hspace*{-0.7cm}\min_{(\an, \bn, \ben, \Sn)}
		\left\{ 
			\text{KL}\left(\dpd(\berv, \srv|\*y) \left\|  \NIGVB \right)\right.
		\right\}.
\end{IEEEeqnarray}
It is straightforward to rewrite the objective function for the above minimization as the Evidence Lower Bound (ELBO) induced by the DPD:
\begin{IEEEeqnarray}{rCl}
	\text{ELBO}_{\text{DPD}} & = & 
		 - \underbrace{\text{KL}\left(\NIGVB \left\|  \NIGz \right)\right.}_{ = Q_1}.\nonumber \\
		& & 
		\quad\quad  \underbrace{- \E_{\text{VB}}\left[ -\sum_{i=1}^n  \loss(\berv, \srv| \*y_i) \right]}_{ = { Q_2 } }.
\end{IEEEeqnarray}

In what follows, closed forms are derived for both $Q_1$ and $Q_2$. Some algebraic tricks will be applied multiple times, and will be referred to by the following symbols:
\begin{itemize}
	\item[$\blacksquare$] Completion of Squares, i.e. $\*u'\*A\*u - 2\*v'\*u = (\*u - \*A^{-1}\*v)'\*A(\*u - \*A^{-1}\*v) - \*v'\*A^{-1}\*v $;
	\item[$I(\mathcal{N})$] Integrating out the Normal density;
	\item[$I(\mathcal{IG})$] Integrating out the Inverse Gamma density.
\end{itemize}
Throughout, the dimensionality of $\berv$ is $p \in \mathbb{N}$, $\mathcal{N}(\berv|\bez, \Sz)$ refers to a normal pdf in $\berv$ with expectation $\bez$, variance $\Sz$ and $\mathcal{IG}(\srv|a,b)$ to an inverse gamma pdf in $\srv$ with shape $a$ and scale $b$.

\subsection{$Q_1$}

First, note that by definition,
\begin{IEEEeqnarray}{rCl}
	Q_1 = \bigintss_{\berv, \srv} \underbrace{\log\left( \dfrac{\NIGVB}{\NIGz}\right)}_{ = Q_1^{\log}}
			 \NIGVB d\berv d\srv.
\end{IEEEeqnarray}
Writing out $Q_1^{\log}$, one obtains a natural sum of three components $C_1, C_2(\srv), C_3(\srv, \berv)$: 
\begin{IEEEeqnarray}{rCl}
	Q_1^{\log} 
	& = 	& \log\left( 
		\dfrac{
			|\Sn|^{-0.5}\frac{\bn^{\an}}{\Gamma(\an)}(\srv)^{-0.5p - \an - 1} 
			\exp\left\{ -\frac{1}{2\srv} \left[ (\berv - \ben)'\Sninv(\berv - 
			\ben) + 2\bn \right]	\right\}
		}
		{
			|\Sz|^{-0.5}\frac{\bz^{\az}}{\Gamma(\az)}(\srv)^{-0.5p - \az - 1} 
			\exp\left\{ -\frac{1}{2\srv} \left[ (\berv - \bez)'\Szinv(\berv - 
			\bez) + 2\bz \right]	\right\}
		}
		\right) \nonumber \\
	& = 	&	
		\underbrace{
		\log
			\left( 
				\dfrac{\bn^{\an}\Gamma(\az)}{\bz^{\az}\Gamma(\an)} 
			\right) + 	
				0.5\log\left| \Sz\Sninv \right|^{0.5}
		}_{ = C_1} + 
		\underbrace{ (\an - \az)\log(\frac{1}{\srv})  
		}_{ = C_2(\srv)}  \nonumber  \\
	&		& \quad\quad  
		-
		\underbrace{ 
			\frac{1}{2\srv}\left[
					(\berv - \ben)'\Sninv(\berv -  \ben)	 -
					(\berv - \bez)'\Szinv(\berv - \bez) + 
					2(\bn - \bz)
			\right]		
		}_{ = C_3(\srv, \berv)}.
\end{IEEEeqnarray}
Next, note that $C_3(\srv, \berv)$ further decomposes into
\begin{IEEEeqnarray}{rCl}
&&
	\underbrace{
		\frac{1}{2\srv}\left[
		\berv'\left(\Sninv - \Szinv\right)\berv - 2\berv'\left(\Sninv\ben - 	
		\Szinv\bez \right)\right]
	}_{ = {C}_4(\srv, \berv)}
	 + 
	\underbrace{
		\frac{1}{\srv}
		\underbrace{
			\left[ \frac{1}{2}\ben'\Sninv\ben - \frac{1}{2}\bez'\Szinv\bez +(\bn - \bz) \right]
		}_{ = C_5 }
	}_{ = C_6(\srv)}.
	\nonumber 
\end{IEEEeqnarray}
Notice that we have isolated the random variable $\berv$ inside ${C}_4(\srv, \berv)$ and that by definition, $\NIGVB = \mathcal{N}^{\text{VB}}(\berv|\ben, \srv\Sn) \cdot \mathcal{IG}^{\text{VB}}(\srv|\an, \bn)$, meaning that
\begin{IEEEeqnarray}{rCl}
	Q_1 & = & 
	C_1 + 
	\bigintssss_{\srv}
		\left\{C_2(\srv) - C_6(\srv)\right\}
		\mathcal{IG}^{\text{VB}}(\srv|\an, \bn)
		d\srv  \nonumber \\
	&&\quad\quad\quad
	- \bigintssss_{\srv}
		\underbrace{
			\left\{
			\bigintssss_{\berv}
				C_4(\srv, \berv)\mathcal{N}^{\text{VB}}(\berv|\ben, \srv\Sn)d\berv
			\right\}
		}_{ = C_7(\srv)}
	\mathcal{IG}^{\text{VB}}(\srv|\an, \bn)d\srv.	
\end{IEEEeqnarray}
The inner integral is available in closed form, and naturally decomposes as
\begin{IEEEeqnarray}{rCl}
	 C_7(\srv) & = & 
		\frac{1}{2\srv}\E_{\mathcal{N}^\text{VB}}\left[ 
			\berv'\left(\Sninv - \Szinv\right)\berv
		\right] - 
		\frac{2}{2\srv} \E_{\mathcal{N}^\text{VB}}\left[ 
			\berv'\right] \left(\Sninv\ben - 	\Szinv\bez \right) \nonumber \\
	& = & 
		\frac{1}{2\srv}\E_{\mathcal{N}^\text{VB}}\left[ 
			\text{tr}\left( 
				\left(\Sninv - \Szinv\right)\berv\berv'	
			\right) 
		\right]  - 
		\frac{1}{\srv}\ben'\left(\Sninv\ben - 	\Szinv\bez \right) \nonumber \\
	& = &
		\frac{1}{2\srv}\text{tr}\left( 
				\left(\Sninv - \Szinv\right)\E_{\mathcal{N}^\text{VB}}\left[ \berv\berv'\right] 	
			\right) 
		 - 
		\frac{1}{\srv}\ben'\left(\Sninv\ben - 	\Szinv\bez \right) \nonumber \\
	& = &
		\frac{1}{2\srv}\text{tr}\left( 
				\left(\Sninv - \Szinv\right)\left[\srv\Sn - \ben\ben' \right]
			\right) 
		 - 
		\frac{1}{\srv}\ben'\left(\Sninv\ben - 	\Szinv\bez \right) \nonumber \\
	& = &
			\underbrace{
				\frac{1}{2}
				\text{tr}\left(I - \Szinv\Sn\right)
			}_{  = C_8 } - 
		\underbrace{
		\frac{1}{\srv}
		\underbrace{
			\left[
				\frac{1}{2}\ben'(\Sninv - \Szinv)\ben - 
				\ben'\left(\Sninv\ben - 	\Szinv\bez \right)
			\right]
		}_{ = C_9}
		}_{ = C_{10}(\srv)}.
\end{IEEEeqnarray}
We may now rewrite $Q_1$ so as to integrate out $\srv$ next:
\begin{IEEEeqnarray}{rCl}
	Q_1 & = & 
	C_1 - C_{8} +
	\bigintssss_{\srv}
		\left\{C_2(\srv) - C_6(\srv) - C_{10}(\srv)\right\}
		\mathcal{IG}^{\text{VB}}(\srv|\an, \bn)
		d\srv.  \nonumber 
\end{IEEEeqnarray}
Using the additivity of integrals, we consider its three components separately and then add them up together afterwards. For $C_2(\srv)$, (I)  apply a change of variable with $z = \frac{\srv}{\bn}$ and then use (II) that $\frac{d}{dx}a^{-x} = -a^x \cdot \log(a) = a^x \cdot \log(a^{-1})$ together with Fubini's Theorem (III) to find that 
\begin{IEEEeqnarray}{rCl}
	C_{11} & 	= &  \bigintssss_{\srv} C_2(\srv) \mathcal{IG}^{\text{VB}}(\srv|\an, \bn) d\srv 
				\nonumber \\
	& = 	& 
		(\an - \az) 
		\bigintssss_{\srv} 
		\log\left(\frac{1}{\srv}\right)
		\dfrac{\bn^{\an}}{\Gamma(\an)}(\srv)^{-\an -1} 
		\exp\left\{ -\frac{\bn}{\srv}\right\} d\srv
		\nonumber \\
	& \overset{\text{(I)}}{=} & 
		(\an - \az) 
		\bigintssss_{z}
		 \log\left(\frac{1}{z\bn}\right)
		 \dfrac{\bn^{\an+1}}{\Gamma(\an)}
		 \left({z}{\bn}\right)^{-\an -1} 
		 \exp\left\{ -\frac{1}{z} \right\} 
		 dz 
		 \nonumber \\
	& = &
		(\an - \az) \frac{1}{\Gamma(\an)}
			\bigintssss_{z}
		 	\left(	-\log(z)	-\log(\bn)	\right)
		 	z^{-\an -1} 
		 	\exp\left\{ -\frac{1}{z} \right\} 
		 	dz 
		 \nonumber \\
	& \overset{I(\mathcal{IG})}{=} & 
		(\an - \az)
		\left[
		 	\frac{1}{\Gamma(\an)}
			\bigintssss_{z}
		 	\left(	-\log(z)		\right)
		 	z^{-\an -1} 
		 	\exp\left\{ -\frac{1}{z} \right\} 
		 	dz 
		 	- \log(\bn)
		\right]
		 \nonumber \\
	& \overset{\text{(II)}}{=} & 
		(\an - \az)
		\left[
		 	\frac{1}{\Gamma(\an)}
			\bigintssss_{z}
			\frac{d}{d\an}
			\left\{
		 		z^{-\an -1} 
		 		\exp\left\{ -\frac{1}{z} \right\} 
		 	\right\}
		 	dz 
		 	- \log(\bn)
		\right]
		 \nonumber \\
	& \overset{\text{(III)}}{=} & 
		(\an - \az)
		\left[
		 	\frac{1}{\Gamma(\an)}
		 	\frac{d}{d\an}
		 		\underbrace{
		 			\left\{
					\bigintssss_{z}	
		 				z^{-\an -1} 
		 				\exp\left\{ -\frac{1}{z} \right\} 
		 				dz 
		 			\right\}
		 			}_{ \overset{I(\mathcal{IG})}{=} \Gamma(\an)}
		 	- \log(\bn)
		\right]
		 \nonumber \\
	& = &
		(\an - \az)\left( \frac{\Gamma'(\an)}{\Gamma(\an)} - \log(\bn) \right)  \nonumber \\
	& = &
		(\an - \az)\left( \Psi(\an)- \log(\bn) \right),  \nonumber
\end{IEEEeqnarray}
where $\Psi$ is the digamma function. 
For $ C_6(\srv) $, one obtains the closed form as
\begin{IEEEeqnarray}{rCl}
	C_{12} & = & 
			\bigintssss_{\srv}
				 C_6(\srv) \mathcal{IG}^{\text{VB}}(\srv|\an, \bn) 
			d\srv 
			\nonumber \\
	& = & C_5 
		\bigintssss_{\srv}  
			\frac{\bn^{\an}}{\Gamma(\an)}
			(\srv)^{-\an-1-1}
			\exp\left\{ -\frac{\bn}{\srv} \right\} 
			d\srv \nonumber \\
	& \overset{I(\mathcal{IG})}{=} &
		 C_5\frac{\Gamma(\an + 1)}{\bn\Gamma(\an)}.
\end{IEEEeqnarray}
Using the exact same steps for $C_{10}(\srv)$, one finds
\begin{IEEEeqnarray}{rCl}
C_{13} & = & 
			\bigintssss_{\srv}
				 C_{10}(\srv) \mathcal{IG}^{\text{VB}}(\srv|\an, \bn) 
			d\srv 
			\nonumber \\
		& \overset{I(\mathcal{IG})}{=} &
		 C_9\frac{\Gamma(\an + 1)}{\bn\Gamma(\an)},
\end{IEEEeqnarray}
finally yielding
\begin{IEEEeqnarray}{rCl}
	Q_1 & = & C_1 - C_8 + C_{11} -C_{12} - C_{13} \nonumber \\
	& = &
		\log
			\left( 
				\dfrac{\bn^{\an}\Gamma(\az)}{\bz^{\az}\Gamma(\an)} 
			\right) + 	
		0.5\log\left| \Sz\Sninv \right|^{0.5}
		- \frac{1}{2}
				\text{tr}\left(I - \Szinv\Sn\right)
		+ (\an - \az)\left( \Psi(\an)- \log(\bn) \right) \nonumber \\
	&&\quad\quad
		- \left[ \frac{1}{2}\ben'\Sninv\ben -  \frac{1}{2}\bez'\Szinv\bez +(\bn - \bz) \right]
			\cdot \frac{\Gamma(\an + 1)}{\bn\Gamma(\an)} \nonumber \\
	&&\quad\quad
		- \left[
				\frac{1}{2}\ben'(\Sninv - \Szinv)\ben - 
				\ben'\left(\Sninv\ben - 	\Szinv\bez \right)
			\right]
			\cdot \frac{\Gamma(\an + 1)}{\bn\Gamma(\an)} \nonumber \\
	& = &
		\log
			\left( 
				\dfrac{\bn^{\an}\Gamma(\az)}{\bz^{\az}\Gamma(\an)} 
			\right) + 	
		0.5\log\left| \Sz\Sninv \right|^{0.5}
		- \frac{1}{2}
				\text{tr}\left(I - \Szinv\Sn\right)
		+ (\an - \az)\left( \Psi(\an)- \log(\bn) \right) \nonumber \\
	&&\quad\quad
		+ \left[ 
				\frac{1}{2} (\bez - \ben)'\Szinv(\bez - \ben) + 2(\bz - \bn)
			\right]
			\cdot \frac{\Gamma(\an + 1)}{\bn\Gamma(\an)} \\
\end{IEEEeqnarray}

\subsection{$Q_2$}
Noting that one can write $Q_2$ as
\begin{IEEEeqnarray}{rCl}
			& = & E_{\text{VB}}\left[ \sum_{i=1}^n  \loss(\berv, \srv| \*y_i) \right] \nonumber \\
		   & = & \bigintssss_{\berv, \srv}
		   			\left\{
		   				\sum_{i=1}^n  \left[ 
		   					\frac{1}{\beta} f(\*y_i|\berv, \srv)^{\beta} - 
		   					\frac{1}{1+\beta} \int_{\mathcal{Y}} f(\*y|\berv, \srv)^{1+\beta}d\*y
		   			\right] \NIGVB \right\} d\srv d\berv \nonumber \\
		    & = & \sum_{i=1}^n  \left[ 
		    			\bigintssss_{\berv, \srv}
		   				\left\{
		   					\frac{1}{\beta} f(\*y_i|\berv, \srv)^{\beta} - 
		   					\frac{1}{1+\beta} \int_{\mathcal{Y}} f(\*y|\berv, \srv)^{1+\beta}d\*y\right\} 
		   			 \NIGVB 
		   			 d\srv d\berv \right]. \label{Q2}
\end{IEEEeqnarray}
The last equation implies that it is sufficient to concern ourselves with the integral for a single term. To this end, observe that the likelihood for a single observation $\*y_i$ with regressor matrix $\*X_i$ is given by
\begin{IEEEeqnarray}{rCl}
	f(\*y_i|\berv, \srv) & = & \mathcal{N}(\*y_i |\*X_i'\berv,  \srv I_d),
\end{IEEEeqnarray}
 where $I_d$ is  the identity matrix of dimension $d$ .
Looking at the likelihood terms inside $\loss$,
 the $\beta$-exponentiated likelihood term can be rewritten as
\begin{IEEEeqnarray}{rCl}
	\frac{1}{\beta} f(\*y_i|\berv, \srv)^{\beta} 
	& = & 
		\underbrace{
			\frac{1}{\beta}(2\pi)^{-0.5d\beta}(\srv)^{-0.5d\beta}
		}_{ = D_1(\srv)}
		\cdot
			\exp\left\{ -\frac{\beta}{2\srv} 
					\left[ 
						(\*y_i - \*X_i'\berv)'(\*y_i - \*X_i'\berv)
					\right] 
			\right\} \nonumber \\
	& =  &
		D_1(\srv) \cdot
		\exp\left\{
			-\frac{\beta}{2\srv} 
				\left[
					\*y_i'\*y_i + \berv'
						\underbrace{(\*X_i\*X_i')}_{ = \Sciinv}
					\berv - 2(\*y_i'\*X_i)\berv
				\right]
		\right\} \nonumber \\
	& \overset{\blacksquare}{=}  & 
		D_1(\srv) \cdot
			\exp\left\{
				-\frac{1}{2\srv} 
					\left[
						\beta(\berv - \underbrace{\Sci(\*X_i'\*y_i)}_{= \beci})'\Sciinv(\berv - \beci) +
						\underbrace{
							 \beta\left[\*y_i'\*y_i  -(\*y_i\*X_i')\Sci(\*X_i\*y_i')\right]
						}_{ = D_{2,i} }
					\right]
			\right\} \nonumber \\
	& = &
		D_1(\srv) \cdot
			\exp\left\{
				-\frac{1}{2\srv} 
					\left[
						\underbrace{
							\beta(\berv -  \beci)'\Sciinv(\berv - \beci) 
						}_{ = D_{3,i}(\berv)}
						+D_{2,i}
					\right]
			\right\}		 \nonumber \\
	& = &
		D_1(\srv) \cdot
			\exp\left\{
				-\frac{1}{2\srv} 
					\left[
						 D_{3,i}(\berv)
						+D_{2,i}
					\right]
			\right\},  \label{f_i_term}
\end{IEEEeqnarray}
 while the integral is available in closed form as
\begin{IEEEeqnarray}{rCl}
	\frac{1}{1+\beta} \int_{\mathcal{Y}} f(\*y|\berv, \srv)^{1+\beta}d\*y & \overset{I(\mathcal{N})}{=}  &
		(\srv)^{-0.5p\beta}
			\underbrace{
				(2\pi)^{-0.5d\beta}(1+\beta)^{-0.5d-1}
			}_{ = D_4}
	\label{f_integral}
\end{IEEEeqnarray}
One can see a neat separation between terms involving $\srv$ and terms involving $\berv$ again, allowing us to rewrite the integral in equation \eqref{Q2} such as to exploit the conditional structure of the normal inverse-gamma distribution in Eqs. \eqref{f_integral}, \eqref{f_i_term}. Looking at integrating out $\srv$ from \eqref{f_i_term} first, note that
\begin{IEEEeqnarray}{rCl}
	L_{1} &=& 
		\bigintsss_{\srv} 
			\left\{ 
				\frac{1}{1+\beta} \int_{\mathcal{Y}} f(\*y|\berv, \srv)^{1+\beta}d\*y			
			\right\}
			\mathcal{IG}^{\text{VB}}(\srv|\an, \bn)d\srv \nonumber \\
	& = & D_4  
		\bigintsss_{\srv} 
			(\srv)^{-0.5d\beta - \an - 1} 
			\frac{\bn^{\an}}{\Gamma(\an)}
			\exp\left\{ -\frac{\bn}{\srv} \right\}
			d\srv \nonumber \\
	& \overset{I(\mathcal{N})}{=} &
		 D_4 \cdot \dfrac{\Gamma(\an + 0.5d\beta)}{\Gamma(\an)\bn^{0.5d\beta}}.
\end{IEEEeqnarray}
For the $\beta$-exponentiated likelihood term, one finds that
\begin{IEEEeqnarray}{rCl}
	L_{2,i} &=& \bigintssss_{\srv, \berv} 
				\frac{1}{\beta} f(\*y_i|\berv, \srv)^{\beta}  
				\NIGVB 
				d\srv d\berv \nonumber \\
	& = & \bigintssss_{\srv}
			D_1(\srv) \cdot \exp\left\{-\frac{1}{2\srv}D_{2,i} \right\}
			\underbrace{
		   		\left[
				\bigintssss_{\berv}
						\exp\left\{
							-\frac{1}{2\srv} 
						 			D_{3,i}(\berv)
						\right\}
					\mathcal{N}^{\text{VB}}(\berv|\ben, \srv\Sn)
					d\berv		
		   		\right]
		   	}_{ = D_{5,i}(\srv)}
		   	\mathcal{IG}^{\text{VB}}(\srv|\an, \bn)
		   	d\srv, \nonumber
\end{IEEEeqnarray}
where we have again exploited the conditional structure of our assumed posterior. The inner integral equals
\begin{IEEEeqnarray}{rCl}
	D_{5,i}(\srv) & = & 
	(2\pi)^{-0.5p}\left| \srv \Sn \right|^{-0.5}
		\underbrace{
		\bigints_{\berv}
						\exp\left\{
							-\frac{1}{2\srv} 
								\underbrace{
								\left[
						 			D_{3,i}(\berv) + (\berv - \ben)'\Sninv(\berv - \ben)
								\right]
								}_{ = D_{6,i}(\berv) }
						\right\}
			}_{ = D_{7,i}(\srv)},
\end{IEEEeqnarray}
indicating that the closed form for the integral is available if one rewrites it as a normal density. To this end, one can use completion of squares to rewrite
\begin{IEEEeqnarray}{rCl}
	D_{6,i}(\berv) 
	& = & 
		\beta(\berv -  \beci)'\Sciinv(\berv - \beci)  + 
		(\berv - \ben)'\Sninv(\berv - \ben) 
		\nonumber \\
	& = &
		\berv'
			\underbrace{
				\left[ \Sninv + \beta \Sciinv \right]
			}_{ = \Stiinv }
		\berv -
		2\left[ \bn'\Sninv + \beta\beci'\Sciinv \right]\berv + 
		\left[ \ben'\Sninv\ben +  \beta \beci'\Sciinv\beci\right] 
		\nonumber \\
	& \overset{\blacksquare}{=}  & 
		\left(\berv - 
			\underbrace{
				\Sti \left[ \Sninv\bn + \beta\Sciinv\beci \right]
			}_{ = \beti}
		\right)'
		\Stiinv (\berv - \beti) + \nonumber \\
	&& \quad\quad\quad
		\underbrace{
			\ben'\Sninv\ben +  \beta \beci'\Sciinv\beci -
			\left(\Sninv\ben + \beta\Sciinv\beci\right)' \Sti 
				\left(\Sninv\ben + \beta\Sciinv\beci\right)
		}_{ = D_{8,i}} \nonumber \\
	& = &
		(\berv - \beti)' \Stiinv (\berv - \beti) + D_{8,i},
\end{IEEEeqnarray}
which then allows integrating out $\berv$ from $D_{7,i}(\srv)$ using the density of a normal random variable:
\begin{IEEEeqnarray}{rCl}
	D_{7,i}(\srv) & =  & 
		\exp\left\{-\frac{1}{2\srv}D_{8,i}\right\} 
			\bigintsss_{\berv}
				\exp\left\{ -\frac{1}{2\srv}(\berv - \beti)' \Stiinv (\berv - \beti)   \right\}
			d \berv 
		\nonumber \\
	& \overset{I(\mathcal{N})}{=} &
		\exp\left\{-\frac{1}{2\srv}D_{8,i}\right\} 
		(2\pi)^{0.5p} | \srv\Sti |^{0.5},
\end{IEEEeqnarray}
so we can finally rewrite the entire integral as
\begin{IEEEeqnarray}{rCl}
	D_{5,i}(\srv) & = & 
			 |\Sn^{-1}\Sti |^{0.5}
		 \exp\left\{-\frac{1}{2\srv}D_{8,i}\right\}, 
\end{IEEEeqnarray}
which enables rewriting $L_{2,i}$ as
\begin{IEEEeqnarray}{rCl}
	L_{2,i} & = & 
		\underbrace{
		\frac{1}{\beta}(2\pi)^{-0.5d\beta}
		|\Sn^{-1}\Sti |^{0.5}	
		}_{ = D_{9,i} }
			\bigintsss_{\srv}
				(\srv)^{-0.5d\beta}
				\exp\left\{-\frac{1}{\srv}\cdot \frac{1}{2}\left[D_{2,i} + D_{8,i} \right] \right\}			
				\mathcal{IG}^{\text{VB}}(\srv|\an, \bn)
		   	d\srv \nonumber \\
	& \overset{I(\mathcal{IG})}{=} &
			\dfrac{
				D_{9,i} \cdot
				\Gamma(\an + 0.5d\beta) \cdot
				\bn^{\an} 
			}{
				\Gamma(\an)  \cdot
				\left[ 
					\bn +
					0.5(D_{2,i} + D_{7,i})
				\right]^{(\an + 0.5d\beta)}
			},
\end{IEEEeqnarray}
finally implying that one may write
\begin{IEEEeqnarray}{rCl}
	Q_2	& =  & 
				\sum_{i=1}^nL_{2,i} - n L_1 
				\nonumber \\
			& = &
				\sum_{i=1}^n
					\left\{ 
						\dfrac{
							D_{9,i} \cdot
							\Gamma(\an + 0.5d\beta) \cdot
							\bn^{\an} 
						}{
							\Gamma(\an)  \cdot
							\left[ 
								\bn +
								0.5(D_{2,i} + D_{8,i})
							\right]^{(\an + 0.5d\beta)}
						}
					\right\} 
				- nD_4 \cdot 
					\dfrac{
						\Gamma(\an + 0.5d\beta)
					}{
						\Gamma(\an)\bn^{0.5d\beta}
					} 
					\nonumber \\
			& = &
				\sum_{i=1}^n
					\left\{ 
						\dfrac{
							\frac{1}{\beta}
								(2\pi)^{-0.5d\beta}
								\left|\Sn^{-1}
									\left[ \Sninv + \beta(\*X_i\*X_i) \right]^{-1}
								\right|^{0.5}	 \cdot
							\Gamma(\an + 0.5d\beta) \cdot
							\bn^{\an} 
						}{
							\Gamma(\an)  \cdot
							\left[ 
								\bn +
								0.5\left( 
										D_{2,i} + D_{8,i}
									\right)
							\right]^{(\an + 0.5d\beta)}
						}
					\right\} 
				\nonumber \\
				&&\quad\quad\quad
				-n \cdot 
					\dfrac{
						(2\pi)^{-0.5d\beta}(1+\beta)^{-0.5d-1} \cdot
						\Gamma(\an + 0.5d\beta)
					}{
						\Gamma(\an)\bn^{0.5d\beta}
					}
					\nonumber.
\end{IEEEeqnarray}
We further simplify this expression by observing that 
\begin{IEEEeqnarray}{rCl}
	D_{2,i} + D_{8,i} & = &
		\beta\left[\*y_i'\*y_i  -(\*y_i\*X_i')\Sci(\*X_i\*y_i')\right] + 
									\ben'\Sninv\ben +  \beta \beci'\Sciinv\beci 
									\nonumber \\
									&& \quad\quad
									- \left( \Sninv\ben + \beta\Sciinv\beci\right)' \Sti 
									\left( \Sninv\ben + \beta\Sciinv\beci\right)
									\nonumber \\
	& = &
		\beta\*y_i'\*y_i  - \beta(\*y_i\*X_i')\Sci(\*X_i\*y_i') + 
			\ben'\Sninv\ben +
			\beta(\*y_i\*X_i')\Sci(\*X_i\*y_i')  			\nonumber \\
	&& \quad\quad
		- \left( \Sninv\ben + \beta(\*X_i'\*y_i)\right)' \Sti 
									\left( \Sninv\ben + \beta(\*X_i'\*y_i)\right)
									\nonumber \\
	& = & \beta\*y_i'\*y_i + \ben'\Sninv\ben - \left( \Sninv\ben + \beta(\*X_i'\*y_i)\right)'	
									\left[ \Sninv + \beta(\*X_i\*X_i) \right]^{-1}
									\left( \Sninv\ben + \beta(\*X_i'\*y_i)\right)
									\nonumber,
\end{IEEEeqnarray}
leaving us with
\begin{IEEEeqnarray}{lCl}
		\hspace*{-1cm} Q_2	= 
			\dfrac{
				\Gamma(\an + 0.5d\beta) \cdot
							\bn^{\an} 
							\cdot 
				|\Sn^{-1}|^{0.5}
			}{
				\beta (2\pi)^{0.5d\beta}\Gamma(\an)
			} \times \nonumber \\[10pt]
		\hspace*{-1cm} 
				\sum_{i=1}^n
					\left\{ 
						\dfrac{
								\left|
									\left[ \Sninv + \beta(\*X_i\*X_i) \right]^{-1}
								\right|^{0.5}
						}{
							\left[ 
								\bn +
								0.5\left( 
										\beta\*y_i'\*y_i + \ben'\Sninv\ben - \left( \Sninv\ben + 	
											\beta(\*X_i'\*y_i)\right)'	
									\left[ \Sninv + \beta(\*X_i\*X_i) \right]^{-1}
									\left( \Sninv\ben + \beta(\*X_i'\*y_i)\right)
									\right)
							\right]^{(\an + 0.5d\beta)}
						}
					\right\} 
				\nonumber \\[10pt]
				\hspace*{-1cm} \quad\quad\quad
				-n \cdot 
					\dfrac{
						\Gamma(\an + 0.5d\beta)
					}{
						\Gamma(\an)\bn^{0.5d\beta}
						(2\pi)^{0.5d\beta}
						(1+\beta)^{0.5d+1}
					}
					. \nonumber
\end{IEEEeqnarray}

\subsection{ELBO}

Putting together the results of the two previous sections, the ELBO is obtained as
\begin{IEEEeqnarray}{rCl}
	ELBO & = & -Q_1 + Q_2 \nonumber \\
		& = &
		-\log
			\left( 
				\dfrac{\bn^{\an}\Gamma(\az)}{\bz^{\az}\Gamma(\an)} 
			\right) -
		0.5\log\left| \Sz\Sninv \right|
		+ \frac{1}{2}
				\text{tr}\left(I - \Szinv\Sn\right)
		- (\an - \az)\left( \Psi(\an)- \log(\bn) \right) \nonumber \\
	&&\quad
		- \left[ 
				\frac{1}{2} (\bez - \ben)'\Szinv(\bez - \ben) + (\bz - \bn)
			\right]
			\cdot \frac{\Gamma(\an + 1)}{\bn\Gamma(\an)} \nonumber \\
	&&\quad
		+ \dfrac{
				\Gamma(\an + 0.5d\beta) \cdot
							\bn^{\an} 
							\cdot 
				|\Sn^{-1}|^{0.5}
			}{
				\beta (2\pi)^{0.5d\beta}\Gamma(\an)
			} \times \nonumber \\[10pt]
	&&\hspace*{-3cm}
		\sum_{i=1}^n
					\left\{ 
						\frac{
								\Rndet^{-0.5}
						}{
							\left[ 
								\bn +
								0.5\left( 
										\beta\*y_i'\*y_i + \ben'\Sninv\ben - \left( \Sninv\ben + 	
											\beta(\*X_i'\*y_i)\right)'	
									\left[ \Sninv + \beta(\*X_i'\*X_i) \right]^{-1}
									\left( \Sninv\ben + \beta(\*X_i'\*y_i)\right)
									\right)
							\right]^{(\an + 0.5d\beta)}
						}
					\right\}  \nonumber \\[10pt]
		&&\quad\quad 
			- n \cdot 
					\dfrac{
						\Gamma(\an + 0.5d\beta)
					}{
						\Gamma(\an)\bn^{0.5d\beta}
						(2\pi)^{0.5d\beta}
						(1+\beta)^{0.5d+1}
					}		
\end{IEEEeqnarray}

\subsection{Differentiation}

In this section, we take derivatives of the ELBO with respect to each variational parameter, i.e. $\an, \bn, \ben, \Sn$. Observing that differentiation with respect to $\Sninv$ is easier than with respect to $\Sn$, parametrize the optimization using the Cholesky decomposition, i.e.  $\Sninv = \L\L'$, where $L$ is a lower triangular matrix and is unique if $\Sn$ (equivalently $\Sninv$) is positive definite\footnote{Note that $\L$ need not be unique if $\Sn$ is positive semi-definite, but this is of no concern for us here: Since we implicitly impose  that $\Sn$ is non-singular (so that $\Sninv$ is unique and well-defined), all covariance matrices $\Sn$ considered have to be positive definite.}. 

\subsubsection{Derivative with respect to $L$}

In what follows, we differentiate the ELBO term by term with respect to the $p(p-1)\frac{1}{2}$ entries in the lower triangular part of $\L$ that can be summarized in the vector $\text{vech}\left(\L \right)$.
To this end, define 
\begin{IEEEeqnarray}{rCl}
	E_1 & = & -0.5\log\left| \Sz\Sninv \right| + 
						\frac{1}{2}\text{tr}\left(I - \Szinv\Sn\right) \label{E_first}  \\
	E_2 & = & 	\underbrace{
						\dfrac{
							\Gamma(\an + 0.5d\beta) \cdot \bn^{\an} 
						}{
								\beta (2\pi)^{0.5d\beta}\Gamma(\an)
						}}_{ = F} 
						|\Sn^{-1}|^{0.5}  \\			
	E_{3,i} & = &	\Rndet^{-0.5}  \\
	E_4 & = &	 \ben'\Sninv\ben  \\
	E_{5,i} & = &	-\ben'\Sninv\Rninv\Sninv\ben  \\
	E_{6,i} & = & -\beta^2(\*y_i'\*X_i)\Rninv(\*X_i'\*y_i),  \\
	E_{7,i} & = & -2\beta\ben'\Sninv\Rninv(\*X_i'\*y_i). \label{E_last}
\end{IEEEeqnarray}
Obtaining the derivative of the ELBO is equivalent to obtaining the derivatives of these newly defined quantities, as 
\begin{IEEEeqnarray}{rCl}
	\DL\left\{ELBO\right\} & = & 
		\DL\left\{ E_1 \right\}  \nonumber \\
		&& + \DL\left\{ E_2 \right\} \cdot 
			\sum_{i=1}^n 
				\left\{ 
					\dfrac{
						E_{3,i}
					}{
						\left[ \bn + 
							0.5\left(
								\beta\*y_i'\*y_i + E_4 + E_{5,i}+ E_{6,i} + E_{7,i}
							\right)
						\right]^{\an + 0.5d\beta}
					}
				\right\} \nonumber \\[10pt]
	&& + E_2 \cdot 
		\sum_{i=1}^n
			\left\{ 
				\dfrac{\DL\left\{ E_{3,i}\right\}}
				{ 
					\left[ \bn + 
							0.5\left(
								\beta\*y_i'\*y_i + E_4 + E_{5,i}+ E_{6,i} + E_{7,i}
							\right)
						\right]^{\an + 0.5d\beta}				
				}
			\right\} \nonumber \\
	&& + E_2 \cdot 
		\sum_{i=1}^n
			\left\{ 
				{\ E_{3,i}	}
				\cdot 
				\DL\left\{
					\left[ \bn + 
							0.5\left(
								\beta\*y_i'\*y_i + E_4 + E_{5,i}+ E_{6,i} + E_{7,i}
							\right)
						\right]^{-\an - 0.5d\beta}	
					\right\}		
			\right\},
\end{IEEEeqnarray}
where the chain and sum rule imply that
\begin{IEEEeqnarray}{rCl}
	&& \DL\left\{
					\left[ \bn + 
							0.5\left(
								\beta\*y_i'\*y_i + E_4 + E_{5,i}+ E_{6,i} + E_{7,i}
							\right)
						\right]^{-\an -0.5d\beta}	
					\right\}	\nonumber \\
	& = &
		\left( -\an -0.5d\beta \right) \cdot
			\left[ \bn + 
							0.5\left(
								\beta\*y_i'\*y_i + E_4 + E_{5,i}+ E_{6,i} + E_{7,i}
							\right)
			\right]^{-\an -0.5d\beta -1} \times \nonumber \\
	&&	\quad\quad
			0.5\cdot \DL \left\{ 
								E_4 + E_{5,i}+ E_{6,i} + E_{7,i}
						\right\},
\end{IEEEeqnarray}
For convenience and simplified notation when taking the derivatives of the expressions defined in Eqs \eqref{E_first} - \eqref{E_last}, also define the following matrices:
\begin{IEEEeqnarray}{rCl}
	\R &=& \Rnbr \\
	\B &=& \ben\ben'.
\end{IEEEeqnarray}
Define also the following symbols to mark operations used in the derivations:
\begin{itemize}
	\item[$ \p$] Switching from differential notation $\p \L$ to the derivative  $\DL$; 
	\item[$\tr$] Properties of the trace like invariance under cyclic permutations, invariance under the transpose, additivity, and the fact that for $c$ a scalar, $\tr(c) = c$.
\end{itemize}
Note than when the differential operator $\p$ is used, its scope is always limited to the next term only, unless brackets are used. Hence $\p \L\L'$ uses the differential only with respect to $\L$, while $\p\left(\L\L'\right)^{-1}$ uses it with respect to the entire expression $\left(\L\L'\right)^{-1}$. It is also worth noting that $\p\L' = \left(\p\L\right)'$ for any matrix $\L$, as this will be used in conjunction with the transpose invariance of the trace throughout to simplify terms. 
Using these symbols and the differential notation, proceed by noting the following:
\begin{IEEEeqnarray}{rCl}
	\p (\L\L') & = & \p\R  = \p \L\L' + \L\p \L' =  \p \L\L' + \L\p \L' = \p \L\L' +\left(\p \L\L'  \right)' \\
	\p (\L\L')^{-1} & = & 
			-(\L\L')^{-1}\left[ 
				\p(\L\L')
			\right] (\L\L')^{-1} \\
	\p |\L\L'| 
		& =& |\L\L'| \cdot 
		\tr\left( 
			(\L\L')^{-1}\left[ \p \L\L' +\left(\p \L\L'  \right)' \right]			
		\right) \nonumber \\
		& \overset{\tr}{=}  &
		 2|\L\L'| 
		\cdot\tr\left( 
			\L'(\L\L')^{-1}\p \L	
		\right) \\
	\p \R^{-1} & = & -\R^{-1}\p \R \R^{-1}  
		= -\R^{-1}\p \L \L' \R^{-1} - \left[ \R^{-1}\p \L \L' \R^{-1}\right]'.
\end{IEEEeqnarray}
With this in place, the derivatives of the quantities defined before are obtained as
{\allowdisplaybreaks
\begin{IEEEeqnarray}{rCl}
	\p E_1 & = & 	-\frac{1}{2}  \p\left\{	\log| \Sz| + \log| \L\L' | \right\} 
						- \frac{1}{2}  \p \left\{	 \tr\left(\Szinv\Sn \right) \right\} \nonumber \\
			& = & -\frac{1}{2} \cdot |\L\L'|^{-1} \cdot \p| \L\L' | 
					- \frac{1}{2}  \tr\left(\Szinv \p\left(\L\L'\right)^{-1} \right) \nonumber \\
			& = & - \frac{1}{2}\tr\left(\L'(\L\L')^{-1} \p \L \right)
						+ \frac{1}{2}  \tr\left(\Szinv 
							(\L\L')^{-1}\left[ \p(\L\L') \right] (\L\L')^{-1} 
						\right) \nonumber \\
			& \overset{\tr}{=} &
				-  \tr\left(\L'(\L\L')^{-1} \p \L \right)
				+ \tr\left(\L'\left(\L\L'\right)^{-1}\Szinv\left(\L\L'\right)^{-1}\p\L \right) \nonumber \\
	\p E_2 & = & F \cdot \p |\L\L'|^{0.5} \nonumber \\
			& = &  \frac{F}{2} \cdot |\L\L'|^{-0.5} \cdot 2|\L\L'|
						\cdot\tr\left( 
								\L'(\L\L')^{-1}\p \L	
						\right) \nonumber \\
			& = & F\cdot |\L\L'|^{0.5} \tr\left( 
								\L'(\L\L')^{-1}\p \L	
						\right) \\
	\p E_{3,i} & = & 
		\p \R^{-0.5} = -\frac{1}{2}|\R|^{-1.5}\p\R  \nonumber \\
		&=& -\frac{1}{2}|\R|^{-0.5}\tr\left(\R^{-1} \p\left(\L\L' \right)\right) \nonumber \\
		&\overset{\tr}{=}& - |\R|^{-0.5}\tr\left(\L'\R^{-1}\p\L\right)\\
	\p E_4 &  \overset{\tr}{=} & \tr\left(\ben'\p(\L\L')\ben\right) \nonumber \\
			& = & \tr\left(\ben'\left[\p\L\L' + \L\p\L'\right]\ben\right) \nonumber \\
			&  \overset{\tr}{=} & 2\cdot \tr\left( \L'\B \p\L \right) \\
	\p E_{5,i} &  \overset{\tr}{=}  &
			-\tr\left(
				\ben'\p\left(\L\L'\right)\R^{-1}\left(\L\L'\right)\ben
			\right)
			-\tr\left(
				\ben'\left(\L\L'\right)\p\R^{-1}\left(\L\L'\right)\ben
			\right) \nonumber \\
			&&
			-\tr\left(
				\ben'\left(\L\L'\right)\R^{-1}\p\left(\L\L'\right)\ben
			\right) \nonumber \\
		& \overset{\tr}{=}  &
			-2\cdot \tr\left(
				\ben' \p \L\L'\R^{-1}\left(\L\L'\right)\ben
			\right)
			+2\cdot \tr\left(
				\ben'\left(\L\L'\right)\R^{-1}\p\L\L'\R^{-1}\left(\L\L'\right)\ben
			\right) \nonumber \\
			&&
			-2\cdot \tr\left(
				\ben'\left(\L\L'\right)\R^{-1}\p\L\L'\ben
			\right) \nonumber \\
		& \overset{\tr}{=}  &
			-2\cdot \tr\left(
				\L'\R^{-1}\left(\L\L'\right)\B \p \L
			\right)
			+2\cdot \tr\left(
				\L'\R^{-1}\left(\L\L'\right)\B\left(\L\L'\right)\R^{-1}\p\L
			\right) \nonumber \\
			&&
			-2\cdot \tr\left(
				\L'\B\left(\L\L'\right)\R^{-1}\p\L
			\right) \nonumber \\		
	\p E_{6,i} & \overset{\tr}{=} & -\beta^2 \tr\left( 
			\left( \*y_i' \*X_i \right) \p \R^{-1} \left( \*X_i' \*y_i \right)
		 \right) \nonumber \\
		& \overset{\tr}{=}  &  2\beta^2 \tr\left( 
					\left( \*y_i' \*X_i \right) \R^{-1}\p\L\L'\R^{-1} \left( \*X_i' \*y_i \right)
			\right) \nonumber \\
		& \overset{\tr}{=}  &  2\beta^2 \tr\left( 
					\L'\R^{-1} \left( \*X_i' \*y_i \right)\left( \*y_i' \*X_i \right) \R^{-1}\p\L
			\right)  \\
	\p E_{7,i} &  \overset{\tr}{=} & -2\beta \cdot \left[
		\tr\left(
			\ben'\p\left(\L\L'\right)\R^{-1}\left(\*X_i\*y_i\right)
		\right) +
		\tr\left(
			\ben'\left(\L\L'\right)\p\R^{-1}\left(\*X_i\*y_i\right)
		\right)	
	\right] \nonumber \\
	& \overset{\tr}{=} &
	-2\beta \cdot \bigg[
		\tr\left(
			\ben'\p\L\L' \R^{-1}\left(\*X_i\*y_i\right)
		\right) 
		+\tr\left(
			\ben'\L\p\L' \R^{-1}\left(\*X_i\*y_i\right)
		\right) \nonumber \\
		&&
		-\tr\left(
			\ben'\left(\L\L'\right)\R^{-1}\p\L\L'\R^{-1}\left(\*X_i\*y_i\right)
		\right)	
		-\tr\left(
			\ben'\left(\L\L'\right)\R^{-1}\L\p\L'\R^{-1}\left(\*X_i\*y_i\right)
		\right)	
	\bigg] \nonumber \\
	& \overset{\tr}{=} &
	-2\beta \cdot \bigg[
		\tr\left(
			\L' \R^{-1}\left(\*X_i\*y_i\right)\ben'\p\L
		\right) 
		+\tr\left(
			\L' \ben\left(\*y_i'\*X_i\right)\R^{-1}\p\L
		\right) \nonumber \\
		&& \hspace*{-0.35cm}
		-\tr\left(
			\L'\R^{-1}\left(\*X_i\*y_i\right)\ben'\left(\L\L'\right)\R^{-1}\p\L
		\right)	
		-\tr\left(
			\L'\R^{-1}\left(\L\L'\right)\ben\left(\*y_i'\*X_i\right)\R^{-1}\p\L
		\right)	
	\bigg] 
\end{IEEEeqnarray}	
}			
This can now be converted into derivative notation and simplified. To this end, first note that for any $p\times\p$ matrix $A$ which is not a function of $\L$,
\begin{IEEEeqnarray}{rCl}
	\tr(A d\L) & = & \sum_{i=1}^p A_{1i} dL_{i1} + \sum_{i=2}^p A_{2i} dL_{i2} + \dots = \sum_{j=1}^p\left\{ \sum_{i=j}^p A_{ji} dL_{ji} \right\},
\end{IEEEeqnarray}
implying in particular that
\begin{IEEEeqnarray}{rCl}
	\DL \tr(A d\L) & = & \v(A^T)
\end{IEEEeqnarray}
and use this by defining $\v^T(A) = \v(A^T)$ to note that
{\allowdisplaybreaks
\begin{IEEEeqnarray}{rCl}
		\DL E_1	& \overset{\p}{=} &
				\v^T\left(-\left[\L'(\L\L')^{-1} \right] 
				+ \left[ \L'\left(\L\L'\right)^{-1}\Szinv\left(\L\L'\right)^{-1}\right]\right) \nonumber \\
			& = &
				\v^T\left(\L'\left(\L\L'\right)^{-1}\left[ \Szinv\left(\L\L' \right)^{-1} -  I_p \right]\right)  \nonumber \\
			& = & \v^T\left(\L^{-1}\left[ \Szinv\left(\L\L' \right)^{-1} - I_p \right]\right) \nonumber \\
			& = & \v\left(\left[ \left(\L\L' \right)^{-1}\Szinv -  I_p \right]\L^{-T}\right) \\
		\DL E_2 & \overset{\p}{=} & F\cdot |\L\L'|^{0.5}\cdot \v^T\left( \L'(\L\L')^{-1}\right) \nonumber \\
		& = & F\cdot |\L\L'|^{0.5}\cdot\v\left( \L^{-T}\right)  \\
		\DL E_{3,i} & \overset{\p}{=} & -|\R|^{-0.5}\cdot \v\left(\R^{-1}\L\right) \\
		\DL E_{4} & \overset{\p}{=} & 2\cdot \v\left( \B \L \right)\\
		\DL E_{5,i} & \overset{\p}{=} &
			\v^T\left( -2
				\L'\R^{-1}\left(\L\L'\right)\B 
			+2
				\L'\R^{-1}\left(\L\L'\right)\B\left(\L\L'\right)\R^{-1} 
			-2
				\L'\B\left(\L\L'\right)\R^{-1} \right) \nonumber \\
			& = &
			2\cdot\v^T\left( \left[ 
				\L'\R^{-1}\left(\L\L'\right)\B\left[ \left(\L\L'\right)\R^{-1} - I_p \right] 
			\right]
			- \left[ \L' \B\left(\L\L'\right) \R^{-1} \right]\right) \nonumber\\
			& = &
			2\cdot\v^T\left(
				\L' \left[
					\R^{-1}\left(\L\L'\right)\B
					\left[
						 \left(\L\L'\right)\R^{-1} - I_p
					\right] 
				 - \B\left(\L\L'\right) \R^{-1}
				\right]\right) \nonumber \\
			& = &
			2\cdot\v\left(
				\left[
					\left[
						 \R^{-1}\left(\L\L'\right) - I_p
					\right] 
					\B\left(\L\L'\right) \R^{-1}
				 - \R^{-1}\left(\L\L'\right)\B
				\right]\L \right) \\
		\DL E_{6,i} &\overset{\p}{=} & 
			2\beta^2 \cdot \v\left(\R^{-1} \left( \*X_i' \*y_i \right)\left( \*y_i' \*X_i \right) \R^{-1}\L\right) \\
		\DL E_{7,i} & \overset{\p}{=} &
			-2\beta \cdot \v^T\bigg(
				\L' \R^{-1}\left(\*X_i\*y_i\right)\ben'
				+\L' \ben\left(\*y_i'\*X_i\right)\R^{-1} \nonumber \\
		&& 
				-\L'\R^{-1}\left(\*X_i\*y_i\right)\ben'\left(\L\L'\right)\R^{-1}
				-\L'\R^{-1}\left(\L\L'\right)\ben\left(\*y_i'\*X_i\right)\R^{-1}	
			\bigg) \nonumber \\
		& = & -2\beta\cdot \v^T \bigg(
				\L'\R^{-1}\left( \*X_i'\*y_i \right) \ben' \left[ I_p - \left( \L\L'\right)\R^{-1} \right] 
				\nonumber \\ && \hspace*{2cm}
				+ \left[I_p - \L'\R^{-1}\L \right]\L'\ben\left(\*y_i'\*X_i\right)\R^{-1}
			\bigg)  \nonumber \\
		& = & -2\beta\cdot \v \bigg(
		\left[ I_p - \R^{-1}\left( \L\L'\right) \right] 
				\ben \left( \*y_i'\*X_i \right) \R^{-1}\L
				\nonumber \\ && \hspace*{2cm}
				+ \R^{-1}\left( \*X_i'\*y_i \right) \ben'\L
				\left[I_p - \L'\R^{-1}\L \right]
			\bigg)
\end{IEEEeqnarray}
}

\subsubsection{Derivative with respect to $\ben$}

Differentiating with respect to $\ben$ is trivial. One proceeds by the same logic as in the section before, to which end one additionally needs to define the new term
\begin{IEEEeqnarray}{rCl}
	E_8 & = & -\frac{1}{2}\left[ 
						(\bez - \ben)'\Szinv(\bez - \ben) + 2(\bz - \bn)
					\right] \cdot \frac{\Gamma(\an + 1)}{\bn \Gamma(\an)},
\end{IEEEeqnarray}
allowing us to write
\begin{IEEEeqnarray}{rCl}
	\Dben\left\{ELBO\right\} & = & 
		\Dben\left\{ E_8 \right\} + \nonumber \\
		&& \hspace{-1.5cm} E_2 \cdot 
		\sum_{i=1}^n
			\left\{ 
				{\ E_{3,i}	} \cdot 				
					\Dben\left\{
					\left[ \bn + 
							0.5\left(
								\beta\*y_i'\*y_i + E_4 + E_{5,i}+ E_{6,i} + E_{7,i}
							\right)
						\right]^{-\an - 0.5d\beta}	
					\right\}			
			\right\},
\end{IEEEeqnarray}
where 
\begin{IEEEeqnarray}{rCl}
	&& \Dben\left\{
					\left[ \bn + 
							0.5\left(
								\beta\*y_i'\*y_i + E_4 + E_{5,i}+ E_{6,i} + E_{7,i}
							\right)
						\right]^{-\an -0.5d\beta}	
					\right\}	\nonumber \\
	& = &
		\left( -\an -0.5d\beta \right) \cdot
			\left[ \bn + 
							0.5\left(
								\beta\*y_i'\*y_i + E_4 + E_{5,i}+ E_{6,i} + E_{7,i}
							\right)
			\right]^{-\an -0.5d\beta -1} \times \nonumber \\
	&&	\quad\quad
			0.5\cdot \Dben \left\{ 
								E_4 + E_{5,i} + E_{7,i}
						\right\},
\end{IEEEeqnarray}
so that obtaining the derivative is achieved by finding $\Dben E_4, \Dben E_{5,i}, \Dben E_{7,i}$ and $\Dben E_8$:
\begin{IEEEeqnarray}{rCl}
	\Dben E_{4} & = &
		2\cdot \ben'\Sninv \\
	\Dben E_{5,i} & = &
		-2 \cdot \ben'\Sninv\R^{-1}\Sninv \\
	\Dben E_{7,i} & = &
		-2\beta\cdot  \left(\*y_i'\*X_i\right) \R^{-1}  \Sninv\\
	\Dben E_8 & = & 
		-\frac{1}{2}\cdot
		\frac{\Gamma(\an + 1)}{\bn\Gamma(\an)}
		\left[ 
			 \Dben\left(\ben'\Szinv\ben \right) 
			 - 2\Dben\left(\ben  \Szinv\bez \right)
		\right] \nonumber \\
	& = &
		-\frac{1}{2}\cdot
		\frac{\Gamma(\an + 1)}{\bn\Gamma(\an)}
		\left[ 
			2\ben'\Szinv
			 - 2 \bez'\Szinv
		\right] \nonumber \\
	& = &
		\frac{\Gamma(\an + 1)}{\bn\Gamma(\an)}
		\left[ 
			\left( \bez- \ben\right)'
			\Szinv
		\right] 
\end{IEEEeqnarray}

\subsubsection{Derivative with respect to $\an$}
We proceed again by the same logic. Define
\begin{IEEEeqnarray}{rCl}
	E_9 & = & -\log
			\left( 
				\dfrac{\bn^{\an}\Gamma(\az)}{\bz^{\az}\Gamma(\an)} 
			\right) \\
	E_{10} & = &- (\an - \az)\left( \Psi(\an)- \log(\bn) \right)  \\
	E_{11} & = & - n \cdot 
					\dfrac{
						\Gamma(\an + 0.5d\beta)
					}{
						\Gamma(\an)\bn^{0.5d\beta}
						(2\pi)^{0.5d\beta}
						(1+\beta)^{0.5d+1}
					}.			
\end{IEEEeqnarray}
Use this to write
\begin{IEEEeqnarray}{rCl}
	\Dan\left\{ELBO\right\} & = & 
		\Dan\left\{ E_8 \right\} + 
		\Dan\left\{ E_9 \right\} + 
		\Dan\left\{ E_{10} \right\} + 
		\Dan\left\{ E_{11} \right\} + \nonumber \\ 
	&&  \hspace{-1.25cm} +\Dan\left\{ E_2 \right\}\sum_{i=1}^n 
				\left\{ 
					\dfrac{
						E_{3,i}
					}{
						\left[ \bn + 
							0.5\left(
								\beta\*y_i'\*y_i + E_4 + E_{5,i}+ E_{6,i} + E_{7,i}
							\right)
						\right]^{\an + 0.5d\beta}
					}
				\right\} \nonumber \\[10pt]
	&& \hspace{-1.25cm} + E_2 \cdot 
		\sum_{i=1}^n
			\left\{ 
				{ E_{3,i}	}\cdot
					\Dan\left\{
					\left[ \bn + 
							0.5\left(
								\beta\*y_i'\*y_i + E_4 + E_{5,i}+ E_{6,i} + E_{7,i}
							\right)
						\right]^{-\an - 0.5d\beta}	
					\right\}		
			\right\},
\end{IEEEeqnarray}
where for $\an$, the inner term equals
\begin{IEEEeqnarray}{rCl}
	&&\hspace*{-1.25cm} \Dan\left\{
					\left[ 
					\underbrace{\bn + 
							0.5\left(
								\beta\*y_i'\*y_i + E_4 + E_{5,i}+ E_{6,i} + E_{7,i}
							\right)
					}_{ = K }
					\right]^{-\an - 0.5d\beta}						
				\right\}	
	=
		-\log\left(K \right) \cdot K^{-\an -0.5d\beta},
\end{IEEEeqnarray}
so that the differentiation with respect to $\an$ requires obtaining the following terms:
\begin{IEEEeqnarray}{rCl}
	\Dan E_2 
		& = &
		\frac{|\Sn^{-1}|^{0.5}}{\beta(2\pi)^{0.5d\beta}}
		\bigg[
				\frac{
					\Dan\left\{ \Gamma(\an+0.5d\beta)\right\}\bn^{\an}
				}{ 
					\Gamma(\an)
				}
				 + \frac{
				 		\Dan\left\{ \bn^{\an} \right\}
				 		\Gamma\left(\an + 0.5d\beta\right)
				 	}{
				 		\Gamma(\an)
				 	}
				 	\nonumber \\
				 && +\Dan\left\{ \Gamma(\an)^{-1} \right\} \cdot 
				 	\bn^{\an}\Gamma(\an + 0.5d\beta) 
		\bigg]
		\nonumber \\
		& = &
		\frac{
			|\Sn^{-1}|^{0.5}\bn^{\an}\Gamma(\an + 0.5d\beta)
		}{
			\beta(2\pi)^{0.5d\beta}\Gamma(\an)
		}
		\left[
					\Psi(\an+ 0.5d\beta)
				 	+ \log(\bn)
				 	- \Psi(\an)
		\right]  \\	
	\Dan  E_8 & = & -\frac{1}{2}\left[ 
						(\bez - \ben)'\Szinv(\bez - \ben) + 2(\bz - \bn)
					\right] \cdot 
				\left[ 
					\frac{\Dan\left\{ \Gamma(\an + 1) \right\}}
							{\bn\Gamma(\an)}
					- \frac{\Dan\left\{ \Gamma(\an) \right\}\Gamma(\an+1)}	
							{\Gamma(\an)^2\bn} 
				\right] \nonumber \\
		& = &-\frac{1}{2}
						\left[
							(\bez - \ben)'\Szinv(\bez - \ben) + 2(\bz - \bn)
						\right]
				\cdot \frac{\Gamma(\an + 1)}{ \bn\Gamma(\an)} \cdot 
				\left[ 
					{\Psi(\an+1)}
					- {\Psi(\an)}
				\right]\\
	\Dan E_9 & = &
		-\Dan\left\{ 
			\an\log(\bn) \right\} 
			+ \Dan\left\{  \log\left(\Gamma(\an)\right)
		\right\} \nonumber \\
		& = & -\log(\bn) + \Psi(\an) \\
	\Dan E_{10} & = &
		\Dan\left\{ 
			\an\log(\bn) \right\} 
			- \Dan\left\{ \left(\an - a_0 \right) \Psi(\an)
		\right\} \nonumber \\
		&= & \log(\bn) - \Psi(\an) - \left(\an - a_0\right)\Psi^{(1)}(\an) \\
	\Dan E_{11} & = & 
					-\frac{
						n
					}{
						\bn^{0.5d\beta}
						(2\pi)^{0.5d\beta}
						(1+\beta)^{0.5d+1}
					}	\cdot
					\Dan
					\left\{	
						\frac{
							\Gamma(\an + 0.5d\beta)
						}{
							\Gamma(\an)
						}	
					\right\}			\nonumber \\
			& = &
				-\frac{
						n
					}{
						\bn^{0.5d\beta}
						(2\pi)^{0.5d\beta}
						(1+\beta)^{0.5d+1}
					}	\cdot
				\frac{
					\Gamma(\an + 0.5d\beta)
				}{
					\Gamma(\an)
				} \cdot
				\left[ 
					\Psi(\an + 0.5d\beta)
					- {\Psi(\an)}
				\right],			
\end{IEEEeqnarray}
where $\Psi^{(1)}$ denotes the trigamma function.

\subsubsection{Derivative with respect to $\bn$}

As for the other variational parameters, note that
\begin{IEEEeqnarray}{rCl}
	\Dbn\left\{ELBO\right\} & = & 
		\Dbn\left\{ E_8 \right\} + 
		\Dbn\left\{ E_9 \right\} + 
		\Dbn\left\{ E_{10} \right\} + 
		\Dbn\left\{ E_{11} \right\} + \nonumber \\ 
	&&  \hspace{-1.25cm} +\Dbn\left\{ E_2 \right\}\sum_{i=1}^n 
				\left\{ 
					\dfrac{
						E_{3,i}
					}{
						\left[ \bn + 
							0.5\left(
								\beta\*y_i'\*y_i + E_4 + E_{5,i}+ E_{6,i} + E_{7,i}
							\right)
						\right]^{\an + 0.5d\beta}
					}
				\right\} \nonumber \\[10pt]
	&& \hspace{-1.25cm} + E_2 \cdot 
		\sum_{i=1}^n
			\left\{ 
				{\ E_{3,i}	}\cdot
					\Dbn\left\{
					\left[ \bn + 
							0.5\left(
								\beta\*y_i'\*y_i + E_4 + E_{5,i}+ E_{6,i} + E_{7,i}
							\right)
						\right]^{-\an -0.5d\beta}	
					\right\}	
			\right\}, 
\end{IEEEeqnarray}
where the chain rule implies that
\begin{IEEEeqnarray}{rCl}
	\hspace*{-2cm}\Dbn\left\{
					\left[
						\underbrace{\bn + 
							0.5\left(
								\beta\*y_i'\*y_i + E_4 + E_{5,i}+ E_{6,i} + E_{7,i}
							\right)
						}_{ = K}
						\right]^{-\an -0.5d\beta}	
					\right\}	
	& = &
	\left( -\an - 0.5d\beta \right)\cdot K^{-\an -0.5d\beta-1}.
\end{IEEEeqnarray}
Thus one proceeds by the same logic as before.
\begin{IEEEeqnarray}{rCl}
	\Dbn E_2 & = & 		
					\frac{
							\an\Gamma(\an + 0.5d\beta) \cdot |\Sn^{-1}|^{0.5}  
						}{
								\beta (2\pi)^{0.5d\beta}\Gamma(\an)
						}
						\cdot 
						 \bn^{\an-1}  \\
		\Dbn E_8 & = &
					\frac{1}{2} \left[ 
						(\bez - \ben)'\Szinv(\bez - \ben) + 2\bz
					\right] \frac{\Gamma(\an + 1)}{\Gamma(\an)}
					\cdot\frac{1}{\bn^2 }
					 \\
		\Dbn E_9 & = & -\frac{\an}{\bn} \\
		\Dbn E_{10} & = & \frac{\an - \az}{\bn} \\
		\Dbn E_{11} & = &
					\dfrac{
						n d\beta \cdot
						\Gamma(\an + 0.5d\beta)
					}{
						2\cdot 
						\Gamma(\an)
						(2\pi)^{0.5d\beta}
						(1+\beta)^{0.5d+1}
					}\cdot \bn^{-0.5d\beta-1}
\end{IEEEeqnarray}

\section{Timing and performance comparisons: Markov Chain Monte Carlo vs Structured Variational Bayes}

{
We ran timing comparisons of \SVI with \MCMC for several subsets of the {well-log} data set. We ran the \BD \BOCPD algorithm implementing an \MCMC inference regime using \textit{stan} \citep{carpenter2016stan} and compared this with our \SVI inference regime. The two inference schemes were then run on 3 datasets of different time-length; the first 200 observations of the {well-log},  the first 500 observations of the {well log} and the full {well-log}, in order to show the impact changing the number of observations has on the timings for the algorithm. For the \SVI used to produce these timings, we perform \textit{full} optimization at every step, which is significantly slower than the \SGD-variant that we present in the paper and that can be found in our repository at \url{https://github.com/alan-turing-institute/rbocpdms}. In spite of this, the \SVI is still orders of magnitudes faster.}

\begin{table}[H]
\centering
\caption{Table of times to run the \BD \BOCPD algorithm under the \MCMC and \SVI with full optimization on the first 200 observations, the first 500 observations  and the full {well log} dataset. }
\label{Tab:MSEX}
\begin{tabular}{rrrr}
   \hline\\[-0.9em]
 & \textbf{T=200} & \textbf{T=500} & \textbf{T=4050} \\
   \hline\\[-0.9em]
 \MCMC & 7615.2 & 20388.7 & 106073.0\\ 
 \SVI (full optimization) & 102.8 & 328.5 & 3240.0\\ 
 \end{tabular}
 \end{table}
 
 Another question of interest is how much the Stochastic Gradient Descent (\SGD) inside our inference procedure provides robustness and how much the \BD itself is responsile for this. To put this question to the test, we ran full vs \SGD-based optimization on the well-log data. As shown in Fig. \ref{figure:well_log_}, the results are very close to identical: No \CPs are declared under one that are not declared in the other, and the run-length distribution's maximum coincides throughout. 
 
 \begin{figure}[h!]
\begin{center}
\centerline{\includegraphics[trim= {1.25cm 1.6cm 4.5cm 2.1cm}, clip, 
width=1.00\columnwidth]{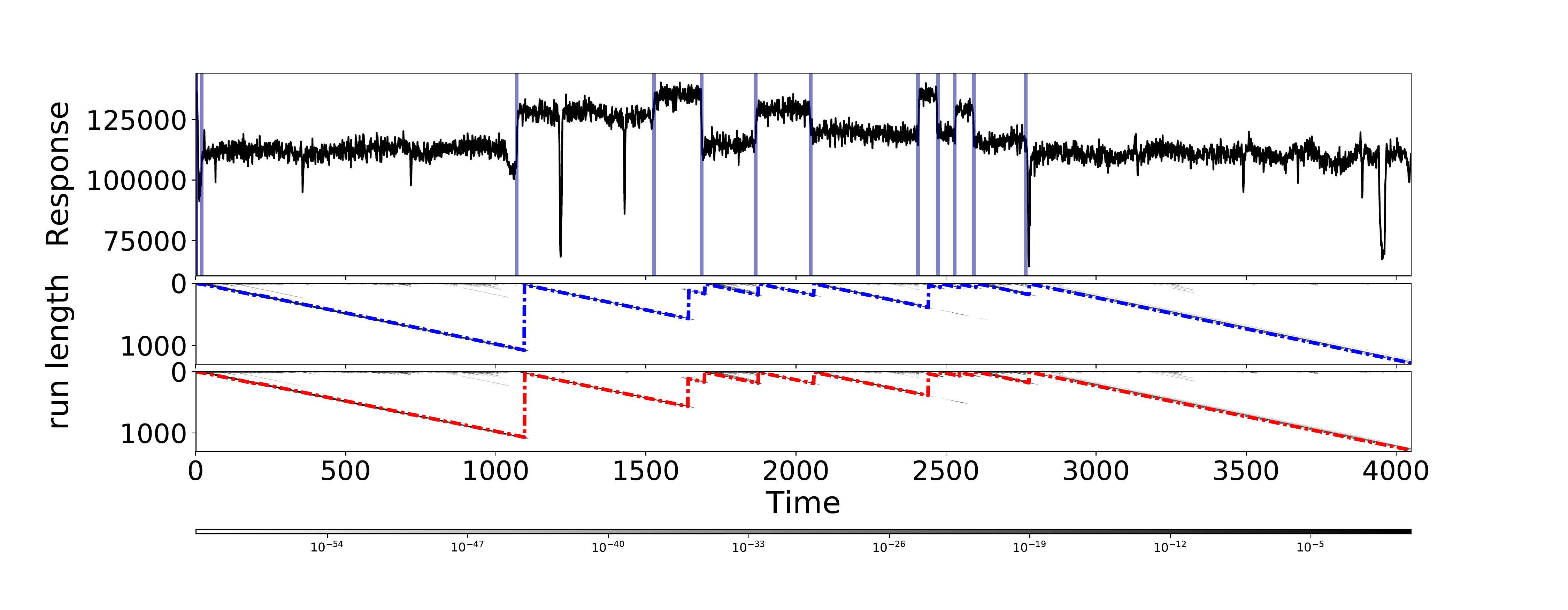}}
\caption{{ MAP segmentation} and run-length distributions of the well-log data. 
{\color{blue}\SGD inference} outcomes in blue, outcomes under {\color{red} full optimization} in red.
The corresponding run-length distributions for \SGD (middle) and full (bottom) optimization
are shown in grayscale with dashed maximum.}
\label{figure:well_log_}
\end{center}
\vskip -0.4in
\end{figure}
 
 \section{Initialization for $\beta_{\text{p}}$}
 
The initialization procedure described in the paper is illustrated in Fig. \ref{figure:init}. Here, the yellow dashed line gives a standard normal density corresponding to our model for the data. The gray dotted vertical line gives the amount of standard deviations from the posterior mean where one wishes to maximize the influence. We have chosen to maximize the influence at observations with $2.75$ standard deviations away from our posterior mean. In the first picture, $\beta_{\text{p}} = 0$ and thus the influence function corresponds to the Kullback-Leibler Divergence. Concordantly, it has no maximum and observations have more influence the further in the tail of our  model they occur. Thus, one needs to increase $\beta_{\text{p}}$ slightly. This is done in the second picture. While observations in the tail get smaller influence now than before, the influence of observations is still increasing beyond $2.75$ standard deviations. So one needs to increase $\beta_{\text{p}}$ two more times, until one finally obtains the desired outcome for $\beta_{\text{p}} = 0.25$ in the fourth picture. Notice that the influence does not immediately drop to $0$ for observations further in the tails than $2.75$ standard deviations away from the posterior mean, but it does decay. In some sense, we have set $\beta_{\text{p}}$ such that we think of observations occuring $2.75$ standard deviations away from the posterior mean as being \textit{most informative}. This is significantly different from what is implied by the Kullback-Leibler divergence, where an observation is most informative if it agrees least with the fitted model. It is intuitive why this produces good inferences if one is in the M-closed world and similarly intuitive why it does not in the M-open world.

\begin{figure}[h!]
\vskip -0.1in
\begin{center}
{\includegraphics[width=1\columnwidth, 
trim= {1.5cm 0.0cm 1.95cm 1.00cm}, clip]{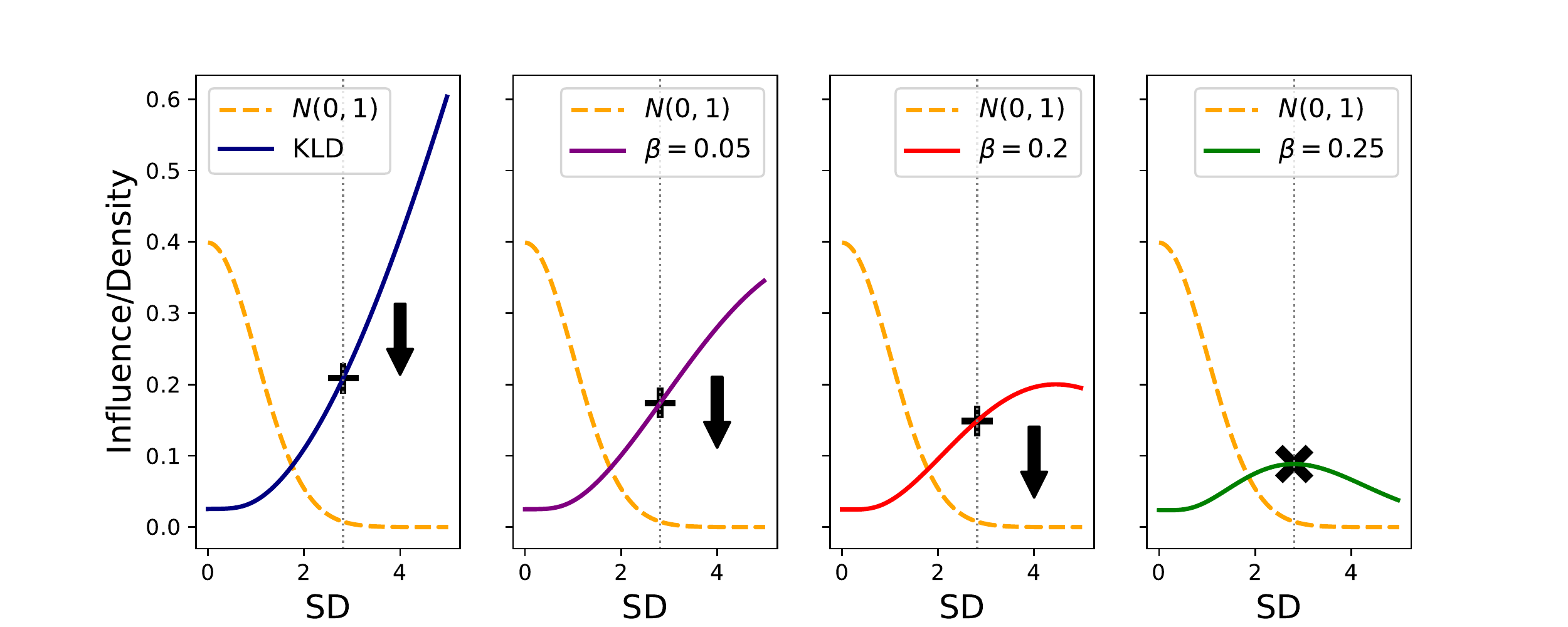}}
\caption{Illustration of the initialization procedure, from left to right.} 
\label{figure:init}
\end{center}
\vskip -0.1in
\end{figure}

\section{Recursive Optimization for $\brlmm$}
Recall that
\begin{IEEEeqnarray}{rCl}
 \widehat{\*y}_t(\*\beta) & = &
 	\sum_{r_t,m_t}\mathbb{E}\left(\*y_t|\*y_{1:(t-1)}, r_{t-1}, m_{t-1}, \bpm\right)p(r_{t-1}, m_{t-1}|\*y_{1:(t-1)}, \brlmm).
\end{IEEEeqnarray}
the issue reduces to finding the partial derivatives $\nabla_{\brlmm } \widehat{\*y}_t(\*\beta)$ and $\nabla_{\bpm } \widehat{\*y}_t(\*\beta)$. 
Notice that for $\nabla_{\brlmm } \widehat{\*y}_t(\*\beta)$, one finds that
\begin{IEEEeqnarray}{rCl}
	\nabla_{\brlmm } \widehat{\*y}_t(\*\beta) & = &
 	\sum_{r_t,m_t}\mathbb{E}\left(\*y_t|\*y_{1:(t-1)}, r_{t-1}, m_{t-1}, \bpm\right)\nabla_{\brlmm }p(r_{t-1}, m_{t-1}|\*y_{1:(t-1)}, \brlmm).
\end{IEEEeqnarray}
Observe now that for $p(\*y_{1:t}) = \sum_{r_t, m_t}p(r_t, m_t, \*y_{1:t}| \brlmm)$,
\begin{IEEEeqnarray}{lll}
&& \nabla_{\brlmm }p(r_t, m_t|\*y_{1:t}, \brlmm) \nonumber \\
& = & 
\nabla_{\brlmm }\left\{\dfrac{p(r_t, m_t, \*y_{1:t}| \brlmm) }{\sum_{r_t, m_t}p(r_t, m_t, \*y_{1:t}| \brlmm)}\right\} \nonumber \\
& = &
\frac{\nabla_{\brlmm }p(r_t, m_t, \*y_{1:t}| \brlmm) }{p(\*y_{1:t}) }
- \frac{p(r_t, m_t,\* y_{1:t}| \brlmm)}{p(\*y_{1:t})^2}
\cdot \sum_{r_t, m_t}\nabla_{\brlmm }p(r_t, m_t, \*y_{1:t}| \brlmm).
\end{IEEEeqnarray}
Thus we have reduced the problem to finding $\nabla_{\brlmm }p(r_t, m_t, \*y_{1:t}| \brlmm)$. Defining for a predictive posterior distribution $f_{m_t}(\*y_t|\mathcal{F}_{t-1} )$ its $\beta$-divergence analogue as
\begin{IEEEeqnarray}{rCl}
	f^{\brlmm}_{m_t}(\*y_t|\mathcal{F}_{t-1} )  & = &
	\exp\left\{ 
		\frac{1}{\brlmm}f_{m_t}(\*y_t|\mathcal{F}_{t-1} )^{\brlmm} 
		- \frac{1}{1+\brlmm}\int_{\mathcal{Y}}f_{m_t}(\*y_t|\mathcal{F}_{t-1} )^{1+\brlmm}d\*y_t
	\right\} \quad
\end{IEEEeqnarray}
 and uppressing the conditioning on $\brlmm$ for convenience, one can using the recursion
\begin{IEEEeqnarray}{rCl}
 \IEEEyesnumber\IEEEyessubnumber*
	p(\*y_{1:t}, r_t, m_t)
	& = &  \hspace*{-0.4cm}
	 \sum_{{m_{t-1}, {r_{t-1}}}}\hspace*{-0.4cm}	  \Bigl\{ 
	 	f^{\brlmm}_{m_t}(\*y_t|\mathcal{F}_{t-1} ) 
	 	q(m_t|\mathcal{F}_{t-1} , m_{t-1}) 
	 	H(r_{t},r_{t-1}) 
	 	p(\*y_{1:(t-1)}, r_{t-1}, m_{t-1})
	 	 \Bigr\},  \label{BOCPD_recursion3} \quad\quad\;
\end{IEEEeqnarray}
compute $\nabla_{\brlmm }p(r_t, m_t, \*y_{1:t})$ from $\nabla_{\brlmm }p(r_{t-1}, m_{t-1}, \*y_{1:(t-1)}| \brlmm)$ for $r_t = r_{t-1}+1$ as
 \begin{IEEEeqnarray}{rCl}
 	&& \nabla_{\brlmm }p(\*y_{1:t}, r_t, m_t) \nonumber \\
	& = &  
	 	 \Bigl\{ 
	 	\nabla_{\brlmm }f^{\brlmm}_{m_t}(\*y_t|\mathcal{F}_{t-1} ) 
	 	q(m_t|\mathcal{F}_{t-1} , m_{t-1}) 
	 	H(r_{t},r_{t-1}) 
	 	p(\*y_{1:(t-1)}, r_{t-1}, m_{t-1})
	 	 \Bigr\} +\nonumber \\	 	 
	 &&
	 	 \Bigl\{ 
	 	f^{\brlmm}_{m_t}(\*y_t|\mathcal{F}_{t-1} ) 
	 	\nabla_{\brlmm }q(m_t|\mathcal{F}_{t-1} , m_{t-1}) 
	 	H(r_{t},r_{t-1}) 
	 	p(\*y_{1:(t-1)}, r_{t-1}, m_{t-1})
	 	 \Bigr\} + \nonumber \\
	 &&
	 	\Bigl\{ 
	 	f^{\brlmm}_{m_t}(\*y_t|\mathcal{F}_{t-1} ) 
	 	q(m_t|\mathcal{F}_{t-1} , m_{t-1}) 
	 	H(r_{t},r_{t-1}) 
	 	\nabla_{\brlmm }p(\*y_{1:(t-1)}, r_{t-1}, m_{t-1})
	 	 \Bigr\}.
 \end{IEEEeqnarray}
 Similarly, for $r_t = 0$ the expression becomes
 \begin{IEEEeqnarray}{rCl}
 	&& \nabla_{\brlmm }p(\*y_{1:t}, r_t, m_t) \nonumber \\
	& = &  
	\nabla_{\brlmm }f^{\brlmm}_{m_t}(\*y_t|\mathcal{F}_{t-1} ) \cdot q(m_t) 
		\sum_{r_{t-1}, m_{t-1}}H(0,r_{t-1})p(\*y_{1:(t-1)}, r_{t-1}, m_{t-1}) + \nonumber \\
	&&f^{\brlmm}_{m_t}(\*y_t|\mathcal{F}_{t-1} ) \cdot q(m_t) 
		\sum_{r_{t-1}, m_{t-1}}H(0,r_{t-1})\nabla_{\brlmm }p(\*y_{1:(t-1)}, r_{t-1}, m_{t-1}).
\end{IEEEeqnarray}
This implies that if $f^{\brlmm}_{m_t}(\*y_t|\mathcal{F}_{t-1} )$ and $ q(m_t|\mathcal{F}_{t-1} , m_{t-1})$ are differentiable with respect to \brlm, then the entire expression can be updated recursively. For most exponential family likelihoods (and in particular the normal likelihood of the Bayesian Linear Regression), $\nabla_{\brlmm }f^{\brlmm}_{m_t}(\*y_t|\mathcal{F}_{t-1} )$ is available analytically. In particular, as long as $\int_{\mathcal{Y}}f_{m_t}(\*y_t|\mathcal{F}_{t-1} )^{1+\brlmm}d\*y_t$ has a closed form, $\nabla_{\brlmm }f^{\brlmm}_{m_t}(\*y_t|\mathcal{F}_{t-1} )$  can be found in analytic form.
In the case of Bayesian Linear Regression where the $d$-dimensional posterior predictive takes the shape of a student-$t$ distribution with $\nu$ degrees of freedom and posterior covariance $\frac{\nu}{\nu-2}\Sigma$, one finds that
\begin{IEEEeqnarray}{rCl}
	\nabla_{\brlmm }f^{\brlmm}_{m_t}(\*y_t|\mathcal{F}_{t-1})  & = &
	\nabla_{\brlmm }g_1(\brlmm)g_2(\brlmm)g_3(\brlmm) + \nonumber \\
	&&g_1(\brlmm)\nabla_{\brlmm }g_2(\brlmm)g_3(\brlmm) + \nonumber \\
	&&g_1(\brlmm)g_2(\brlmm)\nabla_{\brlmm }g_3(\brlmm),
\end{IEEEeqnarray}
where for $\eta = \nu d + d \brlmm + \nu$, 
\begin{IEEEeqnarray}{rCl}
	g_1(\brlmm) & = &
		\left(\dfrac{\Gamma(0.5[\nu+d])}{\Gamma(0.5\nu)}\right)^{1+\brlmm} \nonumber \\
	g_2(\brlmm) & = & \dfrac{\Gamma(0.5\eta)}{\Gamma(0.5[\eta+p])}\nonumber \\
	g_3(\brlmm) & = & (\nu\pi)^{-0.5p\cdot \brlmm} \cdot |\Sigma|^{-\brlmm}, \nonumber 
\end{IEEEeqnarray}
so that their derivatives are given by
\begin{IEEEeqnarray}{rCl}
	\nabla_{\brlmm }g_1(\brlmm) & =& -(\brlmm+1)\cdot \log( g_1(\brlmm)) \cdot g_2(\brlmm)\nonumber \\
	\nabla_{\brlmm }g_1(\brlmm) & =&
		0.5(\nu + p)\left[
			\cdot\dfrac{\Gamma(0.5\eta)\Psi(0.5\eta)}{\Gamma([0.5[\eta+p])}
			- \dfrac{\Gamma(0.5[\eta])\Psi(0.5[p+\eta])}{\Gamma([0.5[\eta+p])}
		\right]
	\nonumber \\
	\nabla_{\brlmm }g_3(\brlmm) & =&
		-g_3(\brlmm) \cdot \log(g_3(\brlmm))\cdot \frac{1}{\brlmm}
	\nonumber \\
\end{IEEEeqnarray}
As for $ \nabla_{\brlmm }q(m_t|\mathcal{F}_{t-1} , m_{t-1})$, one can again obtain it recursively, since for $r_t >0$,
\begin{IEEEeqnarray}{rCl}
	&& \nabla_{\brlmm } q(m_t|\mathcal{F}_{t-1} , m_{t-1}) \nonumber \\
	& = &
	\nabla_{\brlmm } \left\{ \dfrac{p(\*y_{1:(t-1)}, r_{t-1}, m_{t-1})}{\sum_{ m_{t-1}} p(\*y_{1:(t-1)}, m_{t-1})} \right\} \nonumber \\
	& = &
	\dfrac{ \nabla_{\brlmm }p(\*y_{1:(t-1)}, r_{t-1}, m_{t-1})}{\sum_{m_{t-1}} p(\*y_{1:(t-1)}, r_{t-1}, m_{t-1})}
	- \dfrac{\sum_{m_{t-1}} \nabla_{\brlmm } p(\*y_{1:(t-1)}, r_{t-1}, m_{t-1})  }
	{\left( {\sum_{m_{t-1}} p(\*y_{1:(t-1)}}, r_{t-1}, m_{t-1})\right)^2}.
\end{IEEEeqnarray}

\section{Proof of Theorem 2}

\begin{proof}
For ease of notation, we use $\bpm = \beta$.
The model used for the inference is an exponential family model of the form 

\begin{IEEEeqnarray}{rcl}
f(x;\theta)=\exp\left(\eta(\theta)^TT(x)\right)g(\eta(\theta))A(x),
\end{IEEEeqnarray}

where $g(\eta(\theta)):=\left(\int \exp\left(\eta(\theta)^TT(x)\right)A(x)dx\right)^{-1}$. Now under our \SVI routine the \BD posterior originating from this model and its conjugate prior is approximated  by a member of the conjugate prior family. As a result the conjugate prior and variational posterior to the above model have the form

\begin{IEEEeqnarray}{rcl}
\pi_0(\theta|\nu_0,\mathcal{X}_0)&=&g(\eta(\theta))^{\nu_0}\exp\left(\nu_0\eta(\theta)^T\mathcal{X}_0\right)h(\mathcal{X}_0,\nu_0)\\
\pi_n^{VB}(\theta|\nu_n,\mathcal{X}_n)&=&g(\eta(\theta))^{\nu_n}\exp\left(\nu_n\eta(\theta)^T\mathcal{X}_n\right)h(\mathcal{X}_n,\nu_n),
\end{IEEEeqnarray} 

where $\left(\nu_0,\mathcal{X}_0\right)$ are the prior hyperparameters, $\left(\nu_n,\mathcal{X}_n\right)$ represent the variational parameters and $h(\mathcal{X}_i,\nu_i):=\left(\int g(\eta(\theta))^{\nu_i}\exp\left(\nu_i\eta(\theta)^T\mathcal{X}_i\right)d\theta\right)^{-1}$. The resulting ELBO objective function under \GBI has the form 

\begin{IEEEeqnarray}{rcl}
ELBO&&\left(\nu_n,\mathcal{X}_n\right)=\nonumber\\
&&\mathbb{E}_{\pi^{VB}_n}\left[\log\left(\exp\left(\sum_{i=1}^n-\ell^D(x;\theta)\right)\right)\right]-d_{KL}\left(\pi^{VB}_n\left(\theta|\nu_n,\mathcal{X}_n\right),\pi_0\left(\theta|\nu_0,\mathcal{X}_0\right)\right),
\end{IEEEeqnarray}

where for the \BD posterior

\begin{IEEEeqnarray}{rcl}
-\ell^{\beta}(x;\theta)&=&\frac{1}{\beta}\left(\exp\left(\eta(\theta)^TT(x)\right)g(\eta(\theta))A(x)\right)^{\beta}-\nonumber\\
&&\qquad\frac{1}{\beta+1}\int \left(\exp\left(\eta(\theta)^TT(z)\right)g(\eta(\theta))A(x)\right)^{1+\beta}dz\\
&=&\frac{1}{\beta}\exp\left(\beta\eta(\theta)^TT(x)\right)g(\eta(\theta))^{\beta}A(x)^{\beta}-\nonumber\\
&&\qquad\frac{1}{\beta+1}\int \exp\left((1+\beta)\eta(\theta)^TT(z)\right)g(\eta(\theta))^{1+\beta}A(x)^{1+\beta}dz.
\end{IEEEeqnarray}

Therefore the $ELBO\left(\nu_n,\mathcal{X}_n\right)$ has three integrals that need evaluating

\begin{IEEEeqnarray}{rcl}
B_1&=&\sum_{i=1}^n\int \frac{1}{\beta}\exp\left(\beta\eta(\theta)^TT(x_i)\right)g(\eta(\theta))^{\beta}A(x_i)^{\beta}\pi_n^{VB}(\theta|\nu_n,\mathcal{X}_n)d\theta\label{Equ:ExpFamELBOLik}\\
B_2&=&\frac{n}{\beta+1}\int \left\lbrace\int \exp\left((1+\beta)\eta(\theta)^TT(z)\right)g(\eta(\theta))^{1+\beta}A(z))^{1+\beta}dz\right\rbrace\pi_n^{VB}(\theta|\nu_n,\mathcal{X}_n)d\theta\label{Equ:ExpFamELBOInt}\\
B_3&=&d_{KL}\left(\pi^{VB}_n\left(\theta|\nu_n,\mathcal{X}_n\right),\pi_0\left(\theta|\nu_0,\mathcal{X}_0\right)\right)\label{Equ:ExpFamELBOKL}.
\end{IEEEeqnarray}

Now firstly for the term $B_1$ in equation (\ref{Equ:ExpFamELBOLik})

\begin{IEEEeqnarray}{rcl}
B_1&=&\sum_{i=1}^n\int \frac{1}{\beta}\exp\left(\beta\eta(\theta)^TT(x_i)\right)g(\eta(\theta))^{\beta}A(x_i)^{\beta}g(\eta(\theta))^{\nu_n}\exp\left(\nu_n\eta(\theta)^T\mathcal{X}_n\right)h(\mathcal{X}_n,\nu_n)d\theta\\
&=&\sum_{i=1}^n\frac{1}{\beta}A(x_i)^{\beta}h(\mathcal{X}_n,\nu_n)\int g(\eta(\theta))^{\beta+\nu_n}\exp\left(\eta(\theta)^T\left(\beta T(x_i)+\nu_n\mathcal{X}_n\right)\right)d\theta\\
&=&\sum_{i=1}^n\frac{1}{\beta}A(x_i)^{\beta}h(\mathcal{X}_n,\nu_n)\frac{1}{h(\frac{\beta T(x_i)+\nu_n\mathcal{X}_n}{\beta+\nu_n},\beta+\nu_n)}.
\end{IEEEeqnarray}

Where we know that $h(\frac{\beta T(x_i)+\nu_n\mathcal{X}_n}{\beta+\nu_n},\beta+\nu_n)=\int g(\eta(\theta))^{\beta+\nu_n}\exp\left(\eta(\theta)^T\left(\beta T(x_i)+\nu_n\mathcal{X}_n\right)\right)d\theta$ is integrable and closed form as it represents the normalising constant of the same exponential family as the prior and the variational posterior. Next we look at $B_2$ in equation (\ref{Equ:ExpFamELBOInt}). The whole integral is the product of two densities which must be positive and in order for the $ELBO\left(\nu_n,\mathcal{X}_n\right)$ to be defined it must also be integrable. Therefore we can use Fubini's theorem to switch the order of integration 

\begin{IEEEeqnarray}{rcl}
&&B_2=\frac{n}{\beta+1}\int \left\lbrace\int \exp\left((1+\beta)\eta(\theta)^TT(z)\right)g(\eta(\theta))^{1+\beta}\pi_n^{VB}(\theta|\nu_n,\mathcal{X}_n)d\theta\right\rbrace A(z)^{1+\beta}dz\\
&=&\frac{n}{\beta+1}h(\mathcal{X}_n,\nu_n)\int \left\lbrace\int \exp\left(\eta(\theta)^T\left((1+\beta)T(z)+\nu_n\mathcal{X}_n\right)\right)g(\eta(\theta))^{1+\beta+\nu_n}d\theta\right\rbrace A(z)^{1+\beta}dz\\
&=&\frac{n}{\beta+1}h(\mathcal{X}_n,\nu_n)\int \frac{A(z)^{1+\beta}}{h(\frac{(1+\beta) T(z)+\nu_n\mathcal{X}_n}{1+\beta+\nu_n},1+\beta+\nu_n)}dz.
\end{IEEEeqnarray}

once again $h(\frac{(1+\beta) T(z)+\nu_n\mathcal{X}_n}{1+\beta+\nu_n},1+\beta+\nu_n)=\int \exp\left(\eta(\theta)^T\left((1+\beta)T(z)+\nu_n\mathcal{X}_n\right)\right)g(\eta(\theta))^{1+\beta+\nu_n}d\theta$ is the normalisisng constant of the same exponential family as the prior and the variational posterior and is thus closed form. Lastly we look at $B_3$ in equation (\ref{Equ:ExpFamELBOKL})

\begin{IEEEeqnarray}{rcl}
&&B_3=\int \pi^{VB}_n(\theta|\nu_n,\mathcal{X}_n)\log \frac{g(\eta(\theta))^{\nu_n}\exp\left(\nu_n\eta(\theta)^T\mathcal{X}_n\right)h(\mathcal{X}_n,\nu_n)}{g(\eta(\theta))^{\nu_0}\exp\left(\nu_0\eta(\theta)^T\mathcal{X}_0\right)h(\mathcal{X}_0,\nu_0)}\\
&=&\log \frac{h(\mathcal{X}_n,\nu_n)}{h(\mathcal{X}_0,\nu_0)}\int \pi^{VB}_n(\theta|\nu_n,\mathcal{X}_n)\left\lbrace\left(\nu_n-\nu_0\right)\log g(\eta(\theta))+\left(\eta(\theta)^T\left(\nu_n\mathcal{X}_n-\nu_0\mathcal{X}_0\right)\right)\right\rbrace\\
&=&\log \frac{h(\mathcal{X}_n,\nu_n)}{h(\mathcal{X}_0,\nu_0)}\left\lbrace\left(\nu_n-\nu_0\right)\lambda_{n}^{VB}+\left((\mu_n^{VB})^T\left(\nu_n\mathcal{X}_n-\nu_0\mathcal{X}_0\right)\right)\right\rbrace,
\end{IEEEeqnarray}

where $\mu_n^{VB}=\mathbb{E}_{\pi_n^{VB}}\left[\eta(\theta)\right]$ and $\lambda_{n}^{VB}=\mathbb{E}_{\pi_n^{VB}}\left[\log g(\eta(\theta))\right]$.


As a result we get that

\begin{IEEEeqnarray}{rcl}
ELBO(\nu_n,\mathcal{X}_n)&=&B_1-B_2-B_3\\
&=&\sum_{i=1}^n\frac{1}{\beta}A(x_i)^{\beta}h(\mathcal{X}_n,\nu_n)\frac{1}{h(\frac{\beta T(x_i)+\nu_n\mathcal{X}_n}{\beta+\nu_n},\beta+\nu_n)}\nonumber\\
&&\quad-\frac{n}{\beta+1}h(\mathcal{X}_n,\nu_n)\int \frac{A(z)^{1+\beta}}{h(\frac{(1+\beta) T(z)+\nu_n\mathcal{X}_n}{1+\beta+\nu_n},1+\beta+\nu_n)}dz\\
&&\quad -\log \frac{h(\mathcal{X}_n,\nu_n)}{h(\mathcal{X}_0,\nu_0)}\left\lbrace\left(\nu_n-\nu_0\right)\lambda_{n}^{VB}+\left((\mu_n^{VB})^T\left(\nu_n\mathcal{X}_n-\nu_0\mathcal{X}_0\right)\right)\right\rbrace\nonumber.
\end{IEEEeqnarray}
\end{proof}

\section{Complexity Analysis of Inference}

\textbf{Time complexity:} Our \SVRG method crucially hinges on the complexity of the gradient evaluations. For \BLR, we note that evaluating the complete \ELBO gradient derived above for $n$ observations has complexity $\mathcal{O}(n p^3)$, where $p$ is the number of regressors. We proceed by defining $g$ as the (generic) complexity of a gradient evaluation, so for \BLR $g=p^3$.
Clearly, an \SGD step using $b$ observations is of order $\mathcal{O}(bg)$. Similarly, the computation of the anchors is $\mathcal{O}(Bg)$. 
Next, let the optimization routine used for full optimization have complexity $\mathcal{O}(m(n, \text{dim}(\*\theta)))$. Most standard (quasi-) Newton optimization routines such as BFGS or LBFGSB (used in our implementation) are polynomial in $n$ and $\text{dim}(\*\theta)$. 
For such methods, since it holds that at most $W\geq n$ observations are evaluated in the full optimization, and since    $\text{dim}(\*\theta)$ is time-constant, $m(n, \text{dim}(\*\theta))$ is also constant in time. Thus, though these constants can be substantial, all  optimization steps (whether \SVRG steps or full optimization steps) are $\mathcal{O}(1)$ in time. Since one performs $T$ of them for $T$ observations, the computational complexity (in time) is $\mathcal{O}(T)$. 

\textbf{Space complexity}: One needs to store observations $\*y_t$ as well as gradient evaluations. Storing one of them takes $\mathcal{O}(d)$ and $\mathcal{O}(\text{dim}(\*\theta))$ space, respectively. Since we only keep a window $W$ of the most recent observations (and gradients), this means that the space requirement is of order $\mathcal{O}(W(d + \text{dim}(\*\theta)))$ and in particular constant in time.

\section{Additional Details on Experiments}

For all experiment, constrained Limited Memory  Broyden–Fletcher–Goldfarb–Shannon is used for the full optimization step, where the constraints are $\an > 1, \bn > 1$. We use Python's \texttt{scipy.optimize} wrapper, which calls a Fortran implementation.
We also tested whether inference is sensitive to different initializations of $\bpm$ and found that it is fairly stable as long as $\bpm$ is chosen reasonably. For example, for the Air Pollution data, we could recover the same changepoint ($\pm 5$ days) for initializations of $\bpm$ ranging from $0.005$ up to $0.1$.
All experiments were performed on a 2017 MacBook Pro with 16 GB 2133 MHz LPDDR3 and 3.1 GHz Intel Core i7.

\subsection{Well-log data}
\subparagraph{Hyperparameters:}
We set the hyperparameters for standard Bayesian On-line Changepoint Detection slightly differently, the reason being that due to the robustness guarantee of Theorem 1, we can use much less informative priors with the robust version than we can with the standard version: If priors are too flat, the standard version declares far too many changepoints. Thus, for the standard version, we use a constant \CP prior (hazard)
$H(r_t = r_{t-1} + 1| r_{t-1}) = 0.01$,  $a_0 = 1$, $b_0 = 10^4$, $\Sigma_0 = 0.25$, $\*\mu_0 = 1.15 \cdot 10^4$, while for the robust version we can use a less informative prior by instead setting $b_0 = 10^7$. 
By virtue of our initialization procedure for $\beta_p$, this implies setting $\beta_{p,0} \approx 0.05$. To start out close to the \KLD, we initialize $\beta_{\text{rld},0} = 0.0001$.
\subparagraph{Inferential procedure:}
For the robust version, we set $W = 360$, $B= 25$, $b= 10$, $m = 20$, $K = 1$. For both versions, only the $50$ most likely run-lengths are kept. 
For the robust version, the average processing time was $0.487$ per observation.

\subsection{Air Pollution data}

\textbf{Preprocessing \& Model Setup:} The air pollution data is observed every 15 minutes across 29 stations for 365 days. We average the 96 observations made over 24 hours. This is done to move the observed data closer to a normal distribution, as the measurements have significant daily volatility variations. To account for  weekly cycles, we also calculate for each station the mean for each weekday and subtract it from the raw data.. Yearly seasonality is not accounted for. 
Afterwards, the data is normalized station-wise. This is done only for numerical stability, because the internal mechanisms of the used VAR models perform matrix operations (QR-decompositions and matrix multiplications in particular) that can adversely affect numerical stability for observations with large absolute value. Fig. \ref{figure:APdat} shows some of the station's data after these preprocessing steps have been taken. 

 \begin{figure}[h!]
 \vskip -0.08in
 \begin{center}
 {\includegraphics[width=1\columnwidth, 
 trim= {2.5cm 0.0cm 2.5cm 0.00cm}, clip]{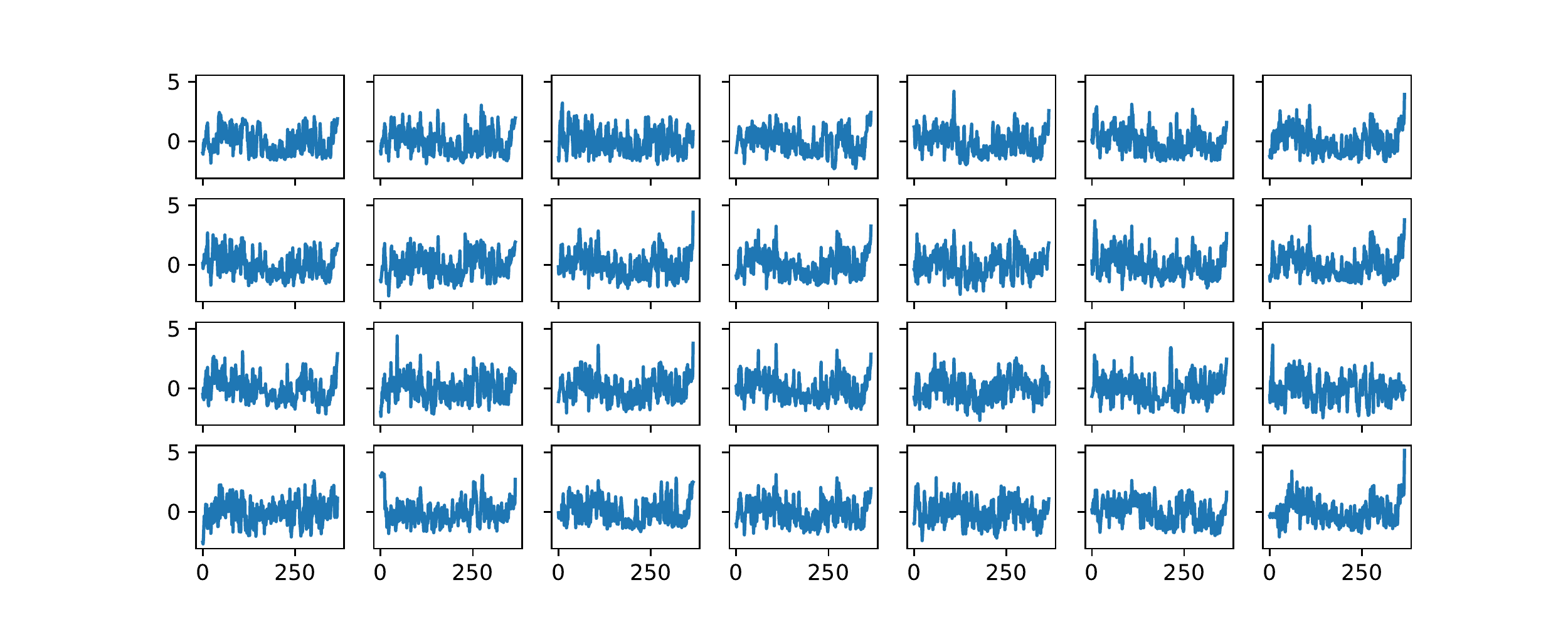}}
 \caption{Some of the stations after preprocessing steps. $x$-axis gives NOX level, $y$-axis the day.} 
 \label{figure:APdat}
 \end{center}
 \end{figure}

The autoregressive models and spatially structured vector autoregressive models (VARs) are chosen to have lag lengths $1,2,3$. These short lag lengths are chosen to explicitly disadvantage the robust model universe: The non-robust run we compare against uses more than $20$ models, with lag lengths $1, 5, 6, 7$, meaning that it is much more expressive and should be able to cope with outliers better. In spite of this, it not only declares more CPs, but also does worse than the robust version in terms of predictive performance.
For both the robust and non-robust model, two spatially structured VARs are included as in \cite{BOCPDMS}. 

%

\subparagraph{Hyperparameters:}
We set $H(r_t = r_{t-1} + 1| r_{t-1}) = 0.001$, $a_0 = 1$, $b_0 = 25$, $\*\mu_0 = \*0$, $\Sigma_0 = I\cdot 20$, which yields initialization $\bpm \approx 0.005$, $\brlmm = 0.1$. The non-robust results are directly taken from \cite{BOCPDMS} and can be replicated running the code available from \url{https://github.com/alan-turing-institute/bocpdms/}
\subparagraph{Inferential procedure:}
We set $W = 300$, $m = 50$, $B=20$ and $b=10$, $K=25$ and retain the $50$ most likely run-lengths. Processing times are more volatile than for the well-log because the full optimization procedure is significantly more expensive to perform. Most observations take significantly less than 20 seconds to process, but some take over a minute (depending on how many of the retained run-lengths are divisible by $m$ at each time point). 

\subsection{Optimizing $\beta$}

Lastly, we investigate the trajectories for $\beta$ as it is being optimized. For all trajectories, a bounded predictive absolute loss was used with threshold $\tau$, i.e. $L(x) = \max\{|x|, \tau \}$. For $\beta_{\text{rld}}$, $\tau = 5/T$ (where $T$ is the length of the time series) while for $\beta_{\text{p}}$, $\tau = 0.1$. The results are not sensitive to these thresholds, and they are picked with the intent that (1) a single observation should not affect $\beta_{\text{p}}$ by more than $0.1$ and (2) that overall, $\beta_{\text{rld}}$ should not change by more than $5$ in absolute magnitude. As the initialization procedure for $\beta_{\text{p}}$ works very well for predictive performance, the on-line optimization never even comes close to making a step with size $\tau$. The picture is rather different for $\beta_{\text{rld}}$, which reaches $\tau$ rather often. We note that this is because the estimated gradients for $\beta_{\text{rld}}$ can be very extreme, which is why the implementation averages $50$ consecutive gradients before performing a step. 
Overall, we note that for the well log data whose trajectories are depicted in Fig. \ref{figure:beta_trajec_well}, the degrees of robustness do not change much relative to their starting points at $\beta_{\text{p}} = 0.05$ and $\beta_{\text{rld}} = 0.001$. In particular, the absolute change over more than $4,000$ observations is $<0.002$ for $\beta_{\text{p}}$ and $<0.015$ for $\beta_{\text{rld}}$. Step sizes are $1/t$ at time $t$. 
\begin{figure}[h!]
 \vskip -0.08in
 \begin{center}
 {\includegraphics[width=1\columnwidth, 
 trim= {0cm 1.0cm 0cm 1.00cm}, clip]{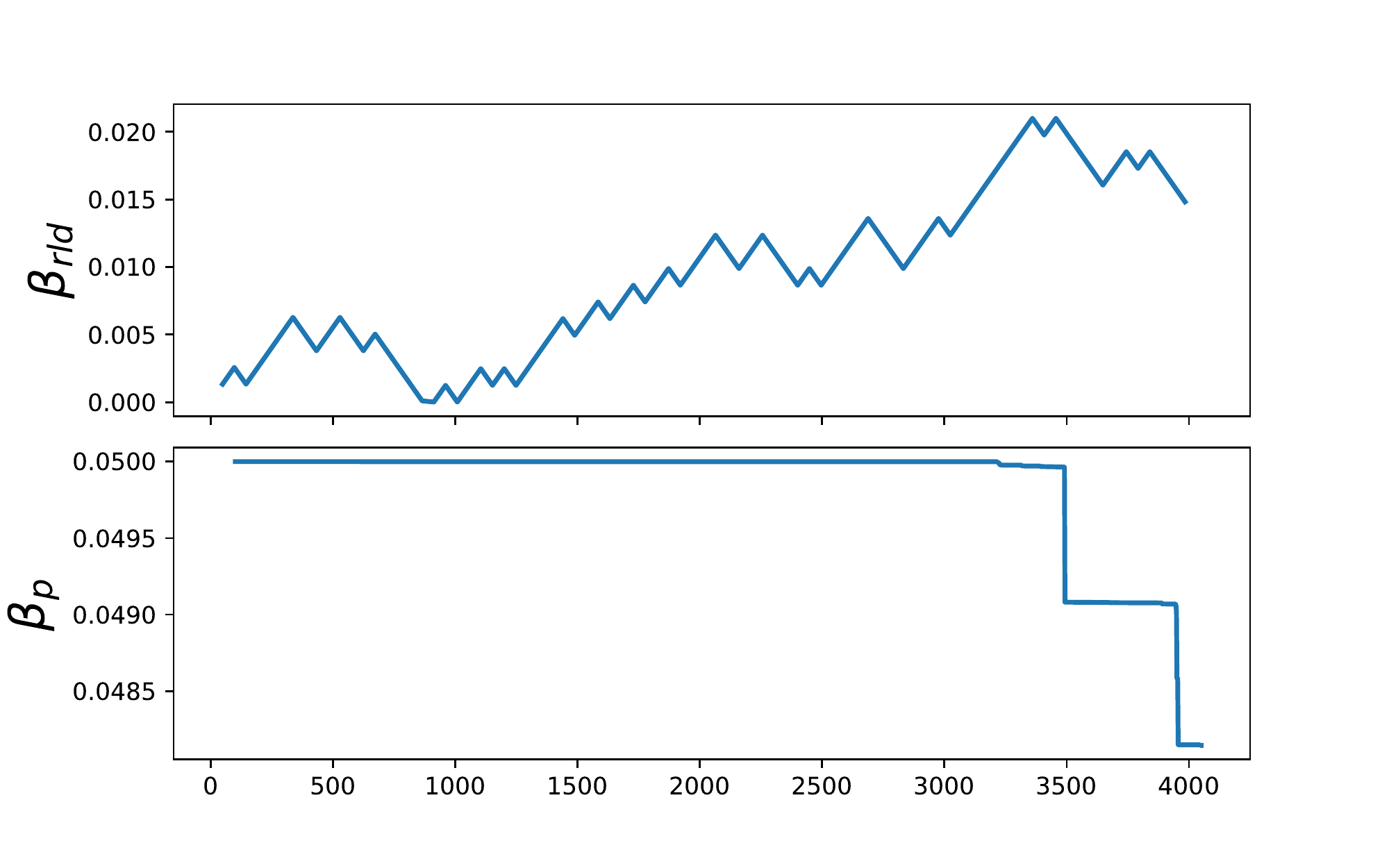}}
 \caption{$\beta$ trajectories for the well-log data. For $\beta_{\text{rld}}$, steps are only taken every $50$ observations to average gradient noise} 
 \label{figure:beta_trajec_well}
 \end{center}
 \end{figure}
 
For the Air Pollution Data, the story is slightly different: Here, $\beta_{\text{p}}$ does not change after the first iteration, where it jumps from $0.005$ directly to $10^{-10}$. While this seems odd, it is mainly due to the fact that for numerical stability reasons\footnote{In particular, working with the \BD implies that one takes the exponential of a density, i.e. $e^{f^{\beta}}$. So even working on a log scale now means working with the densities $f^{\beta}$ \textit{directly}. It should be clear that these quantities become numerically unstable for $\beta$ too large or too small.} , one needs to ensure that $\beta_{\text{p}} > \varepsilon$ for some $\varepsilon>0$; and in our implementation, $\varepsilon = 10^{-10}$. The interpretation of the trace graph is thus that the optimization continuously suggests less robust values for $\beta_{\text{p}}$, but that we cannot admit them due to numerical stability. 
 The downward trend also holds for $\beta_{\text{rld}}$, which is big enough to not endanger numerical stability and hence can drift downwards. 

 Fig. \ref{figure:beta_trajec_well} also shows that the optimization technique used for $\beta$ needs further investigation and research. For starters, the outcomes suggest that a second order method could yield better results than using a first-order \SGD technique. In the future, we would like to explore this in greater detail and also explore more advanced optimization methods like line search or trust region optimization methods for this problem.
\bibliography{library}
\bibliographystyle{plainnat}

\end{document}